\documentclass[runningheads]{llncs}
\usepackage{graphicx}
\usepackage{style}
\usepackage[misc,geometry]{ifsym}

\begin{document}
\title{Local Branching Relaxation Heuristics for \\Integer Linear Programs}
%
%

\author{Taoan Huang (\Letter)\inst{1}\orcidID{0000-0002-8733-2379} \and
Aaron Ferber\inst{1}\orcidID{0000-0002-7422-0044} \and
Yuandong Tian\inst{2}\orcidID{0000-0003-4202-4847} \and
Bistra Dilkina\inst{1}\orcidID{0000-0002-6784-473X} \and \\
Benoit Steiner \inst{2}\orcidID{0000-0002-5767-4976}
}
%
\authorrunning{T. Huang et al.}
%
\institute{University of Southern California, Los Angeles, USA \email{\{taoanhua,aferber,dilkina\}@usc.edu}
\and Meta AI (FAIR), Menlo Park, USA \\ \email{yuandong@meta.com, benoit.papers@gmail.com}
}

\maketitle              
\begin{abstract}

Large Neighborhood Search (LNS) is a popular heuristic algorithm for solving combinatorial optimization problems (COP). It starts with an initial solution to the problem  and iteratively improves it by searching a large neighborhood around the current best solution. LNS relies on heuristics to select neighborhoods to search in. In this paper, we focus on designing effective and efficient heuristics in LNS for integer linear programs (ILP) since a wide range of COPs can be represented as ILPs.  Local Branching (LB) is a heuristic that selects the neighborhood that leads to the largest improvement over the current solution in each iteration of LNS. LB is often slow since it needs to solve an ILP of the same size as input. Our proposed heuristics, \LBRELAX and its variants, use the linear programming relaxation of LB to select neighborhoods. Empirically, \LBRELAX and its variants compute as effective neighborhoods as LB but run faster. They achieve state-of-the-art anytime performance on several ILP benchmarks.

\end{abstract}

\section{Introduction}
Combinatorial optimization problems (COP) concerns a wide variety of real-world applications, including vehicle routing \cite{toth2002vehicle}, path planning \cite{pohl1970heuristic} and resource allocation \cite{manne1960job} problems. 
Many of them are difficult to solve with limited computational resources due to their NP-Hardness.
Nonetheless, the widespread importance of COPs has inspired research in designing algorithms for solving them, including exact algorithms, approximation algorithms, heuristic algorithms and data-driven algorithms.

In this paper, we focus specifically on Integer Linear Programs (ILPs) since it is a powerful tool to model and solve a broad collection of COPs, including graph optimization \cite{song2020general}, mechanism design \cite{de2003combinatorial}, facility location \cite{heragu1991efficient,amaral2008exact} and network design \cite{dilkina2010solving,huang2020enhancing} problems. Branch-and-Bound (BnB) is an optimal and complete tree search algorithm and is one of the state-of-the-art algorithms for ILPs \cite{land2010automatic}. It is also the core of many ILP solvers such as SCIP \cite{BestuzhevaEtal2021OO} and Gurobi \cite{gurobi}. Huge research effort has been made to improve it over the past decades \cite{achterberg2013mixed}. 
However, BnB still falls short of delivering practical impact due to scalability issues \cite{khalil2016learning,gasse2019exact}.
On the other hand, Large Neighborhood Search (LNS) is a powerful heuristic algorithm for hard COPs and has been recently applied to solve ILPs \cite{song2020general,wu2021learning,sonnerat2021learning} in the machine learning (ML) community. 

To solve ILPs, LNS starts with an initial solution, i.e., a feasible assignment of values to the variables. It then iteratively improves the best solution found so far (i.e., the \textit{incumbent solution}), by applying {\it destroy heuristics} to select a subset of variables and solving a sub-ILP that optimizes only the selected variables while leaving others fixed. ML-based destroy heuristics are shown to be efficient and effective but they are often tailored for a specific problem domain and require extensive computational resources for learning.  A few non-ML destroy heuristics have been studied, such as the randomized heuristics \cite{song2020general,sonnerat2021learning} and the Local Branching (LB) heuristic \cite{fischetti2003local,sonnerat2021learning}, but they are either less efficient or effective compared to the ML-based ones.  The randomized heuristics select the neighborhood by quickly randomly sampling a subset of variables which is often of bad quality. LB computes the optimal solution across all possible search neighborhoods that differs from the current incumbent solutions on a limited number of variables; however, LB is computationally expensive since it requires solving an ILP that has the same size as the original problem. 

To strike a balance between efficiency and effectiveness, we propose a simple yet effective destroy heuristic \LBRELAX that is based on the linear programming (LP) relaxation of LB. Instead of solving an ILP to find the neighborhood as LB does, \LBRELAX computes its LP relaxation. It then selects the variables greedily based on the difference between the values in the incumbent solution and the LP relaxation solution. We also propose two other variants, \LBRELAXS and \LBRELAXRR, that deploy a sampling method and combine the randomized heuristic with \LBRELAX to help escape local optima more efficiently, respectively. In experiments, we compare \LBRELAX and its variants against LNS with baseline destroy heuristics and BnB on several ILP benchmarks and show that they achieve state-of-the-art anytime performance. We also show that \LBRELAX achieves competitive results with, sometimes even outperform, the ML-based destroy heuristics. 
We also test \LBRELAX and its variants on selected difficult MIPLIB instances \cite{MIPLIB} that encompass diverse problem domains, structures and sizes and show that they achieve best performance on at least 40\% of the instances. We also empirically show that \LBRELAX and \LBRELAXS find neighborhoods of similar quality but is much faster than LB. They sometimes even outperform LB due to LB being too slow to find good enough neighborhoods within a reasonable time cutoff.


\vspace{-0.05in}
\section{Background}
\vspace{-0.05in}
In this section, we first define ILP and introduce its LP relaxation. We then introduce LNS for ILP solving and the Local Branching (LB) heuristic.

\subsection{ILP and its LP Relaxation}

An \textit{integer linear program (ILP)} is defined as 
\[\min \bc^{\mathsf{T}}\bx\quad \textrm{ s.t. } \bA \bx\leq \bb \textrm{ and } \bx \in \{0,1\}^n,\]
where $\bx = (x_1,\ldots, x_n)^\sfT$ denotes the $n$ binary variables to be optimized, $\bc\in \mathbb{R}^n$ denotes the vector of objective coefficients and $\bA\in \mathbb{R}^{m\times n}$ and $\bb\in \mathbb{R}^{m}$ specify $m$ linear constraints.
A \textit{solution} to the ILP is an feasible assignment of values to the variables.

The \textit{linear programming (LP) relaxation} of an ILP is obtained by relaxing  binary variables in the ILP to continuous variables between 0 and 1, i.e., by replacing the integer constraint $\bx\in \{0,1\}^n$ with  $\bx\in[0,1]^n$. 

Note that, in this paper, we focus on the formulation above that consists of only binary variables, but our methods can also be applied to mixed integer linear programs with continuous variables and/or non-binary integer variables.  

\begin{algorithm}[t]
\small
	\caption{LNS for ILPs}\label{algo::LNSforILP}
	\begin{algorithmic}[1]
	     \State {\bf Input: } An ILP.
	     \State $\bx^0\gets$ Find an intial solution to the input ILP
	     \State $t\gets 0$
	    \While {time limit not exceeded}
		    \State $\calX^t\gets$ Select a subset of variables to destroy
		    \State $\bx^{t+1}\gets$ Solve the ILP with additional constraints $\{x_i = x_i^t: x_i\notin \calX^t\}$
		    \State $t\gets t + 1$
		    
		\EndWhile
		\State \Return $\bx^t$
	\end{algorithmic}
\end{algorithm}

\subsection{LNS for ILP solving}

LNS is a heuristic algorithm that starts with an initial solution and then iteratively reoptimizes a part of the solution by applying the destroy and repair operations until a time limit is exceeded. Let $\bx^0$ be the initial solution. In iteration $t\geq 0$ of the LNS, given the \textit{incumbent solution} $\bx^t$,  defined as the best solution found so far, a destroy operation is done by a \textit{destroy heuristic} where it selects a subset of $k_t$ variables $\calX^t= \{x_{i_1},\ldots, x_{i_{k_t}}\}$. The repair operation is done by solving a sub-ILP with $\calX^t$ being the variables while fixing the values of $x_j\notin \calX^t$ to be the same as in $\bx^t$.
Compared to BnB, LNS is more effective in improving the objective value $\bc^{\mathsf{T}}x$, or the primal bound, especially on difficult instances \cite{song2020general,sonnerat2021learning,wu2021learning}. Compared to other local search methods, LNS explores a large neighborhood in each step and thus, is more effective in avoiding local minima. LNS for ILPs is summarized in Algorithm \ref{algo::LNSforILP}.

\subsection{LB Heuristic}

The LB Heuristic \cite{fischetti2003local} is originally proposed as a primal heuristic in BnB but is also applicable in LNS for ILP solving \cite{sonnerat2021learning,liurevisiting}. 
Given the incumbent solution $\bx^t$ in iteration $t$ of LNS, the LB heuristic \cite{fischetti2003local} aims to find the subset of variables to destroy $\calX^t$ such that it leads to the optimal $\bx^{t+1}$ that differs from $\bx^t$ on at most $k_t$ variables, i.e., it computes the optimal solution $\bx^{t+1}$ that sits within a given Hamming ball of radius $k_t$ centered around $\bx^t$.
To find $\bx^{t+1}$, the LB heuristic solves the LB ILP that is exactly the same ILP from input but with one additional constraint that limits the distance between $\bx^t$ and $\bx^{t+1}$:
$$\sum_{i\in[n]:x^t_i=0}x^{t+1}_i + \sum_{i\in[n]:x^t_i=1}(1-x^{t+1}_i)\leq k_t. $$
The LB ILP is of the same size of the input ILP (i.e., it has the same number of variables and one more constraint), therefore, it is often slow to run in practice.

\section{Related Work}

In this section, we summarize related work on LNS for ILPs, LNS-based primal heuristics in BnB and LNS for other COPs. 

\subsection{LNS for ILPs}
While a lot of effort has been made to improve BnB for ILPs in the past decades, LNS for ILPs has not been studied extensively in the past. Recently, Song et al. \cite{song2020general} show that even a randomized destroy heuristic in LNS can outperform state-of-the-art BnB in runtime. In the same paper, they show that an ML-guided decomposition-based LNS can achieve even better performance, where they apply reinforcement learning and imitation learning to learn the destroy heuristics.
Since then, there have been a few more recent studies on ML-based LNS for ILPs.  Sonnerat et at. \cite{sonnerat2021learning} learn to select variables to destroy via imitating LB. Wu et al. \cite{wu2021learning} learn the same thing but they  use reinforcement learning instead. 
The main difference between \LBRELAX and ML-based heuristics is that \LBRELAX does not require extra computational resource for learning and is agnostic to the underlying problem distributions. \LBRELAX also has a better balance between efficiency and effectiveness than those existing non-ML heuristics.

\subsection{LNS-based primal heuristics in BnB}

LNS-based primal heuristics is one of the rich set of primal heuristics in BnB for ILPs and many techniques have been proposed in past decades. With the same purpose of improving primal bounds of the ILPs, the main differences between the LNS-based primal heuristics in BnB and LNS for ILPs are the following: (1) Since LNS-based primal heuristics are often more expensive to run than the others in BnB, they are executed periodically at different search tree nodes during the main search and the execution schedule is itself dynamic; (2) the destroy heuristics for LNS in BnB are often designed to use information, such as the dual bound and the LP relaxation at a search tree node, that is specific to BnB and not directly applicable in LNS for ILPs in our setting. 

Next, we briefly summarize the destroy heuristics in LNS-based primal heuristics. The Crossover heuristics \cite{rothberg2007evolutionary} destroy variables that have different values in a set of selected known solutions (typically two). The Mutation heuristics \cite{rothberg2007evolutionary} destroys a random subset of variables. 
Relaxation Induced Neighborhood Search (RINS) \cite{danna2005exploring} destroys variables whose values disagree in the solution of the LP relaxation at the current search tree node and the current incumbent solution. 
Relaxation Enforced Neighborhood Search (RENS) \cite{berthold2014rens} restricts the neighborhood to be the feasible roundings of the LP relaxation at the current search tree node. Local Branching \cite{fischetti2003local} restricts the neighborhood to a ball around the current incumbent solution. Distance Induced Neighborhood Search (DINS) \cite{ghosh2007dins} takes the intersection of the neighborhoods of the Crossover, LB and RINS heuristics. Graph-Induced Neighborhood Search (GINS) \cite{maher2017scip} destroys the breadth-first-search neighborhood of a variable in the bipartite graph representation of the ILP. An adaptive LNS primal heuristic that essentially solves a multi armed bandit problem has been proposed to combine the power of these heuristics \cite{hendel2022adaptive}. 

\LBRELAX is closely related to RINS \cite{danna2005exploring} since they both use LP relaxations to select  neighborhoods. However, RINS is more suitable in BnB since it can adapt dynamically to the constraints added by branching. It uses the LP relaxation of the original problem, whereas \LBRELAX uses that of the LB ILP which takes into account the incumbent solutions that could change from iteration to iteration in LNS.

\subsection{LNS for other COPs}

LNS has been applied to solve a wide range of COPs, such as the vehicle routing problem \cite{ropke2006adaptive,azi2014adaptive}, the traveling salesman problem \cite{smith2017glns}, scheduling problems \cite{kovacs2012adaptive,vzulj2018hybrid} and path planning problems \cite{LiIJCAI21,LiAAAI22,HuangICAPS23}. Recently, ML-based methods have been applied to improve LNS for those applications \cite{chen2019learning,lu2019learning,hottung2020neural,li2021learning,huang2022anytime}.

\section{The Local Branching Relaxation Heuristic} \label{sec::methodolgy} 

Recently, designing effective destroy heuristics in LNS for ILPs has been a focus in the ML community \cite{song2020general,sonnerat2021learning,wu2021learning}. However, it is difficult to apply ML-based destroy heuristics to general ILPs since they are often customized for ILPs from certain problem distributions, e.g., graph optimization problems from a given graph distribution or scheduling problems where resources and demands follow the distribution of historical data, and require extra computational resources for training.  There has been a lack of study on destroy heuristics that are agnostic to the underlying distribution of the problem. Existing ones such as randomized heuristics are simple and fast but  sometimes not effective \cite{song2020general,sonnerat2021learning}. LB are effective but not efficient \cite{sonnerat2021learning,liurevisiting} since it exhaustively solves an ILP the same size as input for the best improvement. 

There are well-known approximation algorithms for NP-hard COPs based on LP relaxation \cite{kleinberg2006algorithm}. Typically, they  solve the LP relaxation of the ILP of the original problem and  apply deterministic or randomized rounding afterwards to construct an integral solution. These algorithms often have theoretical guarantee on the effectiveness and are fast, since LP can be solved in polynomial time.  Inspired by those algorithms, we propose destroy heuristic \LBRELAX that first solves the LP relaxation of the LB ILP and then constructs the neighborhood (selects variables $\calX^t$ to destroy) based on the LP relaxation solution. 
Specifically, given an ILP and the incumbent solution $\bx^t$ in iteration $t$, we construct the LB ILP with neighborhood size $k_t$ and solve its LP relaxation. Let $\bar{\bx}^{t+1}$ be the LP relaxation solution to the LB ILP. Also, let $\Delta_i = |\bar{x_i}^{t+1}- x_i^t|$ and $\bar{\calX}^t = \{x_i: \Delta_i> 0, i\in[n]\}$. 
$\Bar{\calX}^t$ includes all the fractional variables in the LP relaxation solution and all integral variables that have different values from $\bx^t$. In the following, we introduce (1) \LBRELAX, (2) \LBRELAXS, a variant of \LBRELAX with randomized sampling and (3) \LBRELAXRR, another variant of \LBRELAX that combines a randomized destroy with \LBRELAX to help avoid local minima more effectively.

{\bf \LBRELAX} first gets the LP relaxation solution $\bar{\bx}^{t+1}$ of the LB ILP and then calculates $\Delta_i$ and $\bar{\calX}^t$ from $\bar{\bx}^{t+1}, \bx^t$. To construct $\calX^t$ (the set of variables to destroy), it then greedily selects $k_t$ variables with the largest $\Delta_i$ and breaks ties uniformly at random. Intuitively, \LBRELAX greedily selects the variables whose values are more likely to change in the incumbent solution $\bx^t$ after solving the LB ILP.  \LBRELAX is summarized in Algorithm \ref{algo::LBRELAX}. Instead of using the LP relaxation of the LB ILP, one could argue that we alternatively use that of the original ILP similar to RINS \cite{danna2005exploring}. However, the advantage of \LBRELAX over using the LP relaxation of the original problem is that, by approximating the solution to the LB ILP, \LBRELAX selects neighborhoods based on the incumbent solutions that change from iteration to iteration, whereas the original LP relaxation is a static and less informative feature that is pre-computed before the LNS procedure. 

\def\NoNumber#1{{\def\alglinenumber##1{}\State #1}\addtocounter{ALG@line}{-1}}

\begin{algorithm}[t]
\small
	\caption{\LBRELAX ({\color{blue}\LBRELAXS})}\label{algo::LBRELAX}
	\begin{algorithmic}[1]
	     \State {\bf Input: } An ILP, incumbent solution $\bx^t$ and neighborhood size $k_t$.
        \State Construct the LB ILP given $\bx^t$ and $k_t$
	     \State $\bar{\bx}^{t+1}\gets$ Solve the LP relaxation of the LB ILP
	     \State $\Delta_i \gets |\bar{x_i}^{t+1}-x_i^t|$ for all $i\in[n]$
        \State $\bar{\calX}^{t} \gets \{x_i:\Delta_i >0, i\in[n]\}$
        \If {$|\bar{\calX}^{t}| \geq k_t$}
        \State $\calX^t\gets$  Select $k_t$ variables greedily with the largest $\Delta_i$ from $\bar{\calX}^{t}$
         \NoNumber{({\color{blue}$\calX^t\gets$  Select $k_t$ variables uniformly at random from $\bar{\calX}^{t}$}) }
        \Else  
        \State $\calX'\gets$ a random subset of $k_t-|\bar{\calX}^{t}|$ variables from $\{x_i:\Delta_i = 0, i\in [n]\}$
        \State $\calX^{t} \gets \bar{\calX}^{t} \cup \calX'$  
        \EndIf
	\State \Return $\calX^t$
	\end{algorithmic}
\end{algorithm}

{\bf \LBRELAXS}  is a variant of \LBRELAX with randomized sampling. To construct $\calX^t$, instead of greedily choosing variables with the largest $\Delta_i$, it selects $k_t$ variables 
from $\bar{\calX}^t$ uniformly at random. If $|\bar{\calX}^t|< k_t$, it selects all variables from $\bar{\calX}^t$ and $k_t-|\bar{\calX}^t|$ variables from the remaining uniformly at random. \LBRELAX is summarized in Algorithm \ref{algo::LBRELAX} where the parts in blue highlight the differences between \LBRELAX and \LBRELAXS. 
Since $0\leq \Delta_i\leq 1$, one could treat $\Delta_i$ as a probability distribution and sample $k_t$ variables accordingly (see \cite{sonnerat2021learning} for an example of how to normalize the distribution to sample $k_t$ variables). However, this variant performs similarly to or slightly worse than \LBRELAXS empirically and require extra hyperparameter tunings for the normalization. We therefore omit it and focus on the simpler variant in this paper.

{\bf \LBRELAXRR} is another variant of \LBRELAX that leverages a randomized destroy to avoid local minima more effectively. Once \LBRELAX fails to find an improving solution in iteration $t$, if we let $k_{t+1} = k_t$, it will solve the exact same LP relaxation of the LB ILP again in the next iteration since the incumbent solution $\bx^{t+1}=\bx^{t}$ and the neighborhood size stay the same. Also, since \LBRELAX uses a greedy rule, it will select the same set of variables with the largest $\Delta_i$'s deterministically, except that it might need to break ties randomly in some cases when there are multiple variables with the same $\Delta_i$. Therefore, it is susceptible to getting stuck at local minima. To tackle this issue, once \LBRELAX fails to find a new incumbent solution, we update $k_{t+1}$ using the adaptive method described in the next paragraph. If it fails again in the next iteration, we switch to a randomized destroy heuristic that uniformly samples variables at random without replacement to construct the neighborhood. We switch back to \LBRELAX after running the randomized destroy heuristic for at least $\gamma$ seconds and a new incumbent solution is found. 

Next, we discuss an adaptive method to set the neighborhood size $k_t$ for \LBRELAX and its variants. The initial neighborhood size $k_0$ is set to a constant or a fraction of the number of variables in the input ILP. In iteration $t$, if LNS finds a new incumbent solution, we let $k_{t+1}=k_t$. Otherwise, we increase $k_t$ by a factor $\alpha >1$. Also, we upper bound the neighborhood size $k_t$ to a fraction $\beta<1$ of the number of variables to make sure the sub-ILP in each iteration is not too difficult to solve, i.e., we let $k_{t+1} =\min\{ \alpha\cdot k_t, \beta\cdot n\}.$ This adaptive way of choosing $k_t$ also helps address the issue of local minima by expanding the search neighborhood when LNS fails to improve the solution. It is applicable to not only  \LBRELAX and its variants but also any destroy heuristics that require a given neighborhood size $k_t$.








\section{Empirical Evaluation}\label{sec::experiment}

In this section, we demonstrate the efficiency and effectiveness of \LBRELAX and its variants through extensive experiments on ILP benchmarks. 

\subsection{Setup}

\subsubsection{Instance Generation} We evaluate on four NP-hard problem benchmarks selected from previous work \cite{wu2021learning,song2020general,scavuzzo2022learning}, which consist of synthetic  minimum vertex cover (MVC), maximum independent set (MIS), set covering (SC) and multiple knapsack (MK) instances.  MVC and MIS instances are generated according to the Barabasi-Albert random graph model \cite{albert2002statistical}, with 9,000 nodes and average degree 5 following \cite{song2020general}. SC instances are generated with 4,000 variables and 5,000 constraints following \cite{wu2021learning}. MK instances are generated with 400 items and 40 knapsacks following \cite{scavuzzo2022learning}. For each problem, we generate 100 instances. 


\subsubsection{Baselines}

We compare \LBRELAX, \LBRELAXRR and \LBRELAXS with the following baselines:
\begin{itemize}
    \item BnB using SCIP (v8.0.1) as the solver with the aggressive mode turned on to focus on improving the primal bound;
    \item LB: LNS which selects the neighborhood with the LB heuristics;
    \item \RANDOM: LNS which selects the neighborhood by uniformly sampling a subset of variables of a given neighborhood size $k_t$;
    \item \GRAPH: LNS which selects the neighborhood based on the bipartite graph representation of the ILP similar to GINS \cite{maher2017scip}. 
    A bipartite graph representation consists of nodes representing the variables and constraints on two sides, respectively, with an edge connecting a variable and a constraint if a variable has a non-zero coefficient in the constraint.
    It runs a breadth-first search starting from a random variable node in the bipartite graph and selects the first $k_t$ variable nodes expanded. 
\end{itemize}
Furthermore, we compare our approaches with state-of-the-art ML approaches:
\begin{itemize}
    \item \DM: LNS which selects the neighborhood using a GCN-based policy obtained by learning to imitate the LB heuristic \cite{sonnerat2021learning}. We implement \DM since the authors do not fully open source the code;
    \item \RL: LNS which selects the neighborhood using a GCN-based policy obtained by reinforcement learning \cite{wu2021learning}. Note that this approach does not require a given neighborhood size $k_t$ since the size is defined implicitly by how the trained policy is used. We use the code made available by the authors.
\end{itemize}

\begin{figure}[tbp]
    \centering
    \includegraphics[width=\textwidth]{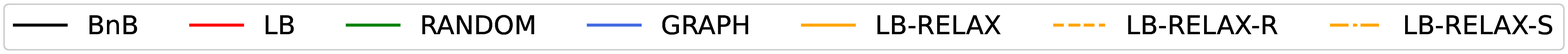}
    \begin{subfigure}[htbp]{0.49\textwidth}
		\centering
		\includegraphics[height=3.3cm]{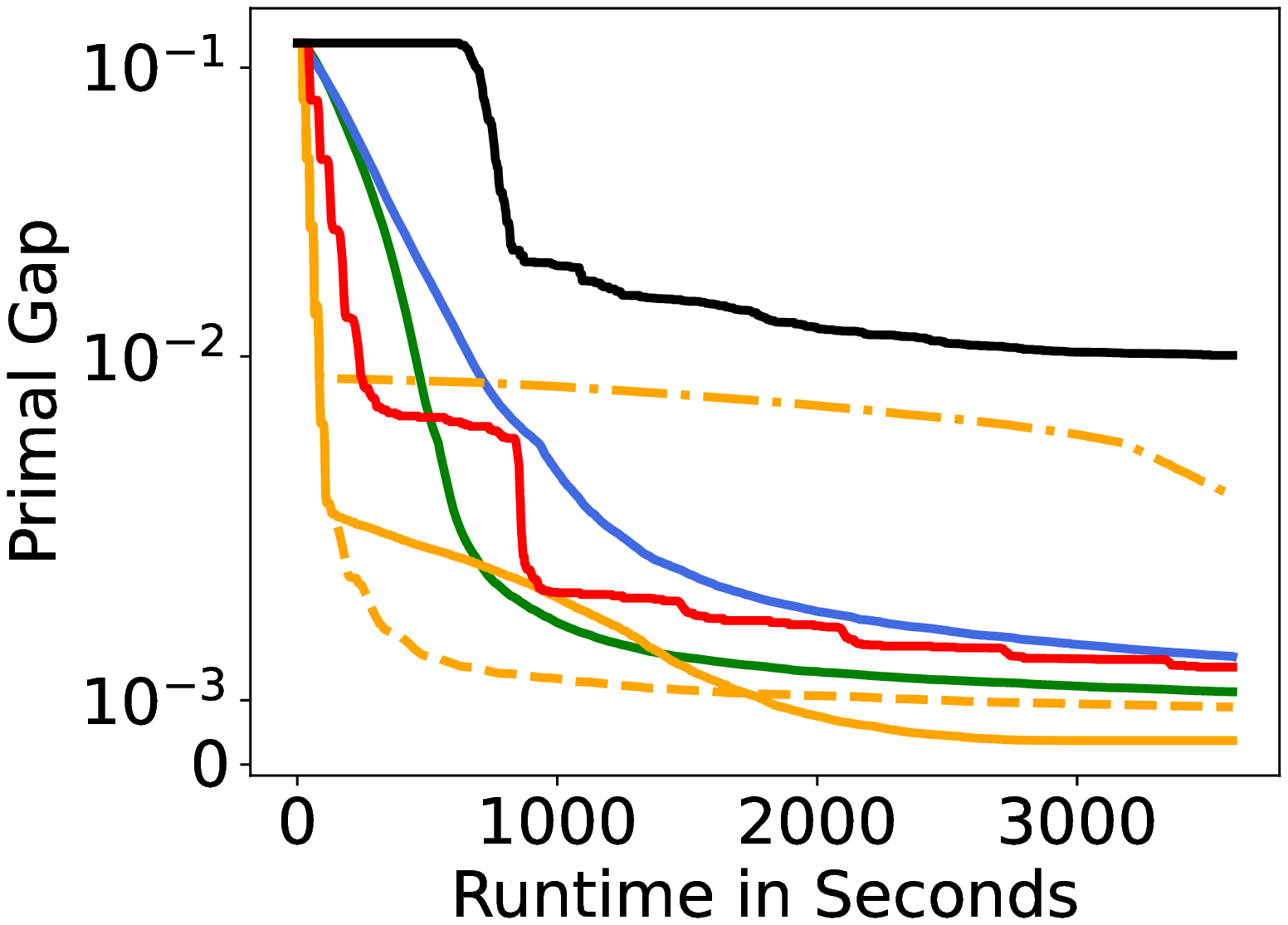}
		\caption{MVC}
	\end{subfigure}\,\,
	\begin{subfigure}[htbp]{0.49\textwidth}
		\centering
		\includegraphics[height=3.3cm]{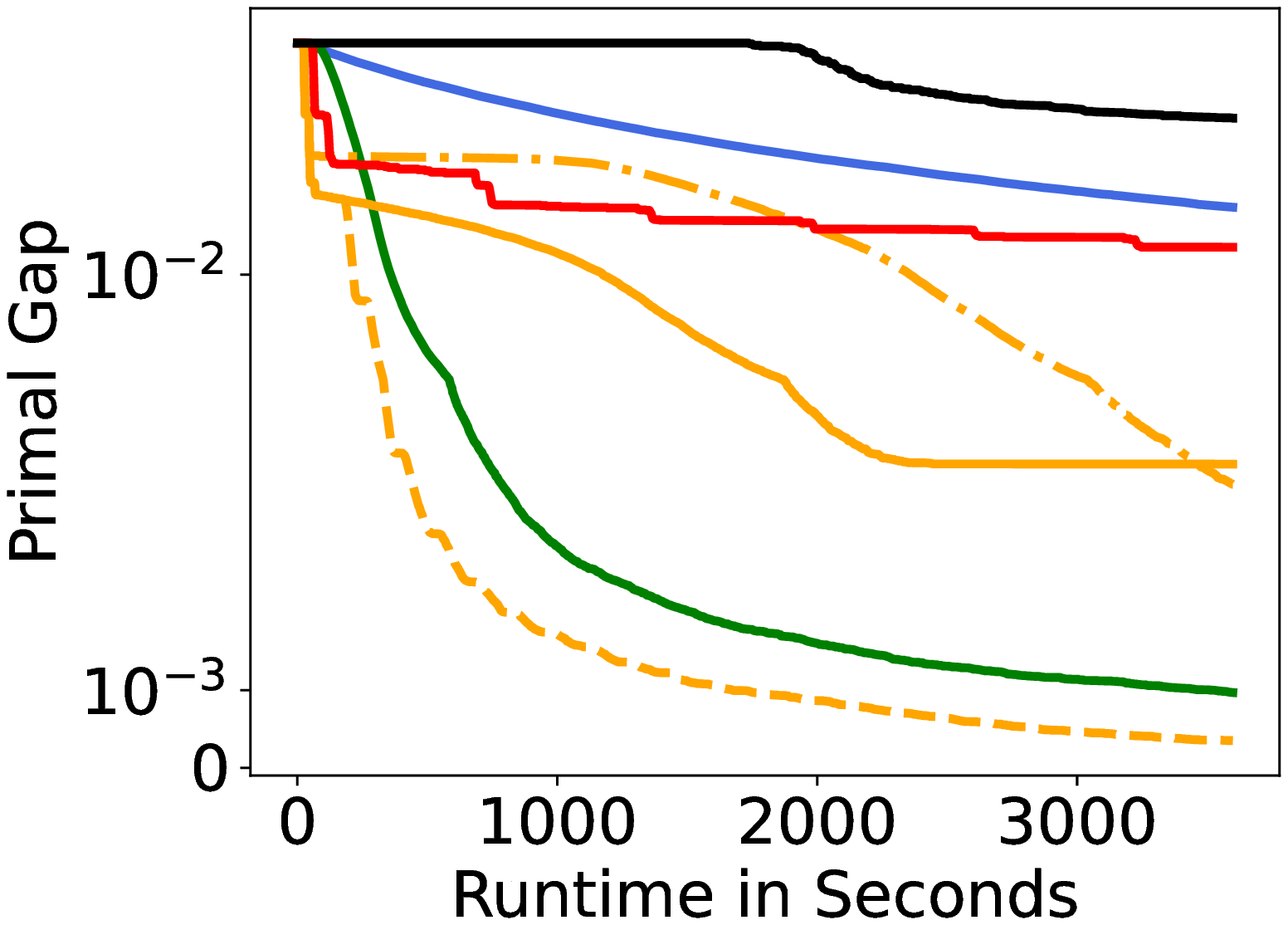}

			\caption{MIS}
	\end{subfigure}
	\begin{subfigure}[htbp]{0.49\textwidth}
		\centering
		\includegraphics[height=3.3cm]{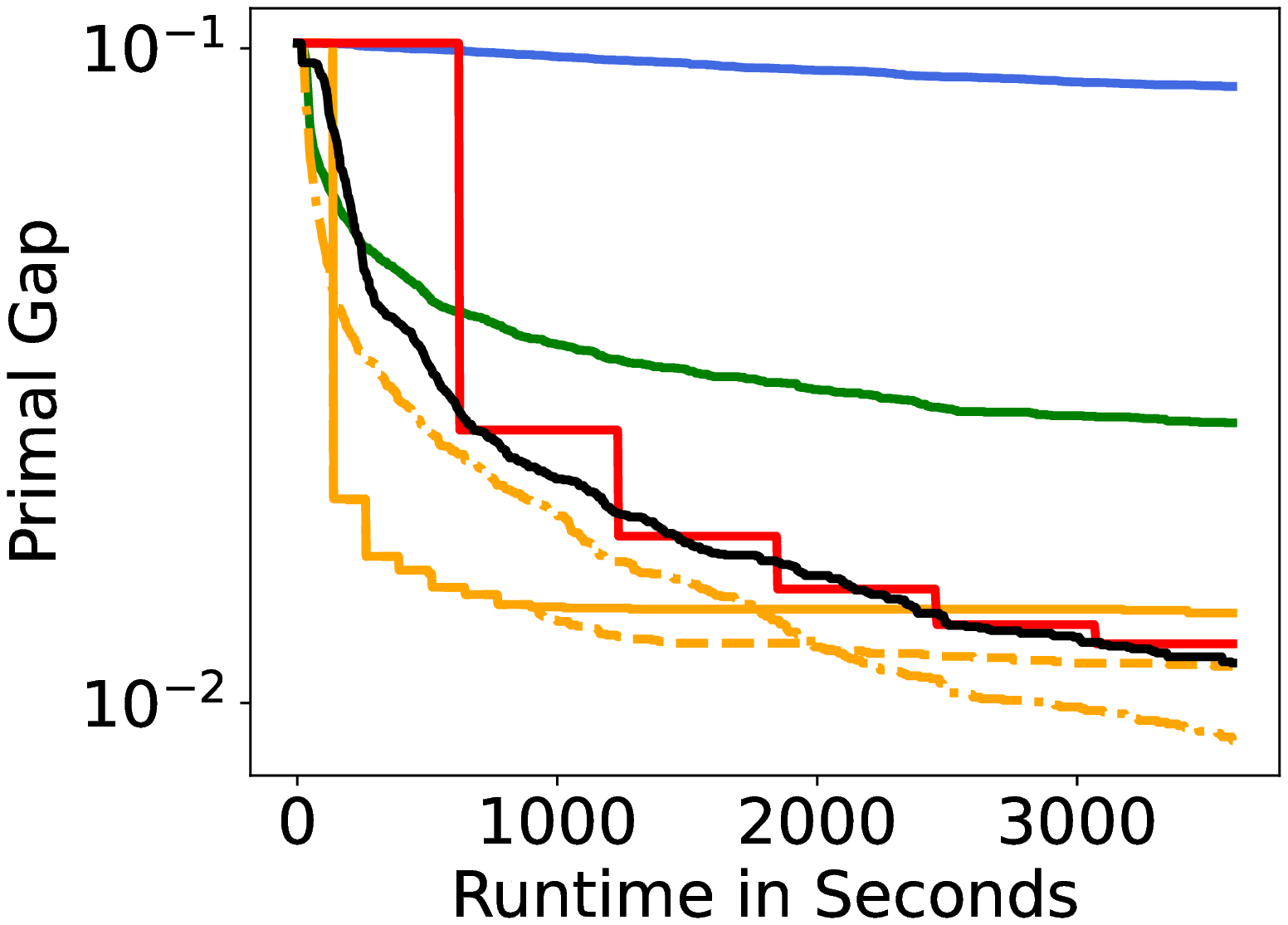}

			\caption{SC}
	\end{subfigure}
	\begin{subfigure}[htbp]{0.49\textwidth}
		\centering
		\includegraphics[height=3.3cm]{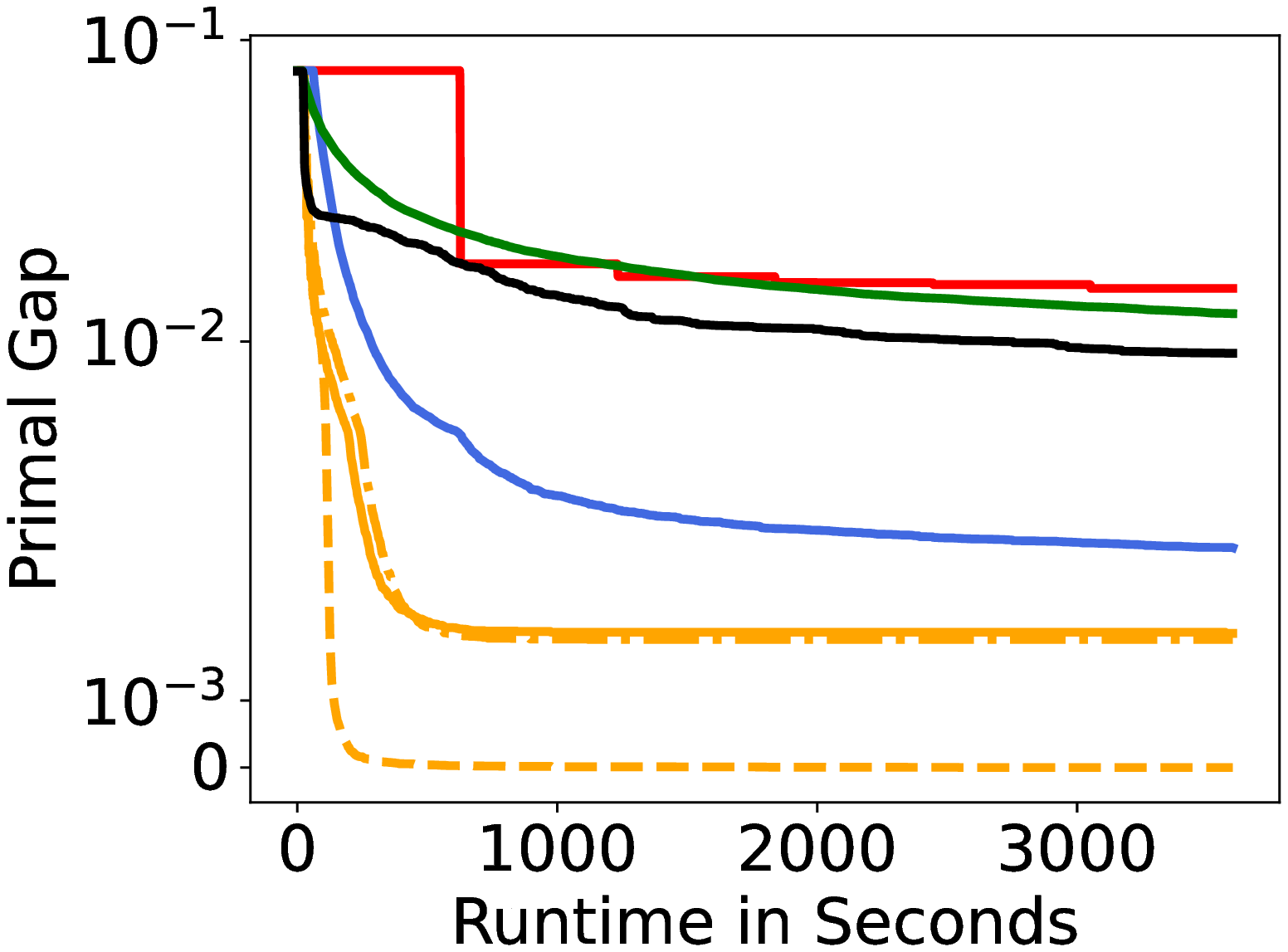}
			\caption{MK}
	\end{subfigure}
    \caption{Comparison with non-ML approaches: The primal gap as a function of time, averaged over 100 instances. \label{res::gap}}
\end{figure}

\subsubsection{Hyperparameters}
We conduct our experiments on 2.5GHz Intel Xeon Platinum 8259CL CPUs with 32 GB RAM. All experiments use the hyperparameters described below unless stated otherwise. 
We use SCIP (v8.0.1) \cite{BestuzhevaEtal2021OO}, the state-of-the-art open source ILP solver for the repair operations in LNS. 
To run LNS, we find an initial solution by running SCIP for 10 seconds for MVC, MIS and SC and 20 seconds for MK. We set the time limit to 60 minutes to solve each instance and 2 minutes for each repair operation in LNS. Except for LB, we set the time limit to 10 minutes for each repair operation since LB solves a larger ILP than other approaches in each iteration and typically requires a longer time limit.
All approaches require a neighborhood size $k_t$ in LNS, except for BnB and \RL. The initial neighborhood size ($k_0$) is set to  $k_0=400,200, 150$ and $400$ for MVC, MIS, SC and MK, respectively.  For fair comparison, all baselines use adaptive neighborhood sizes with  $\alpha=1.02$ and $\beta=0.5$, except for BnB and \RL. 
For \LBRELAXRR, we set $\gamma=30$ seconds. Additional details on tuning hyperparameters are included in Appendix. 

\subsubsection{Metrics}
We use the following metrics to evaluate the efficiency and effectiveness of different approaches:
(1) The \textit{primal bound} is the objective value of the ILP.
(2) The \textit{primal gap} \cite{berthold2006primal} is the normalized difference between the primal bound $v$ and a precomputed best known objective value $v^*$, defined as $\frac{|v-v^*|}{\max(v,v^*,\epsilon)}$ if $v$ exists and $v\cdot v^*\geq 0$, or 1 otherwise. We use $\epsilon=10^{-8}$ to avoid division by zero and $v^*$ is the best primal bound found within 60 minutes by any approach in the portfolio for comparison. 
(3)The \textit{primal integral} \cite{achterberg2012rounding} at time $q$  is the integral on $[0,q]$ of the primal gap as a function of time. It captures the quality of and the speed at which solutions are found. 
(4) The \textit{survival rate}  to meet a certain primal gap threshold is the fraction of instances with the primal gap below the threshold \cite{sonnerat2021learning}. 
Since BnB and LNS are both anytime algorithms, we  show the metrics as a function of time or the number of iterations in LNS (when applicable) to demonstrate their anytime performance.

\begin{figure}[tbp]
    \centering
    \includegraphics[width=\textwidth]{figure/legend_timeVSobj.eps}
    \begin{subfigure}[htbp]{0.49\textwidth}
		\centering
		\includegraphics[height=3.3cm]{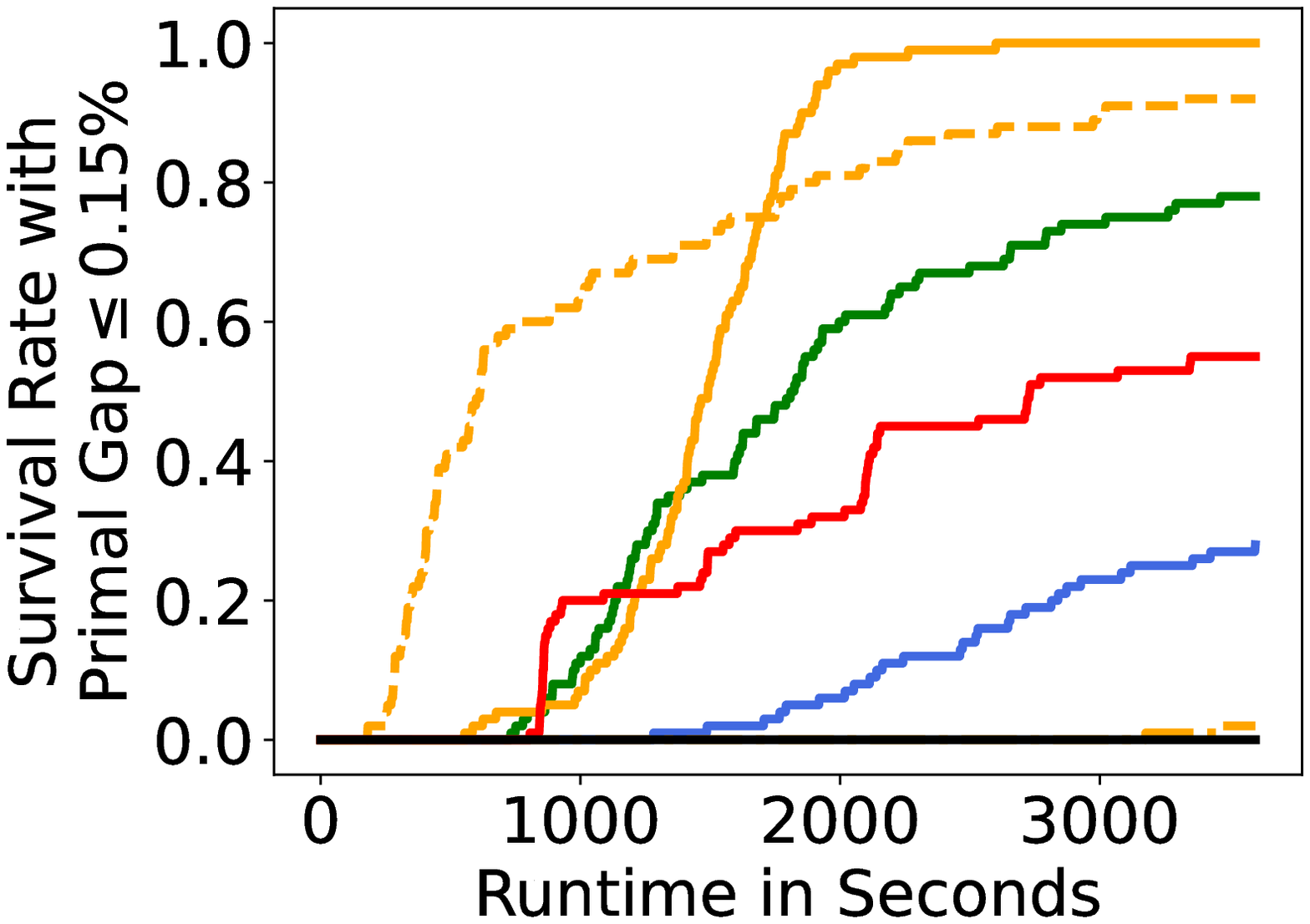}
		\caption{MVC}
	\end{subfigure}\,\,
	\begin{subfigure}[htbp]{0.49\textwidth}
		\centering
		\includegraphics[height=3.3cm]{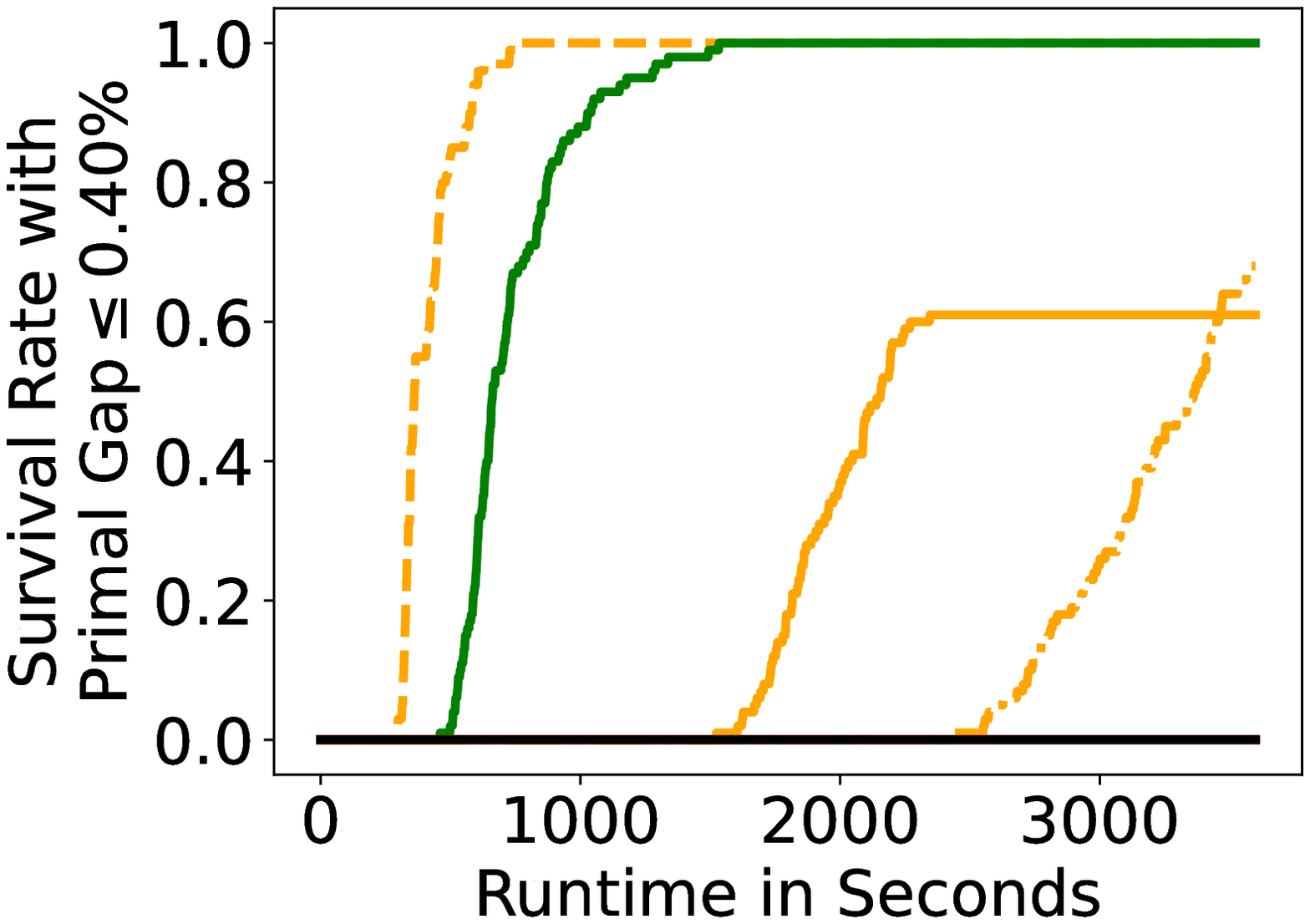}

			\caption{MIS}
	\end{subfigure}
	\begin{subfigure}[htbp]{0.49\textwidth}
		\centering
		\includegraphics[height=3.3cm]{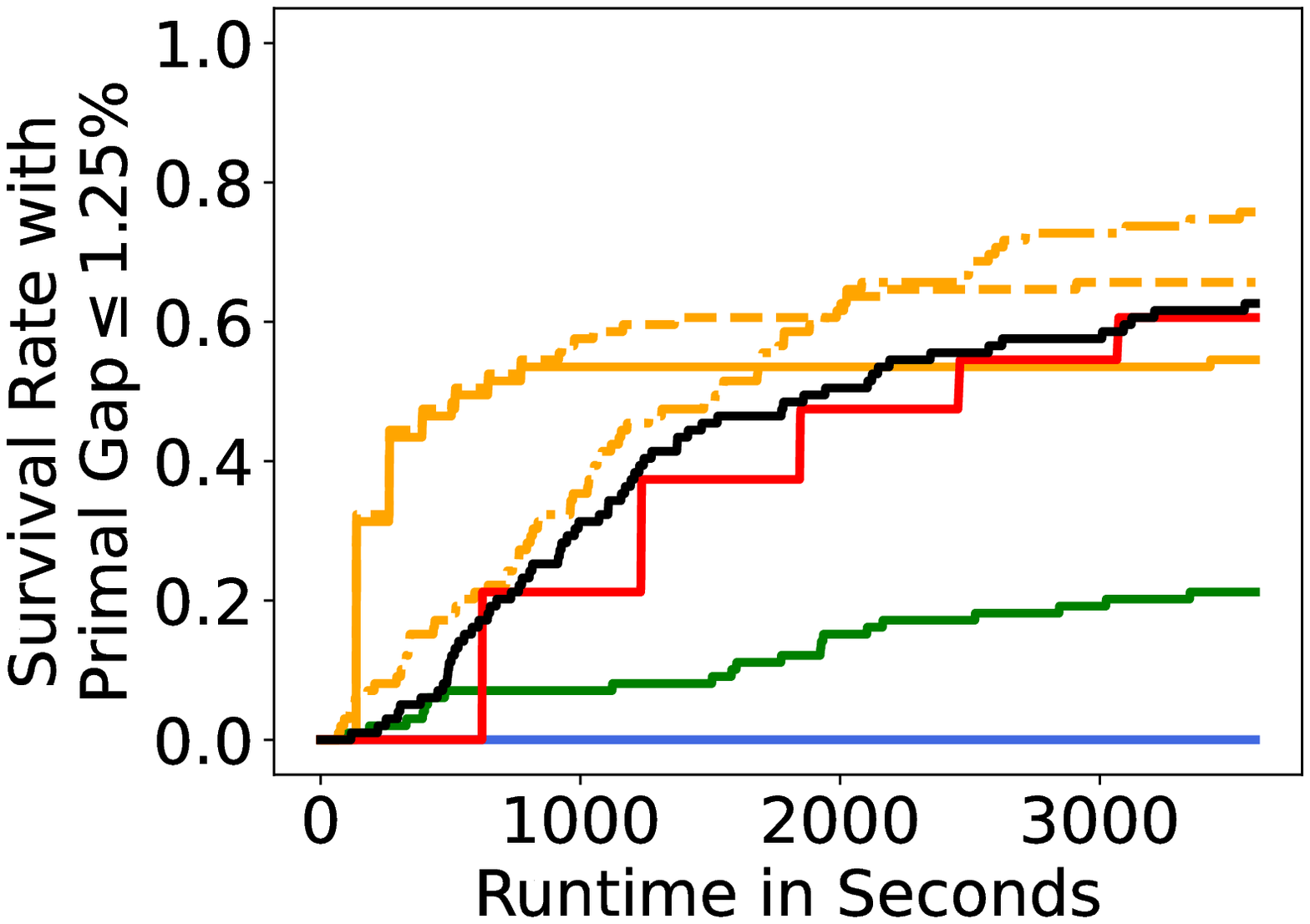}

			\caption{SC}
	\end{subfigure}
	\begin{subfigure}[htbp]{0.49\textwidth}
		\centering
		\includegraphics[height=3.3cm]{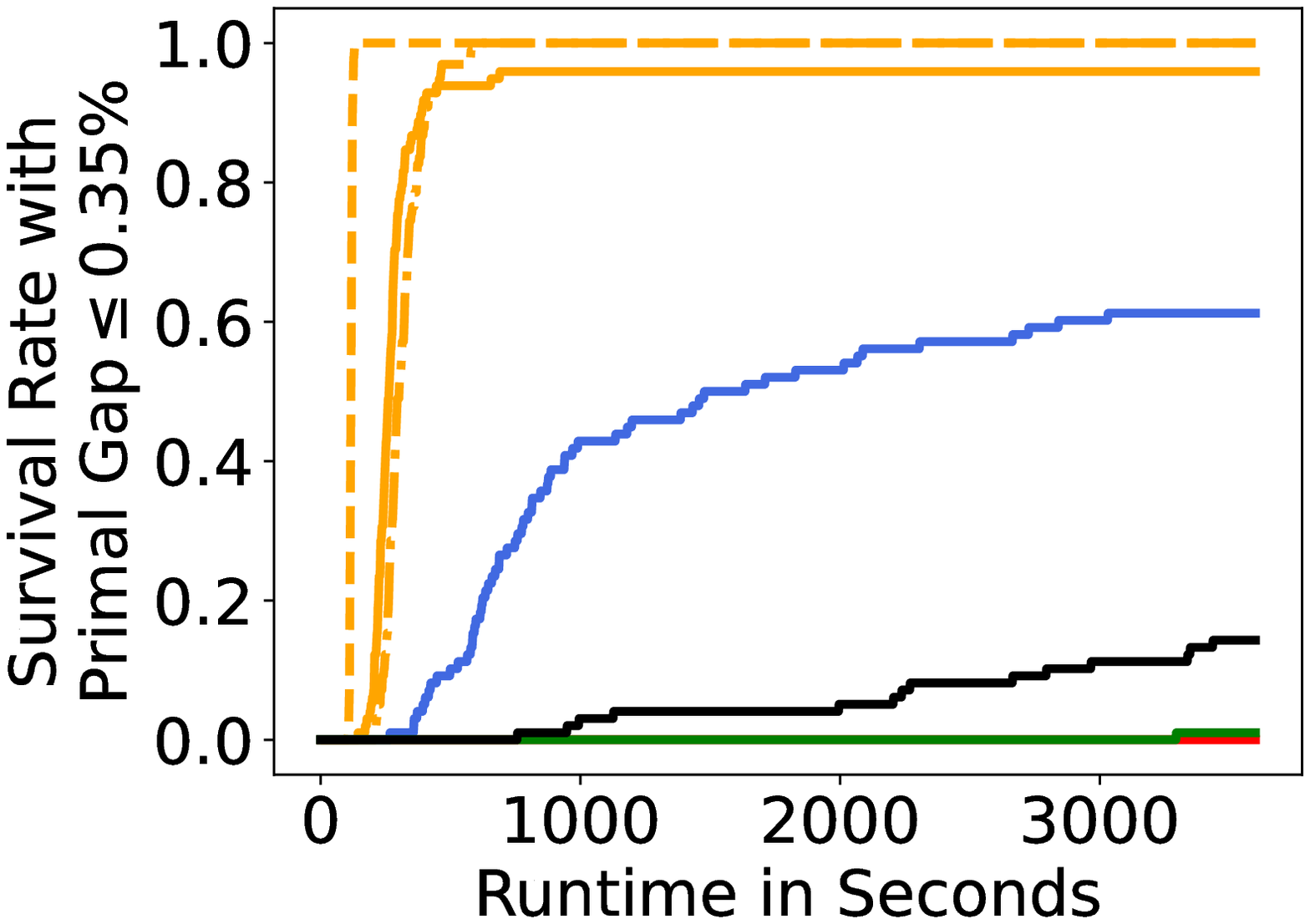}
			\caption{MK}
	\end{subfigure}
    \caption{Comparison with non-ML approaches: The survival rate over 100 instances as a function of time to meet a certain primal gap threshold. The primal gap threshold is chosen from Table \ref{res::nonMLtable} as the median of the average primal gaps at 60 minutes time cutoff over all approaches rounded to the nearest 0.05\%. \label{res::survivalCurve}}
\end{figure}

\subsection{Results}


\subsubsection{Comparison with Non-ML Approaches}

First, we compare \LBRELAX, \LBRELAXRR and \LBRELAXS with non-ML approaches, namely BnB, LB, \RANDOM and \GRAPH.
Figure \ref{res::gap} shows the primal gap as a function of time, averaged over 100 instances. The results show that \LBRELAX, \LBRELAXRR and \LBRELAXS consistently improve the primal gap a lot faster than the baselines in the first few minutes of LNS. \LBRELAX improves the primal gap slightly faster than \LBRELAXS in all cases. 
On average, \LBRELAX is always better than the baselines at any point of time on MK instances and \LBRELAXS is always better than the baselines on SC and MK instances. However, both \LBRELAX and \LBRELAXS could get stuck at some local minima. In those cases, they need some time to escape local minima by adjusting the neighborhood size and sometimes could be outperformed by some baselines with longer time on the MVC and MIS instances. By adding randomization to \LBRELAX, \LBRELAXRR escapes local minima more efficiently than \LBRELAX and \LBRELAXS. On average, \LBRELAXRR is always better than the baselines at any point of time in the search on the MVC, MIS and MK instances. 

\begin{table}[tbp]
\scriptsize
\centering
\caption{\small Primal gap (PG) (in percent) and primal integral (PI) at 60 minutes time cutoff, averaged over 100 instances, and their standard deviations. \label{res::nonMLtable}}

\begin{tabular}{c|rr|rr}
\hline
                       & \multicolumn{2}{c|}{MVC}                                                                        & \multicolumn{2}{c}{MIS}                                                                         \\ \hline
                                    & \multicolumn{1}{c|}{PG (\%)}         & \multicolumn{1}{c|}{PI}              & \multicolumn{1}{c|}{PG (\%)}         & \multicolumn{1}{c}{PI} \\ \hline
BnB                    & \multicolumn{1}{r|}{1.01$\pm$0.46} & 128.6$\pm$14.6           & \multicolumn{1}{r|}{2.80$\pm$1.36} & 144.0$\pm$20.1           \\ \hline
LB                     &  \multicolumn{1}{r|}{0.15$\pm$0.08} & 22.1$\pm$3.6             & \multicolumn{1}{r|}{1.20$\pm$0.31} & 56.3$\pm$9.4             \\ \hline
\RANDOM                 &  \multicolumn{1}{r|}{0.11$\pm$0.05} & 32.3$\pm$2.3             & \multicolumn{1}{r|}{0.10$\pm$0.05} & 18.0$\pm$2.5             \\ \hline
\GRAPH                  &  \multicolumn{1}{r|}{0.17$\pm$0.04} & 40.8$\pm$2.5             & \multicolumn{1}{r|}{1.56$\pm$0.18} & 90.2$\pm$7.6             \\ \hline
\LBRELAX                & \multicolumn{1}{r|}{\bf0.04$\pm$0.03} & 10.3$\pm$1.7              & \multicolumn{1}{r|}{0.39$\pm$0.12} & 29.4$\pm$4.3             \\ \hline
\LBRELAXRR              & \multicolumn{1}{r|}{0.09$\pm$0.04} & {\bf9.6$\pm$1.7}              & \multicolumn{1}{r|}{\bf 0.04$\pm$0.04} & {\bf 9.3$\pm$1.7}             \\ \hline
\LBRELAXS               & \multicolumn{1}{r|}{0.42$\pm$0.20} & 28.8$\pm$8.1             &  \multicolumn{1}{r|}{0.37$\pm$0.11} & 51.7$\pm$10.1            \\ \hline
\multicolumn{1}{l|}{} & \multicolumn{2}{c|}{SC}                                                                         & \multicolumn{2}{c}{MK}                                                                          \\ \hline
BnB                     & \multicolumn{1}{r|}{1.15$\pm$0.98} & 87.4$\pm$38.6            & \multicolumn{1}{r|}{0.91$\pm$0.59} & 60.7$\pm$17.9            \\ \hline
LB                      & \multicolumn{1}{r|}{1.23$\pm$0.98} & 114.1$\pm$35.7            &  \multicolumn{1}{r|}{1.50$\pm$0.48} & 97.7$\pm$13.0            \\ \hline
\RANDOM                 & \multicolumn{1}{r|}{2.68$\pm$1.31} & 124.4$\pm$45.7           & \multicolumn{1}{r|}{1.24$\pm$0.36} & 68.9$\pm$14.7            \\ \hline
\GRAPH                  &\multicolumn{1}{r|}{8.75$\pm$2.15} & 338.2$\pm$77.0           &  \multicolumn{1}{r|}{0.33$\pm$0.14} & 23.6$\pm$4.9             \\ \hline
\LBRELAX                & \multicolumn{1}{r|}{1.37$\pm$0.96} & {63.9$\pm$34.0}            & \multicolumn{1}{r|}{0.20$\pm$0.09} & 11.3$\pm$3.0             \\ \hline
\LBRELAXRR              & \multicolumn{1}{r|}{1.14$\pm$0.90} & {\bf58.9$\pm$31.5}            &  \multicolumn{1}{r|}{\bf0.00$\pm$0.00} & {\bf3.7$\pm$0.4}              \\ \hline
\LBRELAXS               & \multicolumn{1}{r|}{\bf0.88$\pm$0.85} & 63.8$\pm$32.4            & \multicolumn{1}{r|}{0.19$\pm$0.07} & 11.8$\pm$2.4             \\ \hline
\end{tabular}
\end{table}

\begin{table}[htbp]

\centering
\caption{\small The time (in seconds) to improve the inital solution in one iteration and the improvement of the primal bound, averaged over 100 instances. The time for LB is the solving time of the LB ILP. The time for \LBRELAX and \LBRELAXS is the sum of the solving times of the LB relaxation and the sub-ILP. The numbers in parentheses are the speed-ups. The improvement is computed by taking the difference between the initial solution and the new incumbent solution and the numbers in parentheses are the losses in quality in percent compared to LB. $\uparrow$ means higher is better, $\downarrow$ means lower is better. \label{res::firstIter}}

\scriptsize
\begin{tabular}{c|c|r|r|r|rllll}
\cline{1-6}
                          &      & \multicolumn{1}{c|}{MVC} & \multicolumn{1}{c|}{MIS} & \multicolumn{1}{c|}{SC} & \multicolumn{1}{c}{MK} &  &  &  &  \\ \cline{1-6}
\multirow{2}{*}{LB}       & Time$\downarrow$ & 40.2                     & 56.0                     & 600.0                   & 600.0                  &  &  &  &  \\ \cline{2-6}
                          & Imp.$\uparrow$  & 129.79                   & 65.50                    & 12.21                   & 216.51                 &  &  &  &  \\ \cline{1-6}
\multirow{2}{*}{\LBRELAX}  & Time$\downarrow$ & 12.1 (3.3x)           & 19.5 (2.9x)           & 125.3 (4.8x)         & 5.87 (102.2x)         &  &  &  &  \\ \cline{2-6}
                          & Imp.$\uparrow$  & 129.41 (-0.3\%)          & 65.19 (-0.5\%)           & 15.77 (+29.2\%)         & 141.10 (-34.8\%)       &  &  &  &  \\ \cline{1-6}
\multirow{2}{*}{\LBRELAXS} & Time$\downarrow$ & 12.0 (3.4x)           & 19.5 (2.9x)           & 24.51 (24.5x)         & 5.12 (117.6x)         &  &  &  &  \\ \cline{2-6}
                          & Imp.$\uparrow$  & 128.61 (-0.9\%)          & 62.46 (-4.6\%)           & 5.65 (-53.7\%)          & 113.48 (-47.6\%)       &  &  &  &  \\ \cline{1-6}
\end{tabular}

\end{table}

Table \ref{res::nonMLtable} presents the average primal gap and primal integral at 60 minutes time cutoff. 
(See results at 15, 30 and 45 minutes time cutoff in Appendix.) 
On MVC, SC and MK instances, all \LBRELAX, \LBRELAXS and \LBRELAXRR have lower primal gaps and primal integrals on average than any baselines, demonstrating that they not only find higher quality solutions but also find them at a faster speed. On MIS and MK instances, \LBRELAXRR achieves the lowest primal gap and primal integral among all approaches. It also achieves the lowest primal integral on MVC and SC instances. Overall, \LBRELAXRR always comes up in the top 2 in both metrics on all problems. 

Figure \ref{res::survivalCurve} shows the survival rate over 100 instances as a function of time to meet a certain primal gap threshold. On MVC instances, \LBRELAX and \LBRELAXRR achieve final survival rates above 0.9 while the best baseline \RANDOM stays below 0.8. On MIS instances, both \LBRELAXRR and \RANDOM achieve final survival rates of 1.0 but \LBRELAXRR uses shorter time. On SC instances, \LBRELAXS and \LBRELAXRR consistently has a higher survival rate than the baselines. On MK instances, \LBRELAX and its variants achieve survival rates above 0.9 within 15 minutes while the best baseline \GRAPH only gets to around 0.6 with 60 minutes.

One limitation of \LBRELAX and its variants is that they do not perform well on some problem domains, for example the maximum cut and combinatorial auction problems. Please see Appendix for more results.

\begin{figure}[tbp]
    \centering
    \includegraphics[width=0.45\textwidth]{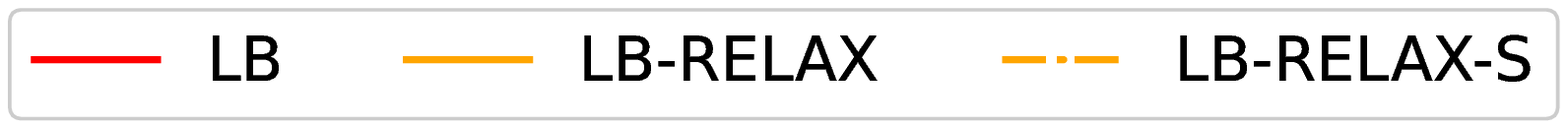}\\
    \begin{subfigure}[htbp]{0.49\textwidth}
		\centering
		\includegraphics[height=3.3cm]{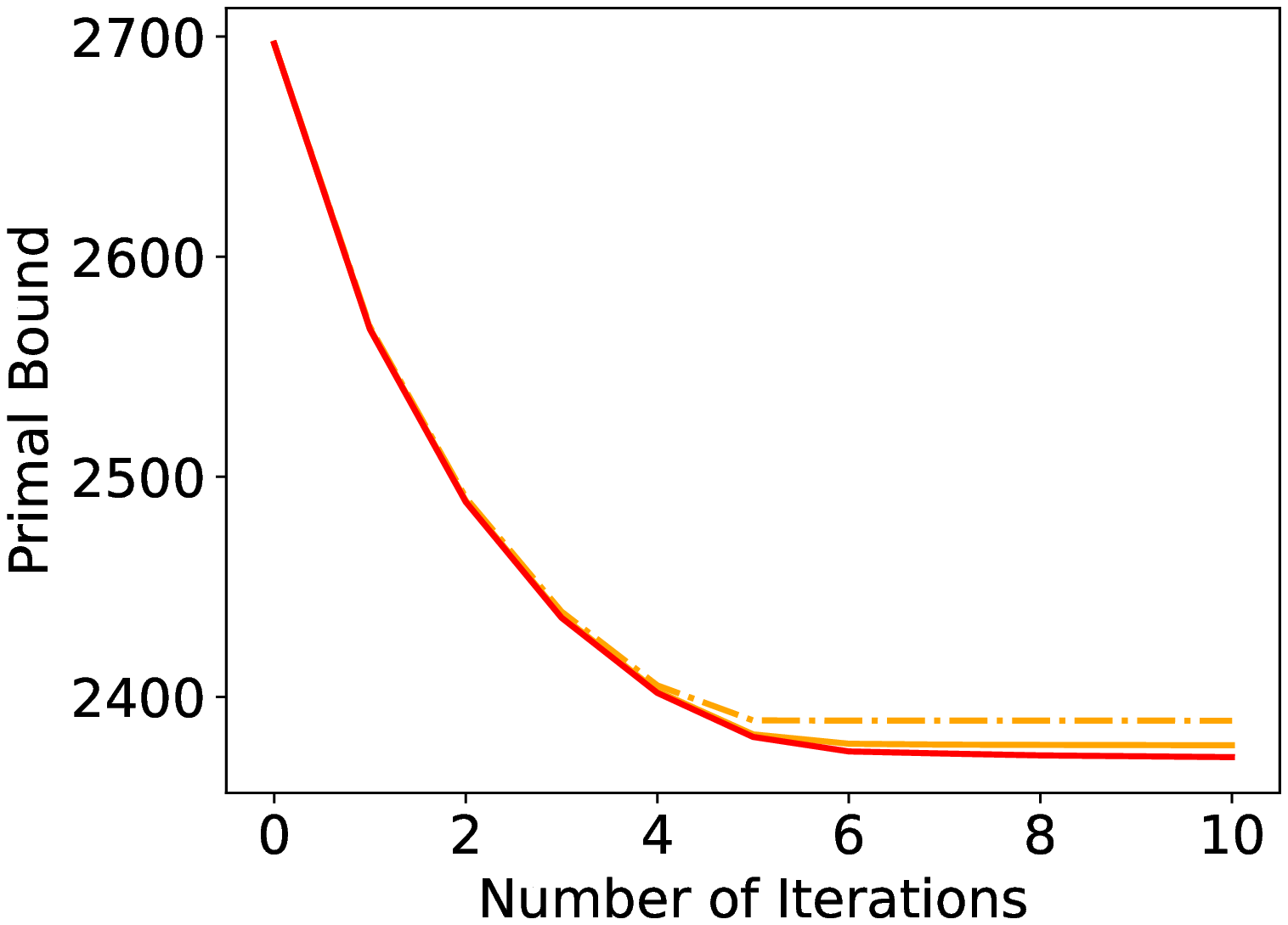}
		\caption{MVC}
	\end{subfigure}\,\,
	\begin{subfigure}[htbp]{0.49\textwidth}
		\centering
		\includegraphics[height=3.3cm]{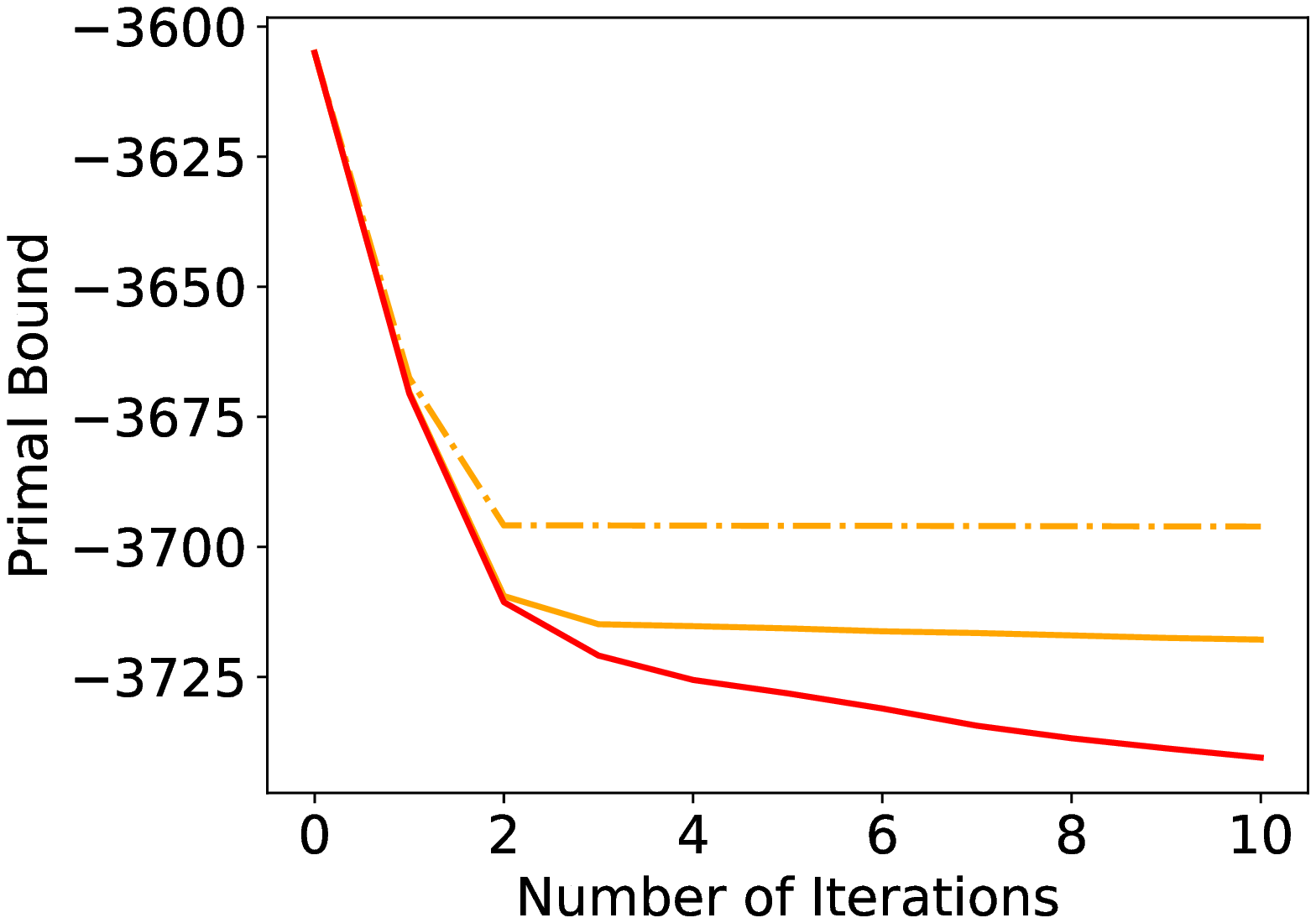}
			\caption{MIS}
	\end{subfigure}
	\begin{subfigure}[htbp]{0.49\textwidth}
		\centering
		\includegraphics[height=3.3cm]{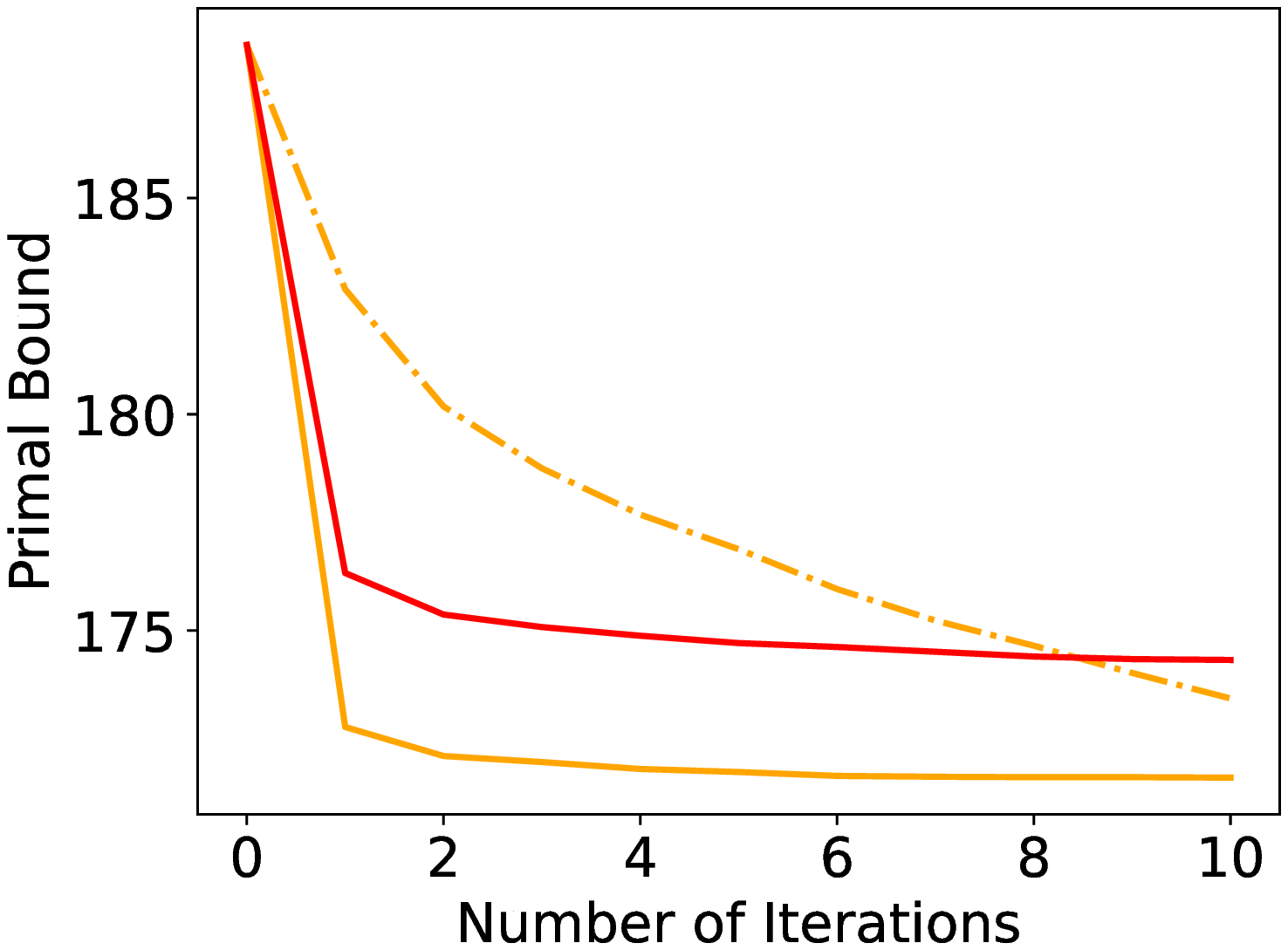}
			\caption{SC}
	\end{subfigure}
	\begin{subfigure}[htbp]{0.49\textwidth}
		\centering
		\includegraphics[height=3.3cm]{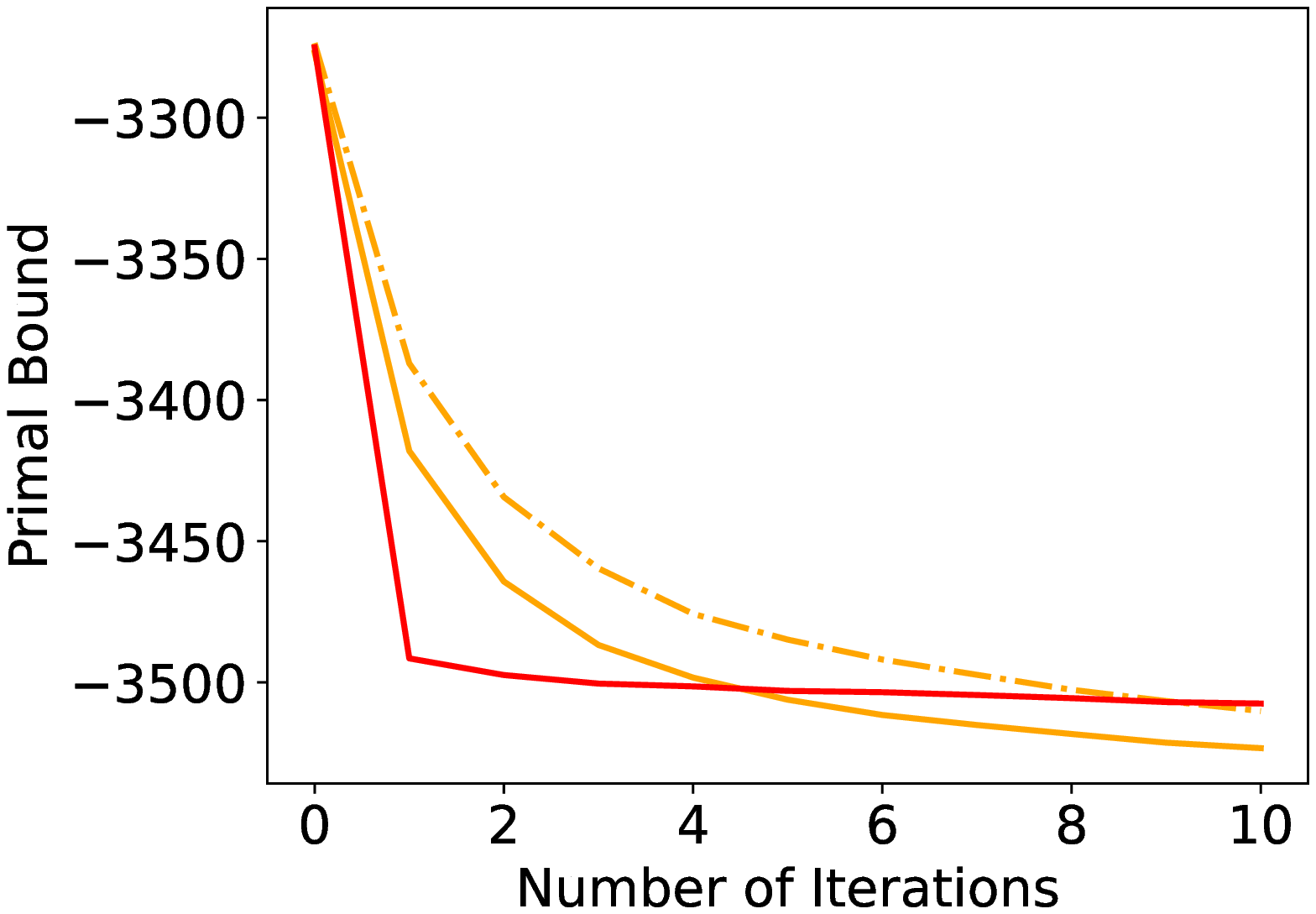}
			\caption{MK}
	\end{subfigure}
    \caption{Comparison with LB: The primal bound as a function of the number of iterations, averaged over 100 instances. \label{res::PBperIter}}
\end{figure}

Next, we run LB, \LBRELAX and \LBRELAXS for 10 iterations to compare their effectiveness. We follow the same setup as earlier described, except that we do not use adaptive neighborhood sizes to make sure they have the same $k_t$ in each iteration $t$.  Note that the time limit for solving the sub-ILP in each iteration is set to 10 minutes for LB and 2 minutes for \LBRELAX and \LBRELAXS.
Table \ref{res::firstIter} shows the average time to improve the initial solutions and the average improvement of the primal bound in the first iteration of LNS. This allows us to compare how closely \LBRELAX and \LBRELAXS approximate the quality of the neighborhood selected by LB and study the trade-off between quality and time. Compared to LB, \LBRELAX and \LBRELAXS have 2.9x-117.6x speed-up but only lose at most 53.7\% in quality. In particular, on MVC and MIS instances, both \LBRELAX and \LBRELAXS lose 0.5\% to 4.6\% in quality but have at least 2.9x speed-up; on SC instances, \LBRELAX even gains 29.2\%  in quality and save 79.1\% in time, due to LB cannot find a good enough neighborhood within its time limit.    

In Figure \ref{res::PBperIter}, we show the primal bound as a function of the number of iterations. 
It allows comparing the effectiveness of different heuristics independently of their speed.
On the MVC instances,  both \LBRELAX and \LBRELAXS perform similarly to but slightly worse than LB. On the SC and MK instances, \LBRELAX achieves better performance than LB, again due to scalability issues of LB, and \LBRELAXS achieves competitive performance with LB after 10 iterations. However on the MIS instances, both \LBRELAX and \LBRELAXS are able to quickly improve the primal bound in the first 2-3 iterations, but afterwards converge to local minima  and the gaps between them and LB increase. To complete the first 10 iterations, both \LBRELAX and \LBRELAXS take less than 21 minutes on SC instances and 3.3 minutes on the others, while LB takes at least 57 minutes and sometimes up to 100 minutes. 


\begin{figure}[tbp]
    \centering
    \includegraphics[width=0.75\textwidth]{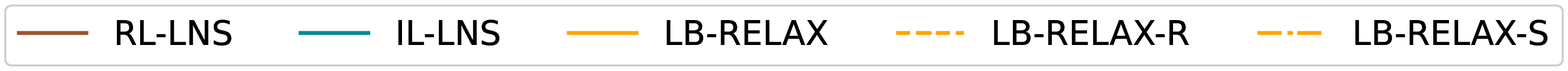}
    \begin{subfigure}[htbp]{0.325\textwidth}
		\centering
		\includegraphics[height=2.8cm]{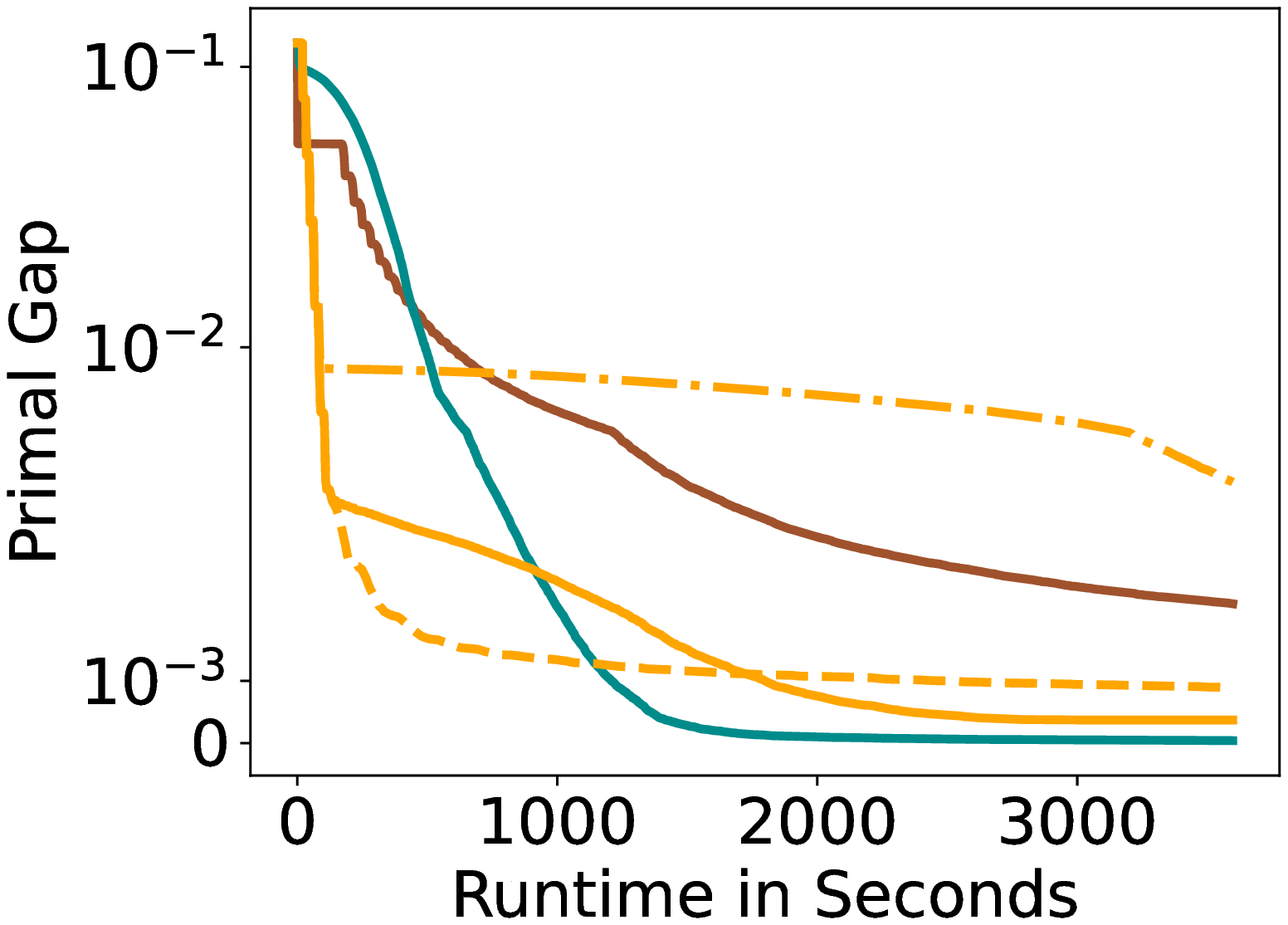}
		\caption{MVC}
	\end{subfigure}
	\begin{subfigure}[htbp]{0.325\textwidth}
		\centering
		\includegraphics[height=2.8cm]{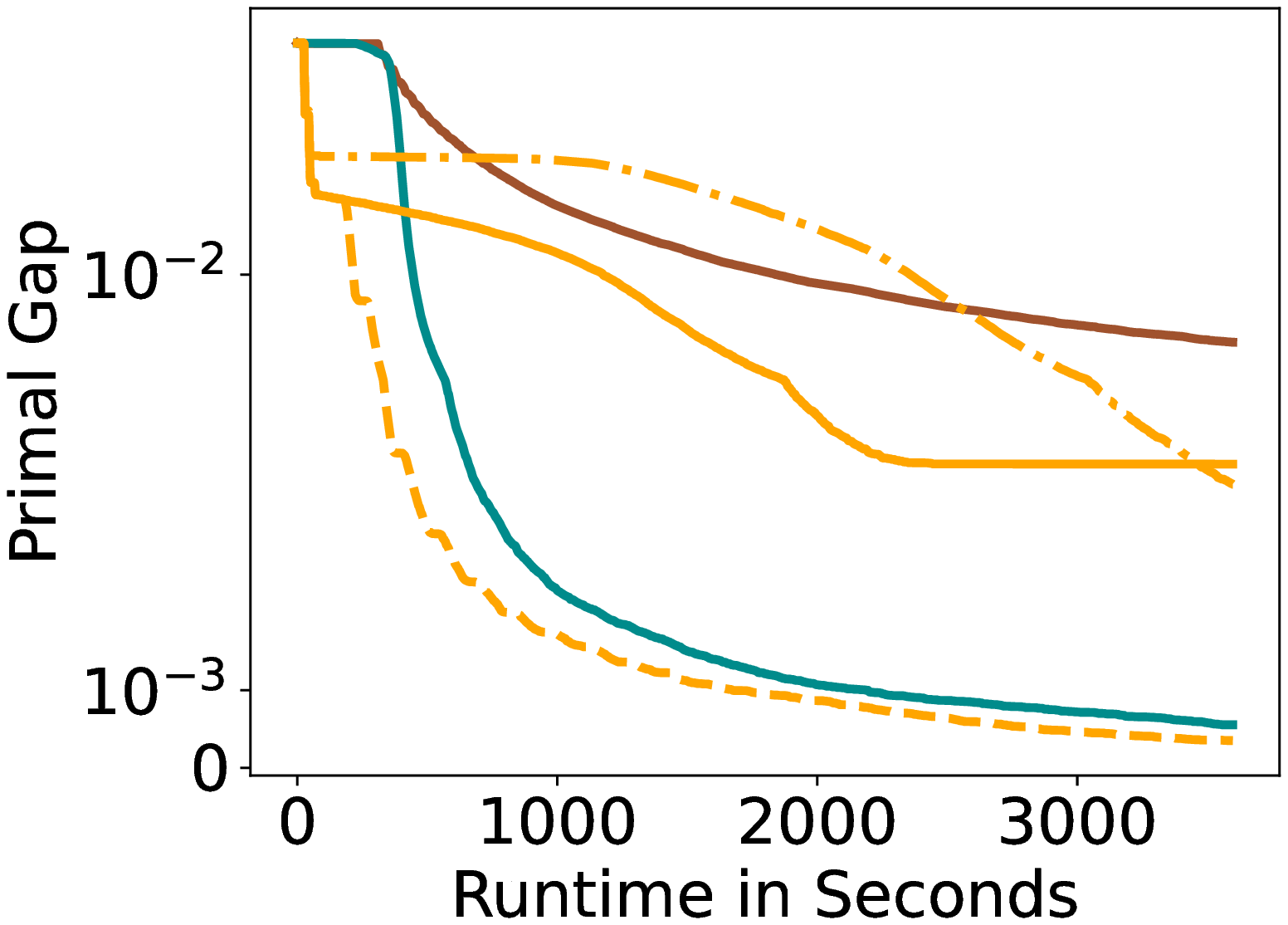}
			\caption{MIS}
	\end{subfigure}
	\begin{subfigure}[htbp]{0.325\textwidth}
		\centering
		\includegraphics[height=2.8cm]{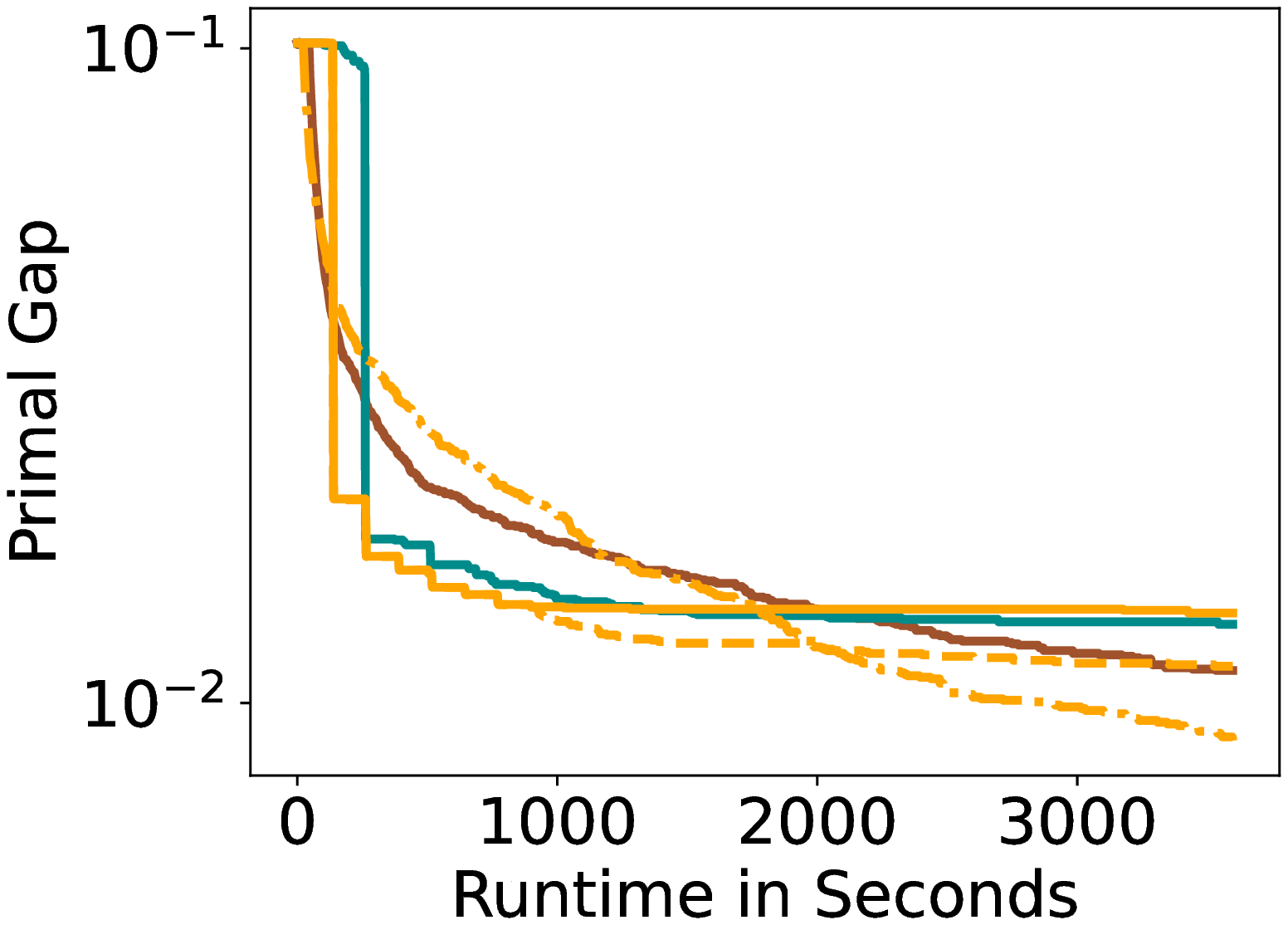}
			\caption{SC}
	\end{subfigure}

    \caption{Comparison with ML approaches: The primal bound as a function of time, averaged over 100 instances.\label{res::MLgap}}
\end{figure}

\begin{figure}[tbp]
    \centering
    \includegraphics[width=0.75\textwidth]{figure/legend_ML_timeVSobj.eps}
    \begin{subfigure}[htbp]{0.325\textwidth}
		\centering
		\includegraphics[height=2.8cm]{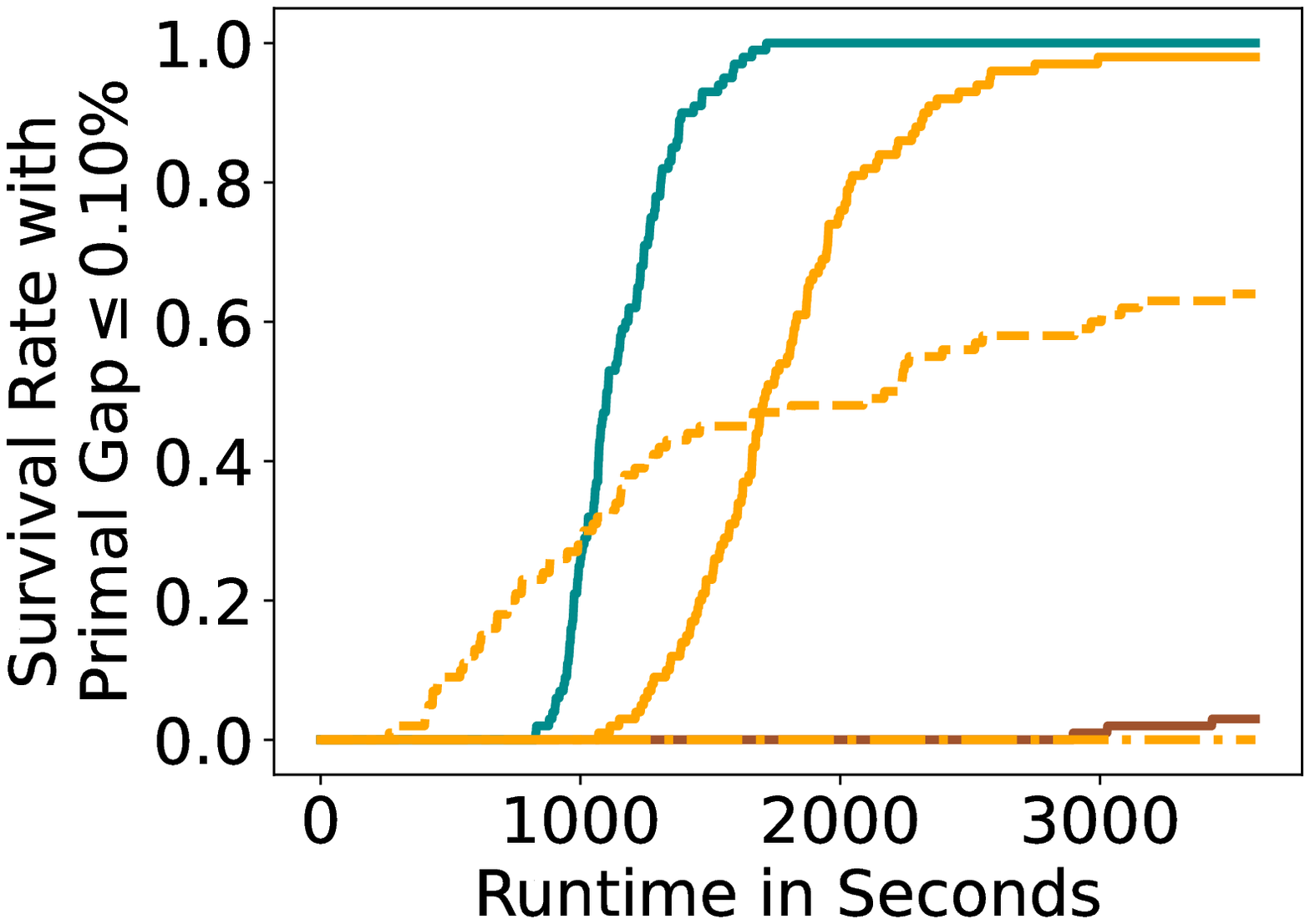}
		\caption{MVC}
	\end{subfigure}
	\begin{subfigure}[htbp]{0.325\textwidth}
		\centering
		\includegraphics[height=2.8cm]{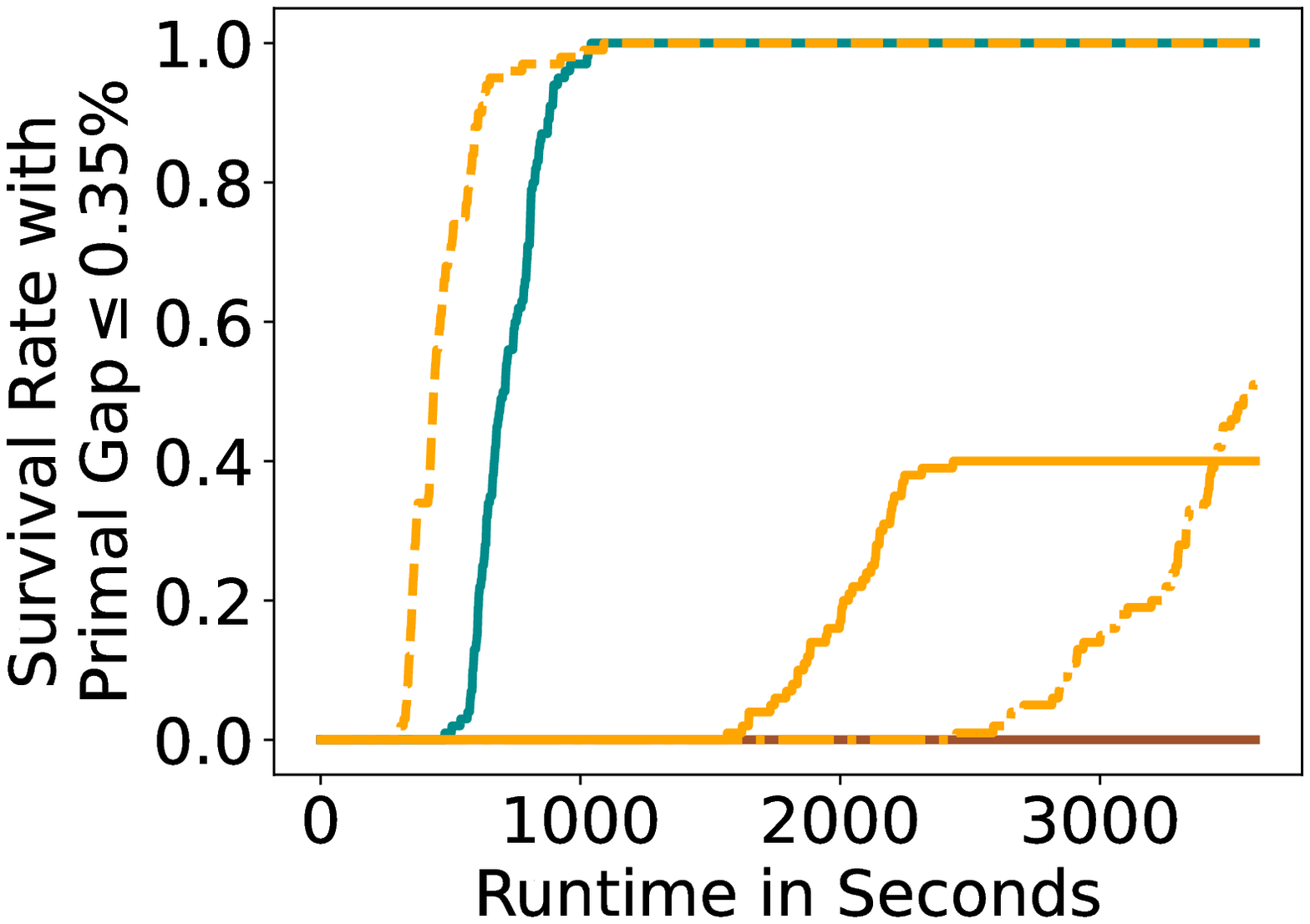}
			\caption{MIS}
	\end{subfigure}
	\begin{subfigure}[htbp]{0.325\textwidth}
		\centering
		\includegraphics[height=2.8cm]{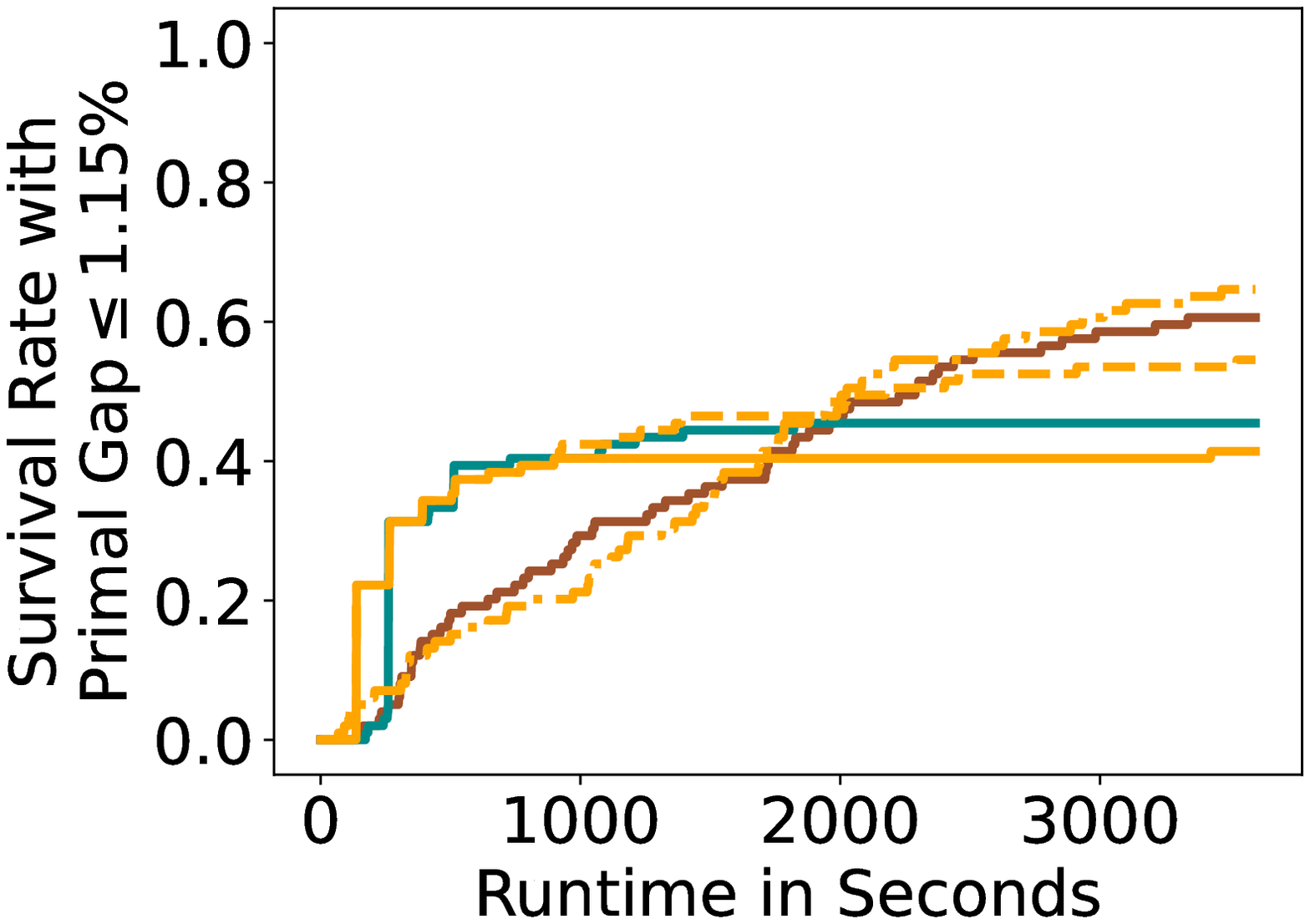}
			\caption{SC}
	\end{subfigure}

    \caption{Comparison with ML approaches: The survival rate over 100 instances as a function of time to meet a certain primal gap threshold. The primal gap thresholds are chosen in the same way as Fig. \ref{res::survivalCurve}. \label{res::MLsurvivalCurve}}
\end{figure}

\subsubsection{Comparison with ML Approaches}

Then, we compare \LBRELAX, \LBRELAXRR and \LBRELAXS on MVC, MIS and SC instances with ML approaches, namely \DM and \RL. 
Figure \ref{res::MLgap} shows the primal gap as a function of time averaged over 100 instances.
The results show that \LBRELAX, \LBRELAXRR and \LBRELAXS consistently improve the primal bound a lot faster than \DM and \RL in the first few minutes of LNS. 
On MVC instances, \DM surpasses \LBRELAXRR with the smallest average primal gap best after 20 minutes and achieve (close-to-)zero gaps after 30 minutes. 
On MIS instances, \LBRELAXRR has a smaller gap than both \DM and \RL throughout the first 60 minutes. On SC instances, \DM is very competitive with \LBRELAX and converges to a similar but slightly higher gap than \LBRELAXRR and \LBRELAXS;
\RL converges to almost the same primal gap as \LBRELAXRR on average but is worse than the best performer \LBRELAXS.
Overall, \LBRELAX and its variants, that do not require extra computational resources for training, are competitive with and more often even better than state-of-the-art ML approaches, suggesting that they are agnostic to the distributions of the instances and easily applicable to different problem domains.

\begin{figure}[tbp]
    \centering
    
		\centering
		    \includegraphics[width=\textwidth]{figure/legend_timeVSobj.eps}
		\includegraphics[height=3.2cm]{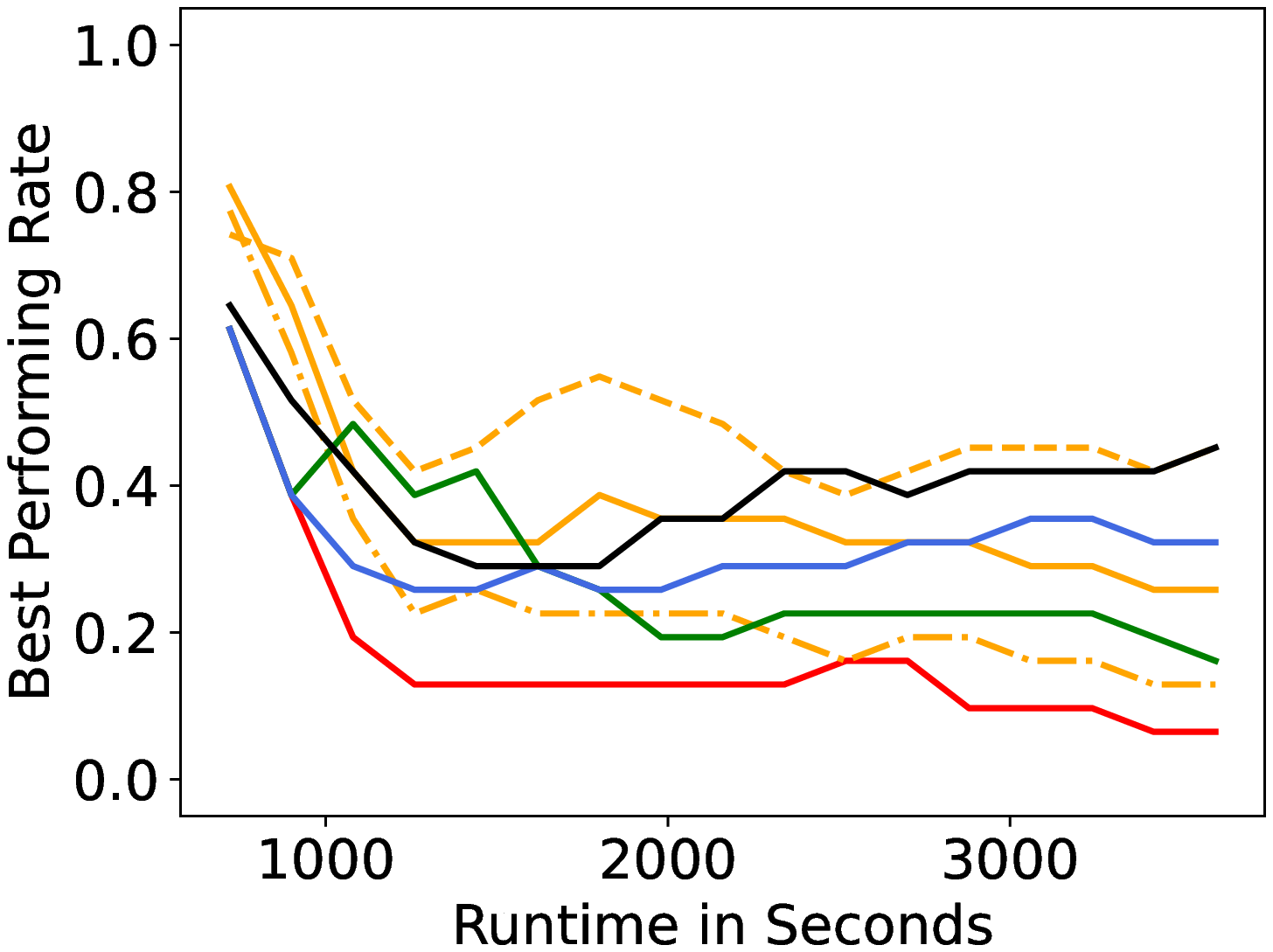}
  \includegraphics[height=3.2cm]{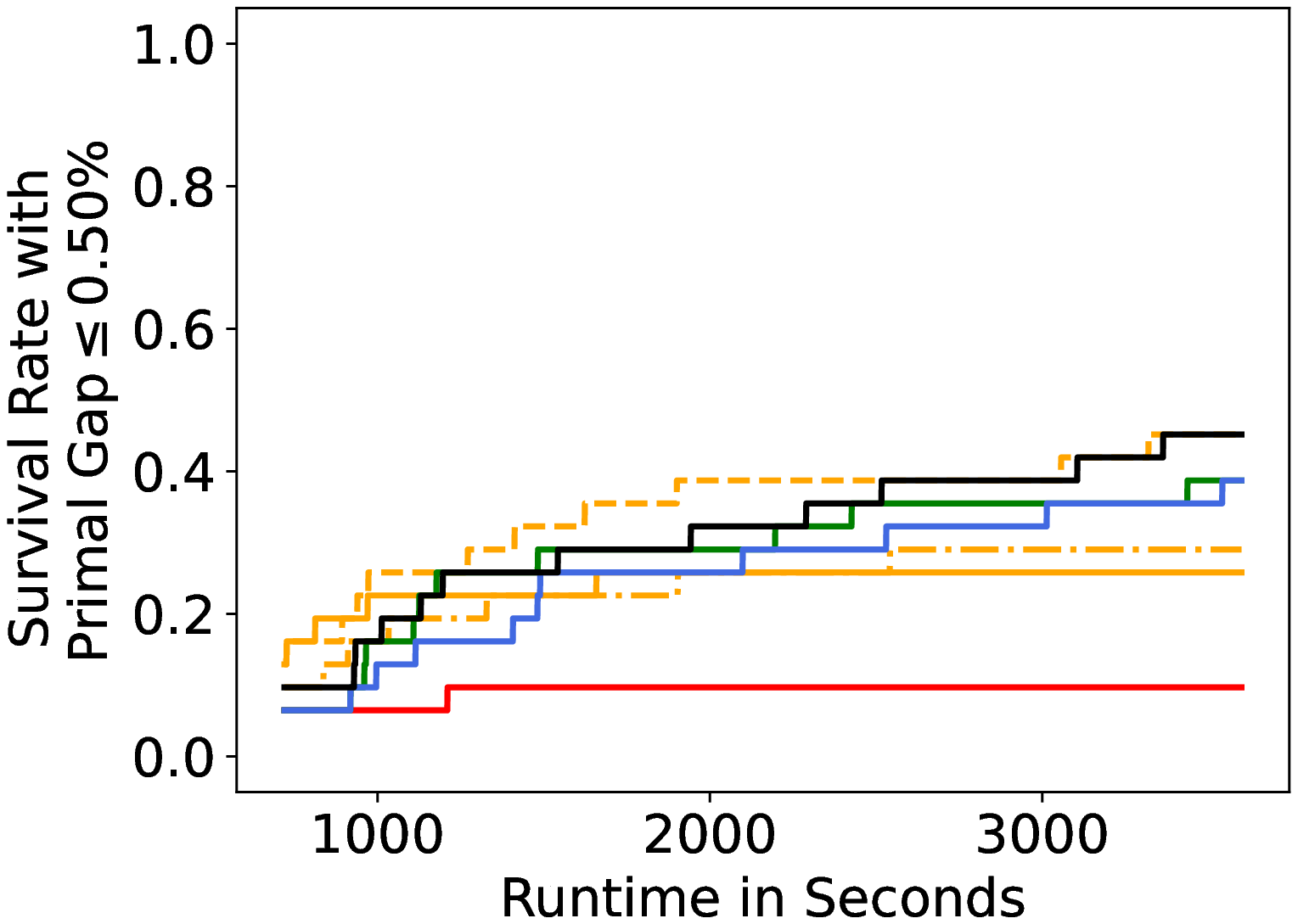}
    \caption{Results on 31 selected MIPLIB instances: The best performing rate as a function of time (left) and the survival rate over 31 instances as a function of time to meet the primal gap threshold 0.50\% (right).\label{res::winrateMIPLIB}}
\end{figure}

\subsubsection{Results on Selected MIPLIB Instances}
Finally, we examine how well \LBRELAX and its variants perform on ILPs that are diverse in structures and sizes. We test them on the MIPLIB dataset \cite{MIPLIB}. MIPLIB contains COPs from various real-world domains. We follow a procedure similar to \cite{wu2021learning} to filter out instances where we first filter to retain ILP instances with only binary variables. Among these, we select instances that are not too easy to solve but relatively easy to find a feasible solution for. Specifically, we filter out those that BnB can optimally solve within 3 hours (too easy) or BnB cannot find any solutions within 10 minutes (too hard), which gives us 35 instances. 
For all LNS approaches, we run BnB for 10 minutes to find the initial solution and set the time limit to 10 minutes for each repair operation. The initial neighborhood size $k_0$ is set to 20\% of the number of binary variables. We compare \LBRELAX, \LBRELAXRR and \LBRELAXS with the non-ML baselines. We further filter out 4 instances that no approach can find a better solution than the initial one, which finally gives us 31 instances.

Figure \ref{res::winrateMIPLIB} shows the winning rate as a function of time for each approach on the 31 instances. The best performing rate at a time $q$ for an approach is  the fraction of instances on which it achieves the best performance (including ties) compared to all approaches in the portfolio. 
\LBRELAX, \LBRELAXRR and \LBRELAXS  achieve the best performance with less than 1000 seconds on 25, 23 and 24 instances out of 35, respectively. \LBRELAXRR has the highest best performing rates at different time cutoffs and ties with BnB at 14 instances at the 60-minute mark. 
Figure \ref{res::winrateMIPLIB} also shows the survival rate over the 31 instances as a function of time to meet the primal gap threshold 0.50\%.
It demonstrates that \RANDOM, \GRAPH and BnB are competitive with our approaches but overall \LBRELAXRR has the highest survival rate over time. 
On some instances,  \LBRELAX and its variants can significantly outperform the baselines and we show the anytime performance on those in Appendix.

\section{Conclusion}

In this paper, we focused on designing effective and efficient destroy heuristics to select neighborhoods in LNS for ILPs. LB is an effective destroy heuristic but is slow to run. We therefore proposed \LBRELAX, \LBRELAXS and \LBRELAXRR  to approximate LB's decisions by solving its  LP relaxation that is a lot faster to run. Empirically, we showed that \LBRELAX, \LBRELAXS and \LBRELAXRR efficiently selected almost as effective neighborhoods as LB and achieved state-of-the-art performance when compared against non-ML and ML approaches. One limitation of our approaches is that they do not work well on some problem domains, however we showed that they still outperformed the baselines on 14 to 25 (depending on the time cutoff) out of 31 difficult MIPLIB instances that are diverse in problem domains, structures and sizes. The other limitation is that they can get stuck at local minima. To address this issue, we proposed techniques to randomize the heuristics and adaptively adjust the neighborhood sizes. For future work, one could improve \LBRELAX and its variants to make them applicable on more problem domains. In addition, instead of using hard-coded rules for scheduling the randomized heuristic in \LBRELAXRR, one could use adaptive LNS to select destroy heuristics to run. It is also future work to develop theoretical claims to help support and explain the effectiveness of \LBRELAX, \LBRELAXS, \LBRELAXRR and possibly their other variants. 

\subsubsection*{Acknowledgements.} 
This paper reports on research done while Taoan Huang and Aaron Ferber were interns at Meta AI (FAIR). The research at the University of Southern California was supported by the National Science Foundation (NSF) under grant number 2112533.

%
%
%
\bibliographystyle{splncs04}
\bibliography{reference}

\begin{thebibliography}{10}
\providecommand{\url}[1]{\texttt{#1}}
\providecommand{\urlprefix}{URL }
\providecommand{\doi}[1]{https://doi.org/#1}

\bibitem{achterberg2012rounding}
Achterberg, T., Berthold, T., Hendel, G.: Rounding and propagation heuristics
  for mixed integer programming. In: Operations research proceedings 2011, pp.
  71--76. Springer (2012)

\bibitem{achterberg2013mixed}
Achterberg, T., Wunderling, R.: Mixed integer programming: Analyzing 12 years
  of progress. In: Facets of combinatorial optimization, pp. 449--481. Springer
  (2013)

\bibitem{albert2002statistical}
Albert, R., Barab{\'a}si, A.L.: Statistical mechanics of complex networks.
  Reviews of modern physics  \textbf{74}(1), ~47 (2002)

\bibitem{amaral2008exact}
Amaral, A.R.: An exact approach to the one-dimensional facility layout problem.
  Operations research  \textbf{56}(4),  1026--1033 (2008)

\bibitem{azi2014adaptive}
Azi, N., Gendreau, M., Potvin, J.Y.: An adaptive large neighborhood search for
  a vehicle routing problem with multiple routes. Computers \& Operations
  Research  \textbf{41},  167--173 (2014)

\bibitem{berthold2006primal}
Berthold, T.: Primal heuristics for mixed integer programs. Ph.D. thesis, Zuse
  Institute Berlin (ZIB) (2006)

\bibitem{berthold2014rens}
Berthold, T.: Rens. Mathematical Programming Computation  \textbf{6}(1),
  33--54 (2014)

\bibitem{BestuzhevaEtal2021OO}
Bestuzheva, K., Besan{\c{c}}on, M., Chen, W.K., Chmiela, A., Donkiewicz, T.,
  van Doornmalen, J., Eifler, L., Gaul, O., Gamrath, G., Gleixner, A.,
  Gottwald, L., Graczyk, C., Halbig, K., Hoen, A., Hojny, C., van~der Hulst,
  R., Koch, T., L{\"u}bbecke, M., Maher, S.J., Matter, F., M{\"u}hmer, E.,
  M{\"u}ller, B., Pfetsch, M.E., Rehfeldt, D., Schlein, S., Schl{\"o}sser, F.,
  Serrano, F., Shinano, Y., Sofranac, B., Turner, M., Vigerske, S.,
  Wegscheider, F., Wellner, P., Weninger, D., Witzig, J.: {The SCIP
  Optimization Suite 8.0}. Technical report, Optimization Online (December
  2021), \url{http://www.optimization-online.org/DB_HTML/2021/12/8728.html}

\bibitem{chen2019learning}
Chen, X., Tian, Y.: Learning to perform local rewriting for combinatorial
  optimization. Advances in Neural Information Processing Systems  \textbf{32}
  (2019)

\bibitem{danna2005exploring}
Danna, E., Rothberg, E., Pape, C.L.: Exploring relaxation induced neighborhoods
  to improve mip solutions. Mathematical Programming  \textbf{102}(1),  71--90
  (2005)

\bibitem{de2003combinatorial}
De~Vries, S., Vohra, R.V.: Combinatorial auctions: A survey. INFORMS Journal on
  computing  \textbf{15}(3),  284--309 (2003)

\bibitem{dilkina2010solving}
Dilkina, B., Gomes, C.P.: Solving connected subgraph problems in wildlife
  conservation. In: CPAIOR. vol.~6140, pp. 102--116. Springer (2010)

\bibitem{fischetti2003local}
Fischetti, M., Lodi, A.: Local branching. Mathematical programming
  \textbf{98}(1),  23--47 (2003)

\bibitem{gasse2019exact}
Gasse, M., Ch{\'e}telat, D., Ferroni, N., Charlin, L., Lodi, A.: Exact
  combinatorial optimization with graph convolutional neural networks. Advances
  in Neural Information Processing Systems  \textbf{32} (2019)

\bibitem{ghosh2007dins}
Ghosh, S.: Dins, a mip improvement heuristic. In: International Conference on
  Integer Programming and Combinatorial Optimization. pp. 310--323. Springer
  (2007)

\bibitem{MIPLIB}
Gleixner, A., Hendel, G., Gamrath, G., Achterberg, T., Bastubbe, M., Berthold,
  T., Christophel, P.M., Jarck, K., Koch, T., Linderoth, J., L\"ubbecke, M.,
  Mittelmann, H.D., Ozyurt, D., Ralphs, T.K., Salvagnin, D., Shinano, Y.:
  {MIPLIB 2017: Data-Driven Compilation of the 6th Mixed-Integer Programming
  Library}. Mathematical Programming Computation  (2021).
  \doi{10.1007/s12532-020-00194-3},
  \url{https://doi.org/10.1007/s12532-020-00194-3}

\bibitem{gurobi}
{Gurobi Optimization, LLC}: {Gurobi Optimizer Reference Manual} (2022),
  \url{https://www.gurobi.com}

\bibitem{hendel2022adaptive}
Hendel, G.: Adaptive large neighborhood search for mixed integer programming.
  Mathematical Programming Computation  \textbf{14}(2),  185--221 (2022)

\bibitem{heragu1991efficient}
Heragu, S.S., Kusiak, A.: Efficient models for the facility layout problem.
  European Journal of Operational Research  \textbf{53}(1),  1--13 (1991)

\bibitem{hottung2020neural}
Hottung, A., Tierney, K.: Neural large neighborhood search for the capacitated
  vehicle routing problem. In: ECAI 2020, pp. 443--450. IOS Press (2020)

\bibitem{huang2020enhancing}
Huang, T., Dilkina, B.: Enhancing seismic resilience of water pipe networks.
  In: Proceedings of the 3rd ACM SIGCAS Conference on Computing and Sustainable
  Societies. pp. 44--52 (2020)

\bibitem{huang2022anytime}
Huang, T., Li, J., Koenig, S., Dilkina, B.: Anytime multi-agent path finding
  via machine learning-guided large neighborhood search. In: Proceedings of the
  AAAI Conference on Artificial Intelligence (AAAI). pp. 9368--9376 (2022)

\bibitem{HuangICAPS23}
Huang, T., Shivashankar, V., Caldara, M., Durham, J., Li, J., Dilkina, B.,
  Koenig, S.: Deadline-aware multi-agent tour planning. In: Proceedings of the
  International Conference on Automated Planning and Scheduling (ICAPS) (2023)

\bibitem{khalil2016learning}
Khalil, E., Le~Bodic, P., Song, L., Nemhauser, G., Dilkina, B.: Learning to
  branch in mixed integer programming. In: Proceedings of the AAAI Conference
  on Artificial Intelligence. vol.~30 (2016)

\bibitem{kleinberg2006algorithm}
Kleinberg, J., Tardos, E.: Algorithm design. Pearson Education India (2006)

\bibitem{kovacs2012adaptive}
Kovacs, A.A., Parragh, S.N., Doerner, K.F., Hartl, R.F.: Adaptive large
  neighborhood search for service technician routing and scheduling problems.
  Journal of scheduling  \textbf{15}(5),  579--600 (2012)

\bibitem{land2010automatic}
Land, A.H., Doig, A.G.: An automatic method for solving discrete programming
  problems. In: 50 Years of Integer Programming 1958-2008, pp. 105--132.
  Springer (2010)

\bibitem{leyton2000towards}
Leyton-Brown, K., Pearson, M., Shoham, Y.: Towards a universal test suite for
  combinatorial auction algorithms. In: Proceedings of the 2nd ACM conference
  on Electronic commerce. pp. 66--76 (2000)

\bibitem{LiIJCAI21}
Li, J., Chen, Z., Harabor, D., Stuckey, P.J., Koenig, S.: Anytime multi-agent
  path finding via large neighborhood search. In: Proceedings of the
  International Joint Conference on Artificial Intelligence (IJCAI). pp.
  4127--4135 (2021)

\bibitem{LiAAAI22}
Li, J., Chen, Z., Harabor, D., Stuckey, P.J., Koenig, S.: {MAPF-LNS2}: Fast
  repairing for multi-agent path finding via large neighborhood search. In:
  Proceedings of the AAAI Conference on Artificial Intelligence (AAAI). pp.
  10256--10265 (2022)

\bibitem{li2021learning}
Li, S., Yan, Z., Wu, C.: Learning to delegate for large-scale vehicle routing.
  Advances in Neural Information Processing Systems  \textbf{34},  26198--26211
  (2021)

\bibitem{liurevisiting}
Liu, D., Fischetti, M., Lodi, A.: Revisiting local branching with a machine
  learning lens

\bibitem{lu2019learning}
Lu, H., Zhang, X., Yang, S.: A learning-based iterative method for solving
  vehicle routing problems. In: International conference on learning
  representations (2019)

\bibitem{maher2017scip}
Maher, S.J., Fischer, T., Gally, T., Gamrath, G., Gleixner, A., Gottwald, R.L.,
  Hendel, G., Koch, T., L{\"u}bbecke, M., Miltenberger, M., et~al.: The scip
  optimization suite 4.0  (2017)

\bibitem{manne1960job}
Manne, A.S.: On the job-shop scheduling problem. Operations research
  \textbf{8}(2),  219--223 (1960)

\bibitem{pohl1970heuristic}
Pohl, I.: Heuristic search viewed as path finding in a graph. Artificial
  intelligence  \textbf{1}(3-4),  193--204 (1970)

\bibitem{ropke2006adaptive}
Ropke, S., Pisinger, D.: An adaptive large neighborhood search heuristic for
  the pickup and delivery problem with time windows. Transportation science
  \textbf{40}(4),  455--472 (2006)

\bibitem{rothberg2007evolutionary}
Rothberg, E.: An evolutionary algorithm for polishing mixed integer programming
  solutions. INFORMS Journal on Computing  \textbf{19}(4),  534--541 (2007)

\bibitem{scavuzzo2022learning}
Scavuzzo, L., Chen, F.Y., Ch{\'e}telat, D., Gasse, M., Lodi, A., Yorke-Smith,
  N., Aardal, K.: Learning to branch with tree mdps. arXiv preprint
  arXiv:2205.11107  (2022)

\bibitem{smith2017glns}
Smith, S.L., Imeson, F.: Glns: An effective large neighborhood search heuristic
  for the generalized traveling salesman problem. Computers \& Operations
  Research  \textbf{87},  1--19 (2017)

\bibitem{song2020general}
Song, J., Yue, Y., Dilkina, B., et~al.: A general large neighborhood search
  framework for solving integer linear programs. Advances in Neural Information
  Processing Systems  \textbf{33},  20012--20023 (2020)

\bibitem{sonnerat2021learning}
Sonnerat, N., Wang, P., Ktena, I., Bartunov, S., Nair, V.: Learning a large
  neighborhood search algorithm for mixed integer programs. arXiv preprint
  arXiv:2107.10201  (2021)

\bibitem{toth2002vehicle}
Toth, P., Vigo, D.: The vehicle routing problem. SIAM (2002)

\bibitem{wu2021learning}
Wu, Y., Song, W., Cao, Z., Zhang, J.: Learning large neighborhood search policy
  for integer programming. Advances in Neural Information Processing Systems
  \textbf{34},  30075--30087 (2021)

\bibitem{vzulj2018hybrid}
{\v{Z}}ulj, I., Kramer, S., Schneider, M.: A hybrid of adaptive large
  neighborhood search and tabu search for the order-batching problem. European
  Journal of Operational Research  \textbf{264}(2),  653--664 (2018)

\end{thebibliography}

\newpage
\appendix
\section*{\LARGE Appendix}

\begin{figure}[bp]
    \centering
    \includegraphics[width=\textwidth]{figure/legend_timeVSobj.eps}
    \begin{subfigure}[htbp]{0.49\textwidth}
		\centering
		\includegraphics[height=4cm]{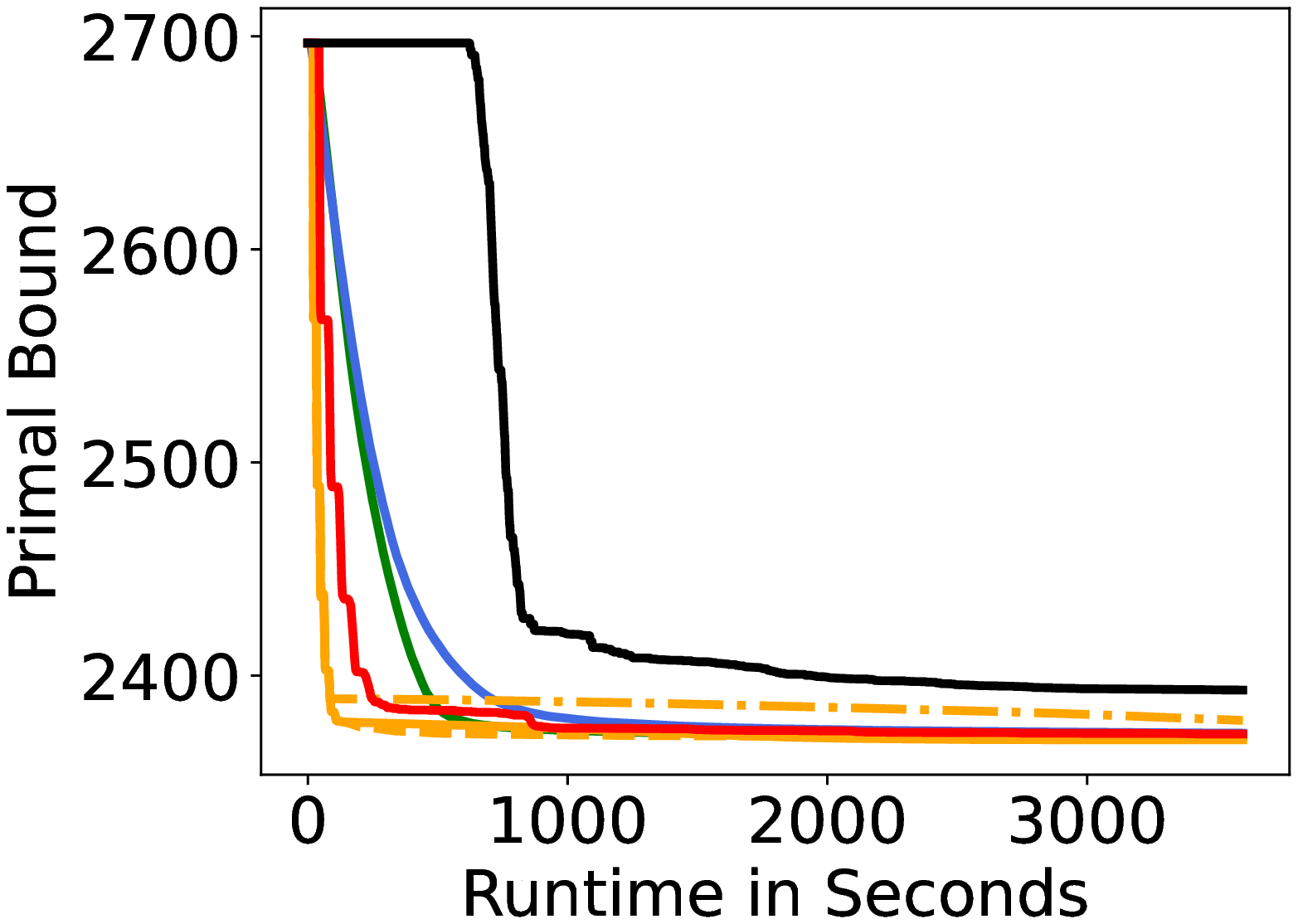}
		\caption{MVC}
	\end{subfigure}\,\,
	\begin{subfigure}[htbp]{0.49\textwidth}
		\centering
		\includegraphics[height=4cm]{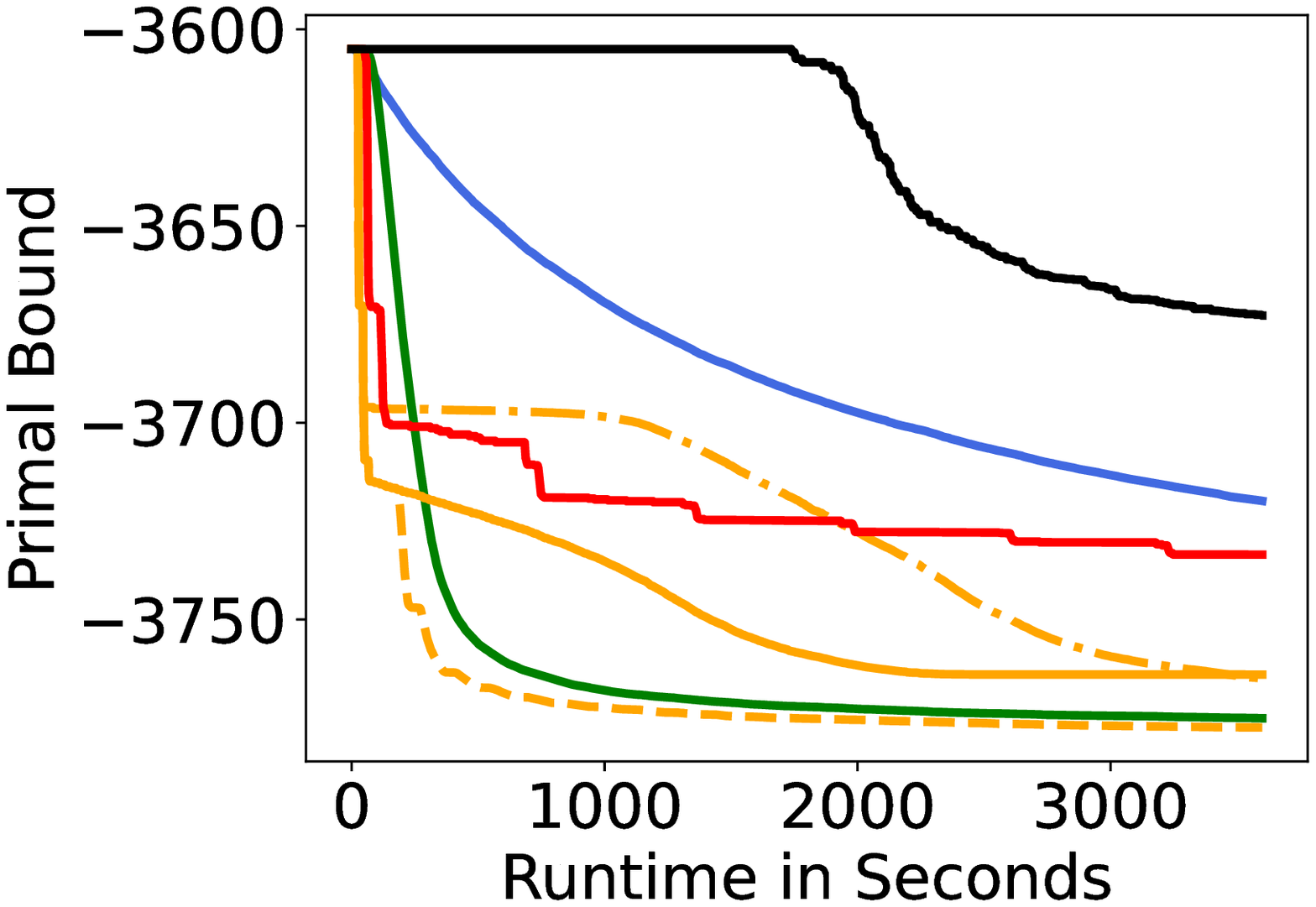}

			\caption{MIS}
	\end{subfigure}
	\begin{subfigure}[htbp]{0.49\textwidth}
		\centering
		\includegraphics[height=4cm]{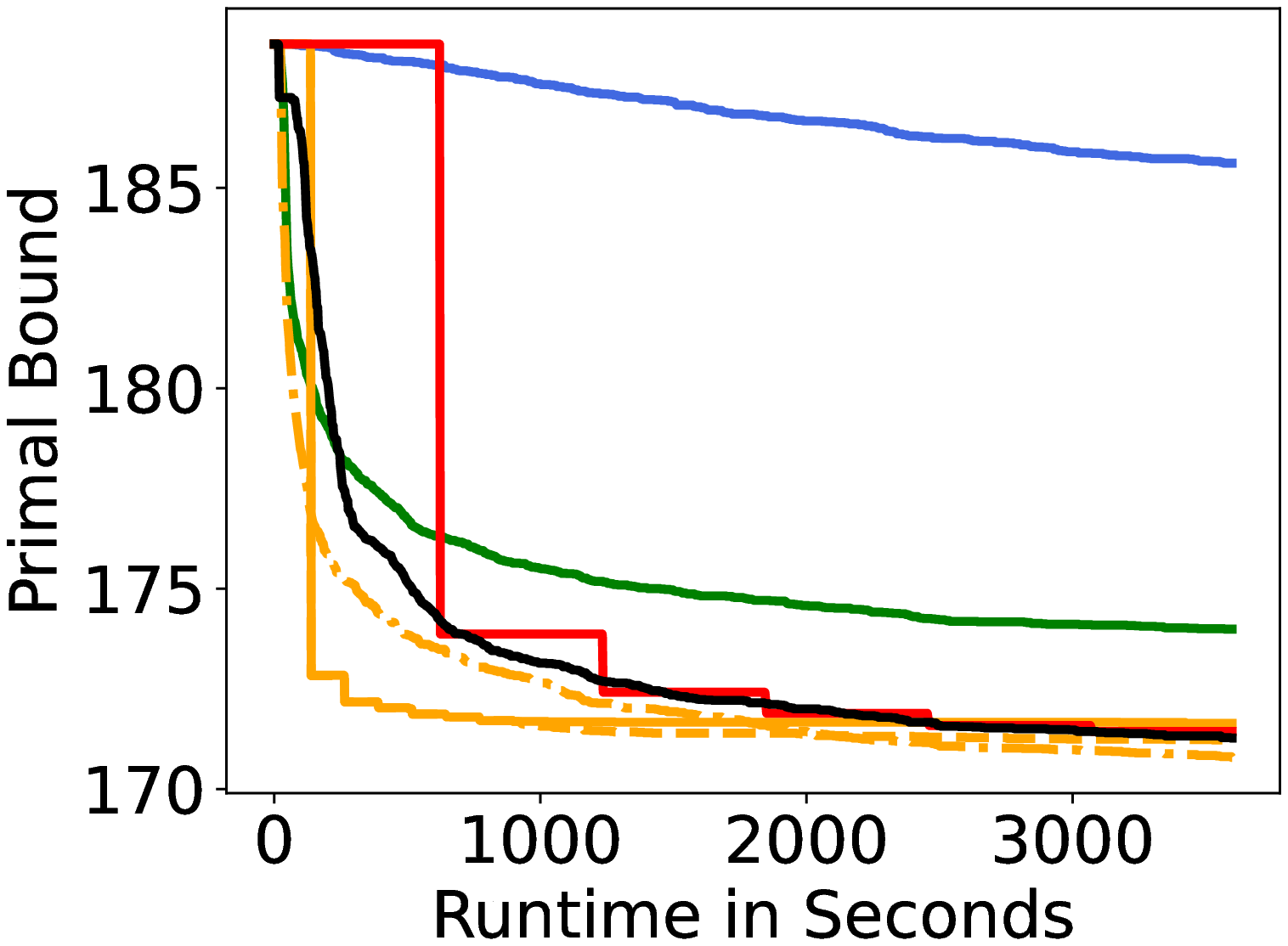}

			\caption{SC}
	\end{subfigure}
	\begin{subfigure}[htbp]{0.49\textwidth}
		\centering
		\includegraphics[height=4cm]{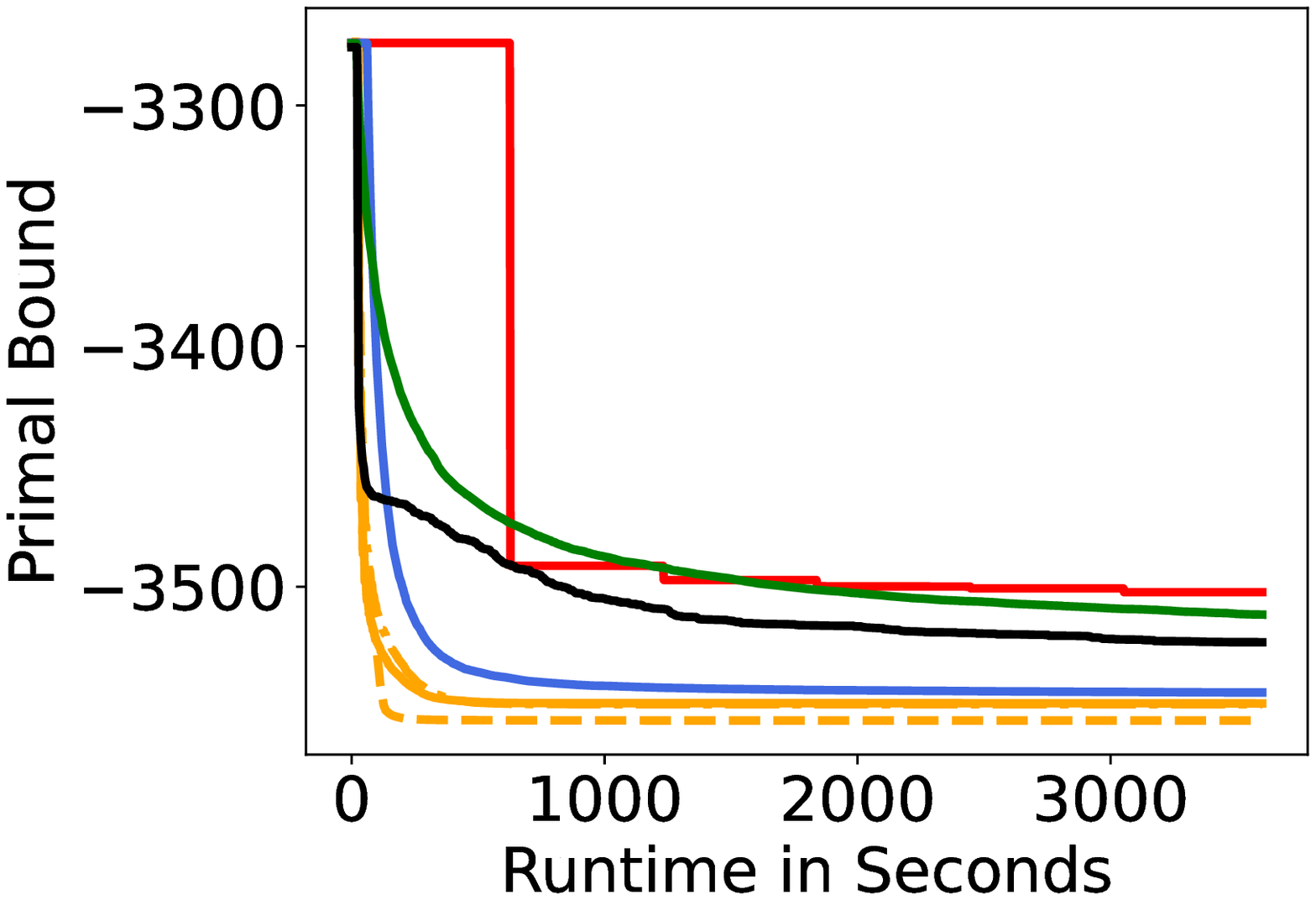}
			\caption{MK}
	\end{subfigure}
    \caption{Comparison with non-ML approaches: The primal bound as a function of time as a function of time, averaged over 100 instances. \label{res::bound}}
\end{figure}

\begin{table}[htbp]
\scriptsize
\centering
\caption{\small Primal gap (PG) (in percent) and primal integral (PI) at 15 minutes time cutoff, averaged over 100 instances, and their standard deviations. \label{res::nonMLtable15}}

\begin{tabular}{c|rr|rr}
\hline
                       & \multicolumn{2}{c|}{MVC}                                                                        & \multicolumn{2}{c}{MIS}                                                                         \\ \hline
                                    & \multicolumn{1}{c|}{PG (\%)}         & \multicolumn{1}{c|}{PI}              & \multicolumn{1}{c|}{PG (\%)}         & \multicolumn{1}{c}{PI} \\ \hline
BnB                    & \multicolumn{1}{r|}{2.12$\pm$1.48} & 93.3$\pm$6.2           & \multicolumn{1}{r|}{4.60$\pm$0.31} & 43.2$\pm$2.8           \\ \hline
LB                     &  \multicolumn{1}{r|}{0.30$\pm$0.15} & 16.7$\pm$1.5             & \multicolumn{1}{r|}{1.58$\pm$0.28} & 19.5$\pm$2.7             \\ \hline
\RANDOM                 &  \multicolumn{1}{r|}{0.24$\pm$0.07} & 28.3$\pm$1.7             & \multicolumn{1}{r|}{0.32$\pm$0.10} & 13.6$\pm$1.3             \\ \hline
\GRAPH                  &  \multicolumn{1}{r|}{0.53$\pm$0.09} & 34.0$\pm$1.9             & \multicolumn{1}{r|}{3.01$\pm$0.26} & 33.3$\pm$2.4             \\ \hline
\LBRELAX                & \multicolumn{1}{r|}{0.28$\pm$0.08} & 7.8$\pm$0.8              & \multicolumn{1}{r|}{1.22$\pm$0.19} & 14.5$\pm$1.9             \\ \hline
\LBRELAXRR              & \multicolumn{1}{r|}{\bf0.14$\pm$0.05} & {\bf6.7$\pm$0.6}              & \multicolumn{1}{r|}{\bf 0.18$\pm$0.08} & {\bf 7.0$\pm$0.9}             \\ \hline
\LBRELAXS               & \multicolumn{1}{r|}{0.80$\pm$0.26} & 11.7$\pm$2.2             &  \multicolumn{1}{r|}{2.14$\pm$0.42} & 20.3$\pm$3.7            \\ \hline
\multicolumn{1}{l|}{} & \multicolumn{2}{c|}{SC}                                                                         & \multicolumn{2}{c}{MK}                                                                          \\ \hline
BnB                     & \multicolumn{1}{r|}{2.30$\pm$1.36} & 46.0$\pm$13.1            & \multicolumn{1}{r|}{1.49$\pm$0.67} & 31.5$\pm$3.6            \\ \hline
LB                      & \multicolumn{1}{r|}{2.61$\pm$1.32} & 70.7$\pm$15.8            &  \multicolumn{1}{r|}{1.81$\pm$0.45} & 54.6$\pm$6.5            \\ \hline
\RANDOM                 & \multicolumn{1}{r|}{3.60$\pm$1.45} & 43.7$\pm$14.0           & \multicolumn{1}{r|}{1.99$\pm$0.45} & 28.7$\pm$4.9           \\ \hline
\GRAPH                  &\multicolumn{1}{r|}{9.78$\pm$2.19} & 90.0$\pm$20.0           &  \multicolumn{1}{r|}{0.41$\pm$0.14} & 14.1$\pm$1.9             \\ \hline
\LBRELAX                & \multicolumn{1}{r|}{\bf1.40$\pm$0.98} & {\bf26.3$\pm$8.8}            & \multicolumn{1}{r|}{0.20$\pm$0.09} & 5.8$\pm$0.8             \\ \hline
\LBRELAXRR              & \multicolumn{1}{r|}{\bf1.40$\pm$0.98} & {\bf26.3$\pm$8.8}          &  \multicolumn{1}{r|}{\bf0.00$\pm$0.01} & {\bf3.7$\pm$0.4}              \\ \hline
\LBRELAXS               & \multicolumn{1}{r|}{2.04$\pm$1.26} & 30.1$\pm$11.7            & \multicolumn{1}{r|}{0.19$\pm$0.07} & 6.6$\pm$0.7             \\ \hline
\end{tabular}
\end{table}

\begin{table}[htbp]
\scriptsize
\centering
\caption{\small Primal gap (PG) (in percent) and primal integral (PI) at 30 minutes time cutoff, averaged over 100 instances, and their standard deviations.  \label{res::nonMLtable30}}

\begin{tabular}{c|rr|rr}
\hline
                       & \multicolumn{2}{c|}{MVC}                                                                        & \multicolumn{2}{c}{MIS}                                                                         \\ \hline
                                    & \multicolumn{1}{c|}{PG (\%)}         & \multicolumn{1}{c|}{PI}              & \multicolumn{1}{c|}{PG (\%)}         & \multicolumn{1}{c}{PI} \\ \hline
BnB                    & \multicolumn{1}{r|}{1.36$\pm$0.57} & 108.5$\pm$9.0           & \multicolumn{1}{r|}{4.51$\pm$0.55} & 84.6$\pm$5.6           \\ \hline
LB                     &  \multicolumn{1}{r|}{0.22$\pm$0.12} & 18.9$\pm$2.2             & \multicolumn{1}{r|}{1.43$\pm$0.33} & 32.9$\pm$4.9             \\ \hline
\RANDOM                 &  \multicolumn{1}{r|}{0.15$\pm$0.05} & 29.9$\pm$1.8             & \multicolumn{1}{r|}{0.18$\pm$0.07} & 15.7$\pm$1.8             \\ \hline
\GRAPH                  &  \multicolumn{1}{r|}{0.26$\pm$0.07} & 37.1$\pm$2.1             & \multicolumn{1}{r|}{2.27$\pm$0.22} & 56.6$\pm$4.4             \\ \hline
\LBRELAX                & \multicolumn{1}{r|}{\bf0.10$\pm$0.05} & 9.5$\pm$1.3              & \multicolumn{1}{r|}{0.52$\pm$0.12} & 22.1$\pm$3.1             \\ \hline
\LBRELAXRR              & \multicolumn{1}{r|}{0.11$\pm$0.05} & {\bf7.8$\pm$1.0}              & \multicolumn{1}{r|}{\bf 0.10$\pm$0.05} & {\bf 8.2$\pm$1.3}             \\ \hline
\LBRELAXS               & \multicolumn{1}{r|}{0.70$\pm$0.23} & 18.5$\pm$4.3             &  \multicolumn{1}{r|}{1.52$\pm$0.26} & 37.4$\pm$6.8            \\ \hline
\multicolumn{1}{l|}{} & \multicolumn{2}{c|}{SC}                                                                         & \multicolumn{2}{c}{MK}                                                                          \\ \hline
BnB                     & \multicolumn{1}{r|}{1.65$\pm$1.22} & 63.2$\pm$22.8            & \multicolumn{1}{r|}{1.11$\pm$0.62} & 42.7$\pm$8.3            \\ \hline
LB                      & \multicolumn{1}{r|}{1.80$\pm$1.13} & 89.6$\pm$22.6            & \multicolumn{1}{r|}{1.64$\pm$0.46} & 69.9$\pm$7.5            \\ \hline
\RANDOM                 & \multicolumn{1}{r|}{3.09$\pm$1.38} & 73.5$\pm$24.9            & \multicolumn{1}{r|}{1.54$\pm$0.42} & 44.3$\pm$8.4             \\ \hline
\GRAPH                  &\multicolumn{1}{r|}{9.34$\pm$2.30} & 175.9$\pm$39.1           &  \multicolumn{1}{r|}{0.36$\pm$0.14} & 17.5$\pm$2.8             \\ \hline
\LBRELAX                & \multicolumn{1}{r|}{1.39$\pm$0.97} & {38.9$\pm$16.9}            & \multicolumn{1}{r|}{0.20$\pm$0.09} & 7.7$\pm$1.5             \\ \hline
\LBRELAXRR              & \multicolumn{1}{r|}{\bf1.23$\pm$0.91} & {\bf37.7$\pm$16.2}            &  \multicolumn{1}{r|}{\bf0.00$\pm$0.00} & {\bf3.7$\pm$0.4}              \\ \hline
\LBRELAXS               & \multicolumn{1}{r|}{1.37$\pm$1.04} & 44.8$\pm$19.9            & \multicolumn{1}{r|}{0.19$\pm$0.07} & 8.4$\pm$1.2             \\ \hline
\end{tabular}
\end{table}

\begin{table}[htbp]
\scriptsize
\centering
\caption{\small Primal gap (PG) (in percent) and primal integral (PI) at 45 minutes time cutoff, averaged over 100 instances, and their standard deviations.  \label{res::nonMLtable45}}

\begin{tabular}{c|rr|rr}
\hline
                       & \multicolumn{2}{c|}{MVC}                                                                        & \multicolumn{2}{c}{MIS}                                                                         \\ \hline
                                    & \multicolumn{1}{c|}{PG (\%)}         & \multicolumn{1}{c|}{PI}              & \multicolumn{1}{c|}{PG (\%)}         & \multicolumn{1}{c}{PI} \\ \hline
BnB                    & \multicolumn{1}{r|}{1.08$\pm$0.48} & 119.3$\pm$11.7           & \multicolumn{1}{r|}{3.09$\pm$1.22} & 117.8$\pm$10.6           \\ \hline
LB                     &  \multicolumn{1}{r|}{0.18$\pm$0.10} & 20.7$\pm$3.0             & \multicolumn{1}{r|}{1.29$\pm$0.31} & 45.1$\pm$7.2             \\ \hline
\RANDOM                 &  \multicolumn{1}{r|}{0.13$\pm$0.05} & 31.2$\pm$2.0             & \multicolumn{1}{r|}{0.12$\pm$0.06} & 17.0$\pm$2.2             \\ \hline
\GRAPH                  &  \multicolumn{1}{r|}{0.20$\pm$0.05} & 39.1$\pm$2.3             & \multicolumn{1}{r|}{1.84$\pm$0.21} & 75.0$\pm$6.0             \\ \hline
\LBRELAX                & \multicolumn{1}{r|}{\bf0.04$\pm$0.03} & 10.0$\pm$1.5              & \multicolumn{1}{r|}{0.39$\pm$0.12} & 25.9$\pm$3.6             \\ \hline
\LBRELAXRR              & \multicolumn{1}{r|}{0.10$\pm$0.05} & {\bf8.7$\pm$1.3}              & \multicolumn{1}{r|}{\bf 0.06$\pm$0.04} & {\bf 8.9$\pm$1.5}             \\ \hline
\LBRELAXS               & \multicolumn{1}{r|}{0.59$\pm$0.21} & 24.3$\pm$6.3             &  \multicolumn{1}{r|}{0.68$\pm$0.23} & 47.3$\pm$9.1            \\ \hline
\multicolumn{1}{l|}{} & \multicolumn{2}{c|}{SC}                                                                         & \multicolumn{2}{c}{MK}                                                                          \\ \hline
BnB                     & \multicolumn{1}{r|}{1.29$\pm$1.03} & 76.4$\pm$31.1            & \multicolumn{1}{r|}{1.00$\pm$0.61} & 52.2$\pm$13.1            \\ \hline
LB                      & \multicolumn{1}{r|}{1.32$\pm$0.99} & 102.7$\pm$28.9            &  \multicolumn{1}{r|}{1.55$\pm$0.47} & 84.0$\pm$9.8            \\ \hline
\RANDOM                 & \multicolumn{1}{r|}{2.78$\pm$1.30} & 99.8$\pm$35.5           & \multicolumn{1}{r|}{1.36$\pm$0.38} & 57.3$\pm$11.6            \\ \hline
\GRAPH                  &\multicolumn{1}{r|}{9.01$\pm$2.24} & 258.4$\pm$58.5           &  \multicolumn{1}{r|}{0.34$\pm$0.13} & 20.6$\pm$3.8             \\ \hline
\LBRELAX                & \multicolumn{1}{r|}{1.39$\pm$0.97} & {51.4$\pm$25.2}            & \multicolumn{1}{r|}{0.20$\pm$0.09} & 9.5$\pm$2.2             \\ \hline
\LBRELAXRR              & \multicolumn{1}{r|}{1.17$\pm$0.89} & {\bf48.5$\pm$23.9}            &  \multicolumn{1}{r|}{\bf0.00$\pm$0.00} & {\bf3.7$\pm$0.4}              \\ \hline
\LBRELAXS               & \multicolumn{1}{r|}{\bf1.01$\pm$0.89} & 55.1$\pm$26.3            & \multicolumn{1}{r|}{0.19$\pm$0.07} & 10.1$\pm$1.8             \\ \hline
\end{tabular}
\end{table}

\begin{figure}[tbp]
    \centering
    \includegraphics[width=\textwidth]{figure/legend_timeVSobj.eps}
    \begin{subfigure}[htbp]{0.49\textwidth}
		\centering
		\includegraphics[height=4cm]{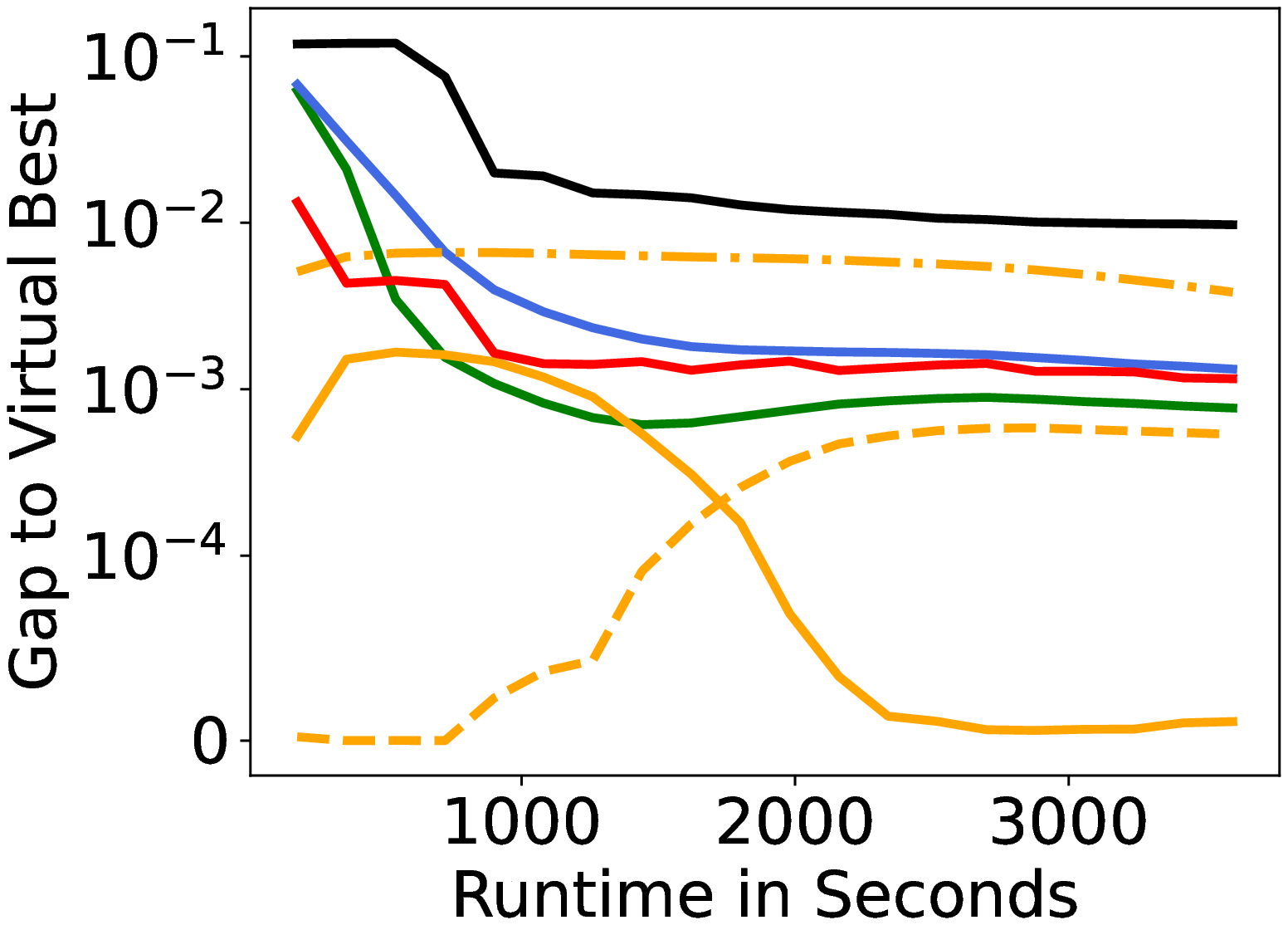}
		\caption{MVC}
	\end{subfigure}\,\,
	\begin{subfigure}[htbp]{0.49\textwidth}
		\centering
		\includegraphics[height=4cm]{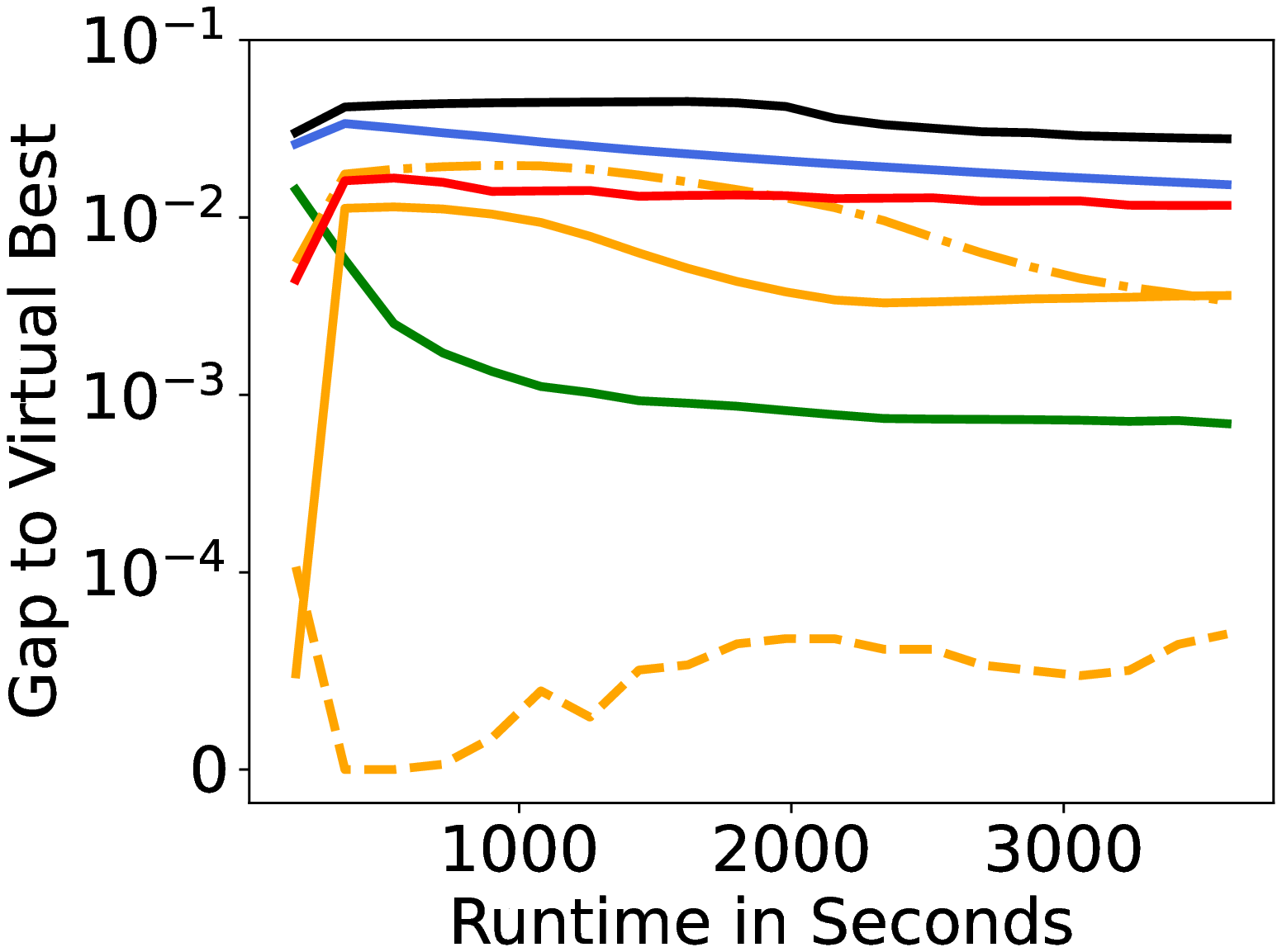}

			\caption{MIS}
	\end{subfigure}
	\begin{subfigure}[htbp]{0.49\textwidth}
		\centering
		\includegraphics[height=4cm]{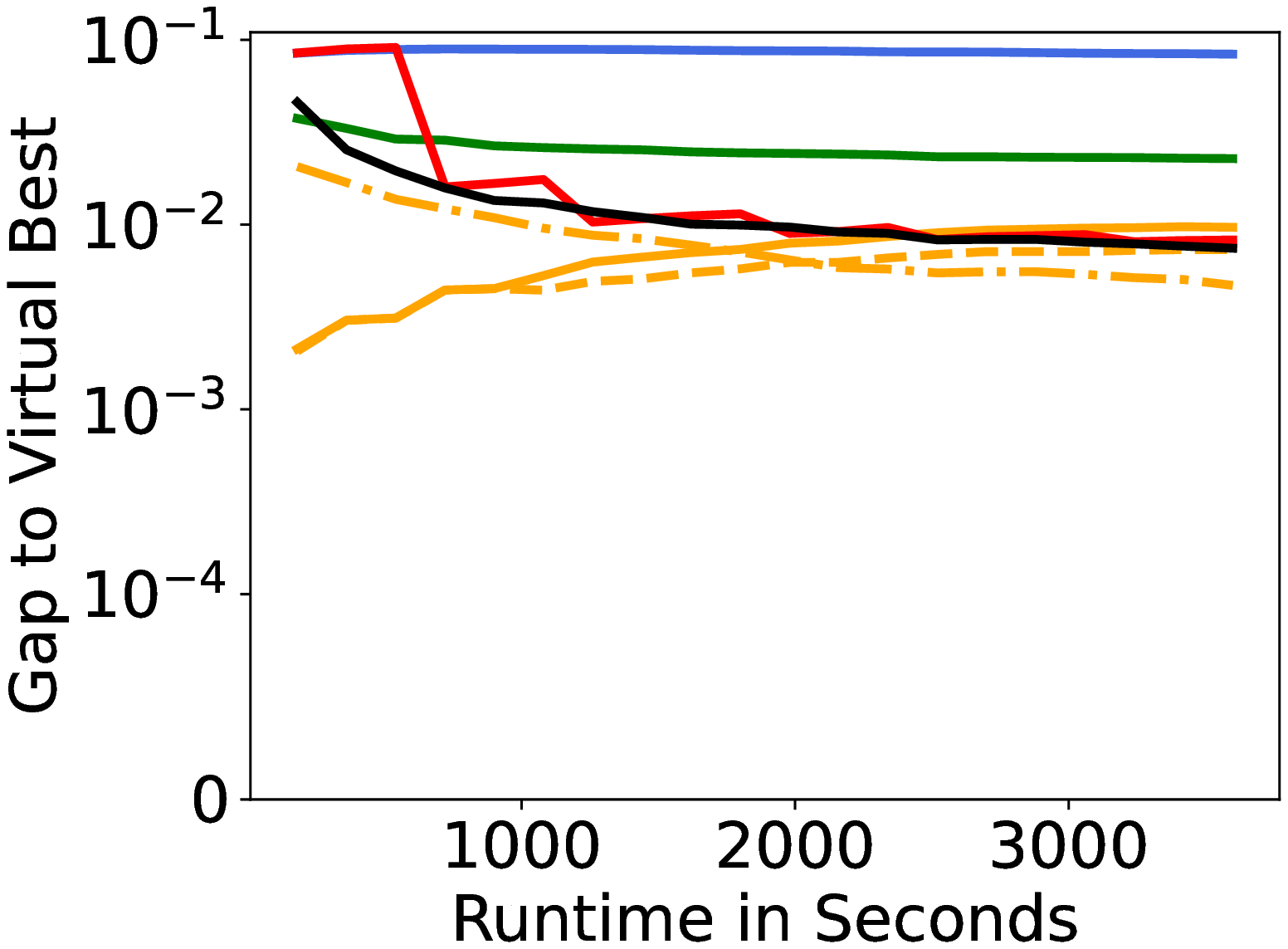}

			\caption{SC}
	\end{subfigure}
	\begin{subfigure}[htbp]{0.49\textwidth}
		\centering
		\includegraphics[height=4cm]{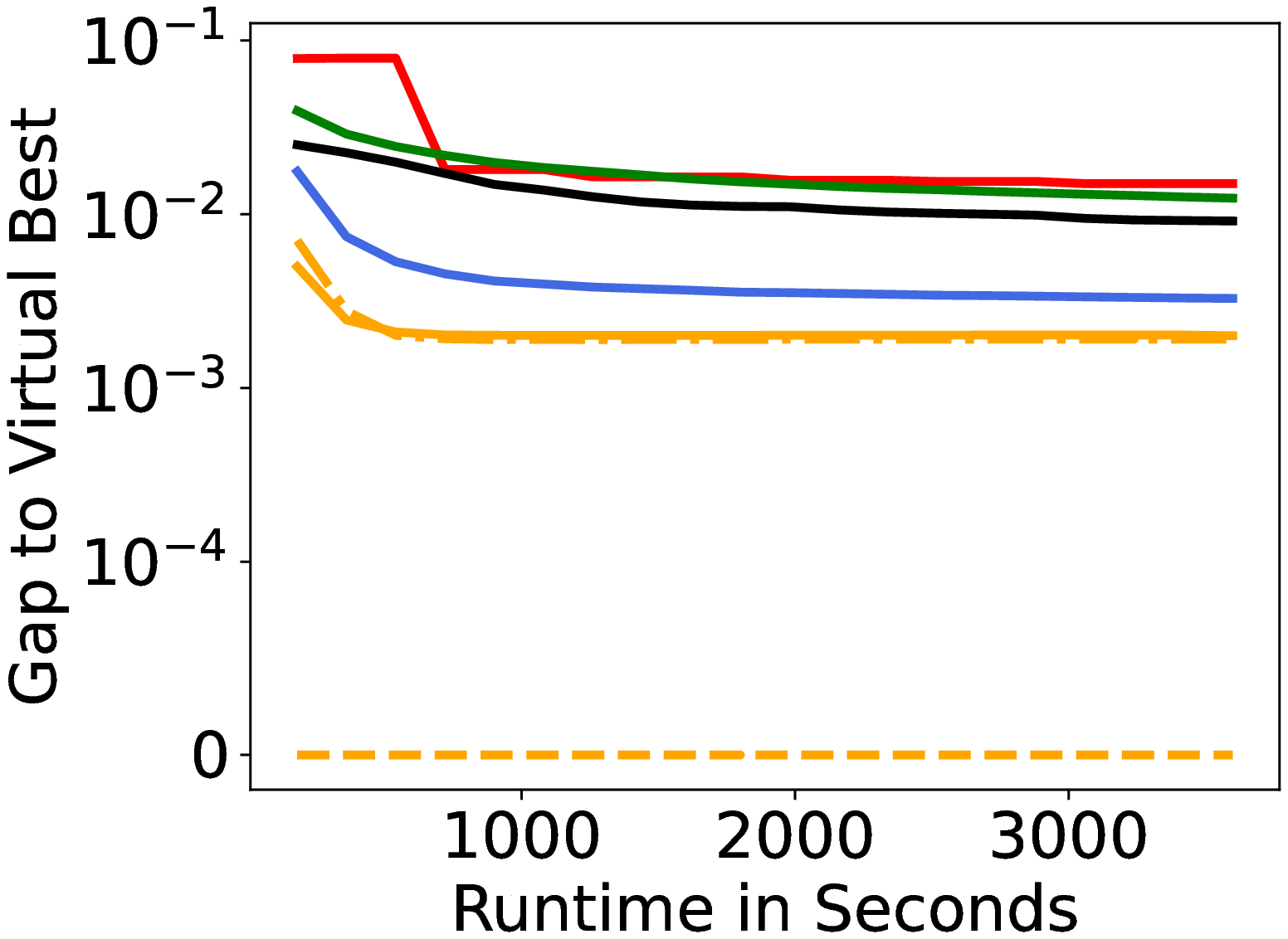}
			\caption{MK}
	\end{subfigure}
    \caption{Comparison with non-ML approaches: The gap to virtual best as a function of time, averaged over 100 instances. \label{res::virtualBest}}
\end{figure}

\begin{figure}[tbp]
    \centering
    \includegraphics[width=\textwidth]{figure/legend_timeVSobj.eps}
    \begin{subfigure}[htbp]{0.49\textwidth}
		\centering
		\includegraphics[height=4cm]{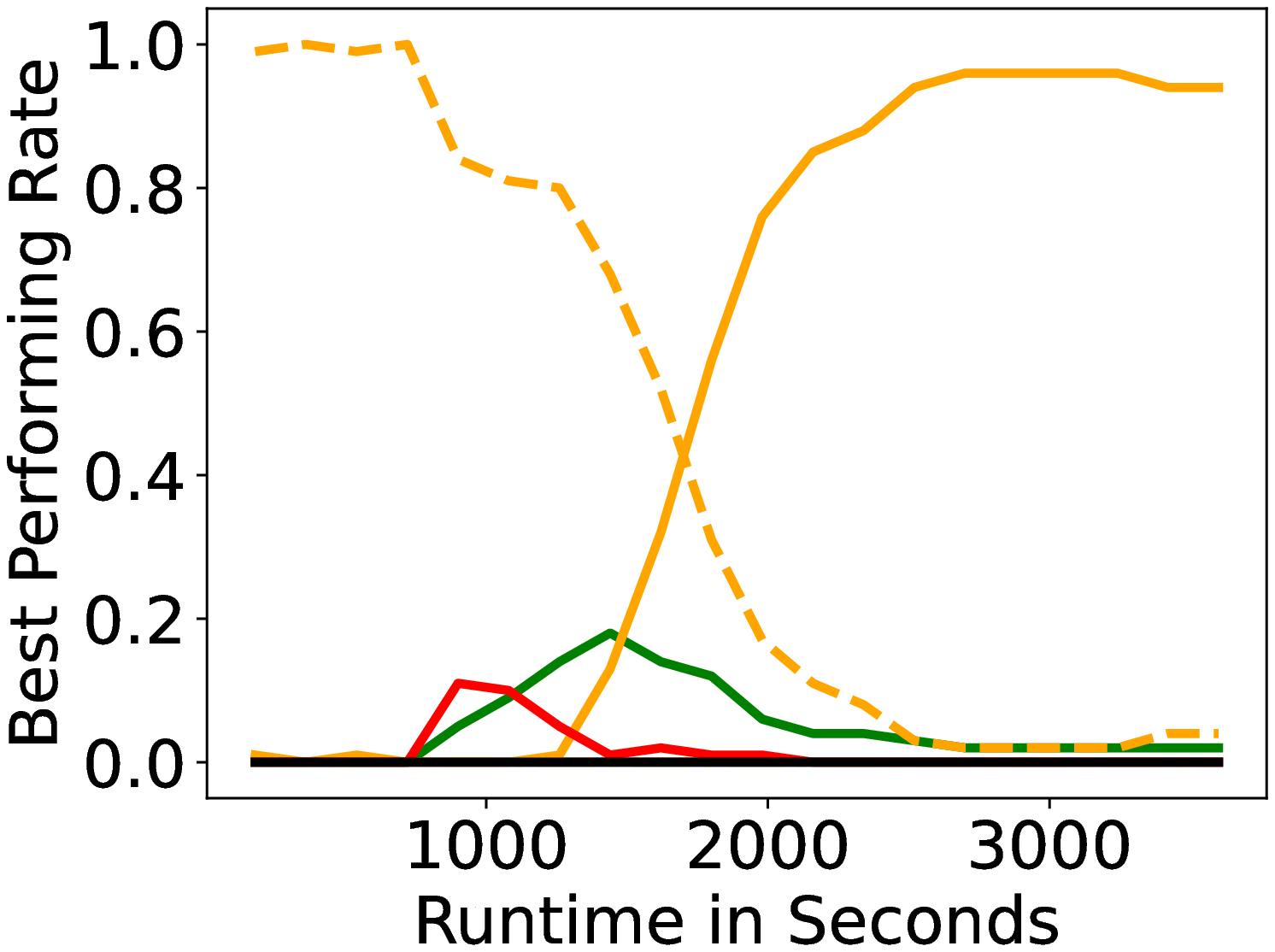}
		\caption{MVC\label{fig::exmp1}}
	\end{subfigure}\,\,
	\begin{subfigure}[htbp]{0.49\textwidth}
		\centering
		\includegraphics[height=4cm]{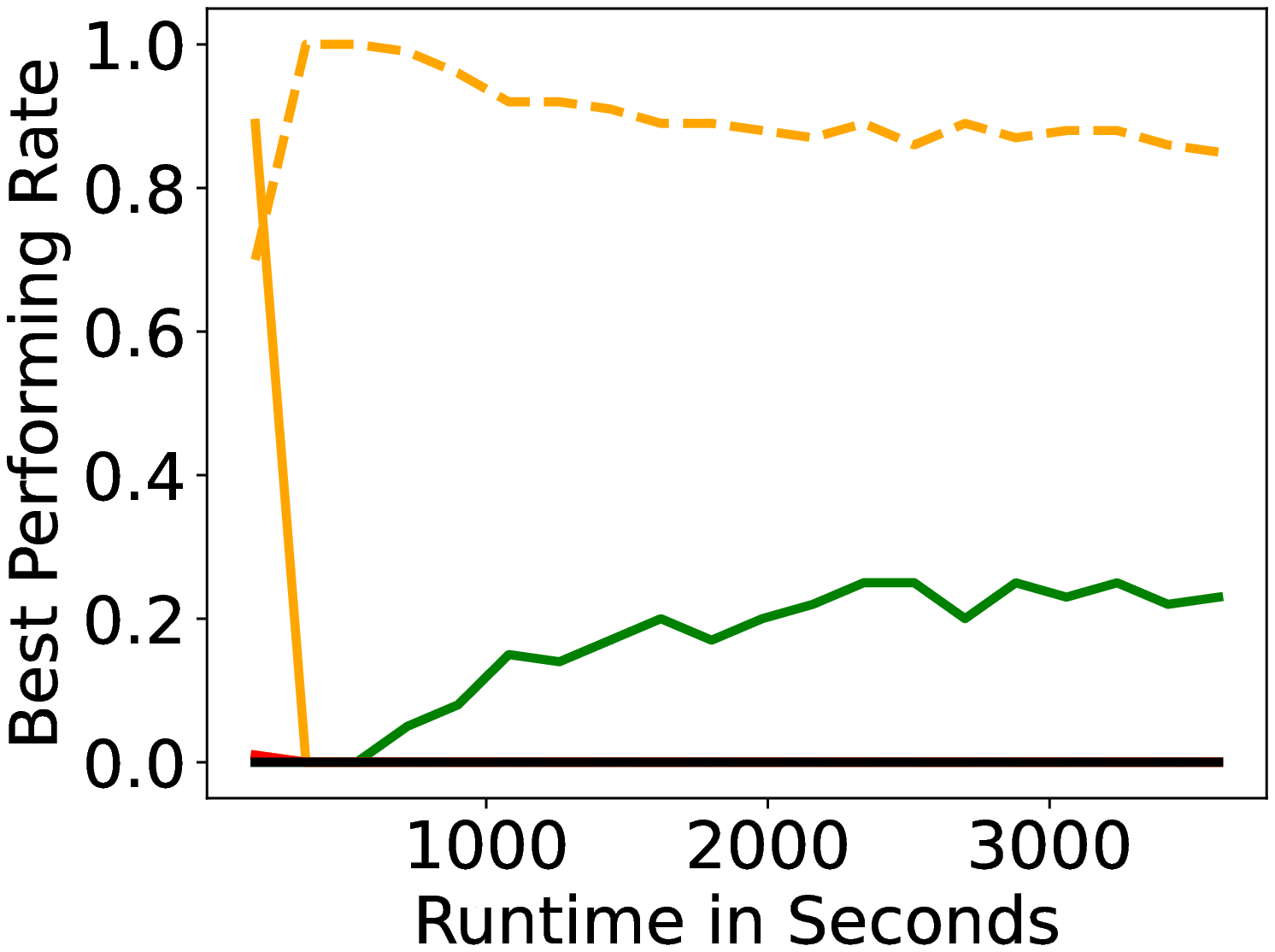}
			\caption{MIS\label{fig::exmp2}}
	\end{subfigure}
	\begin{subfigure}[htbp]{0.49\textwidth}
		\centering
		\includegraphics[height=4cm]{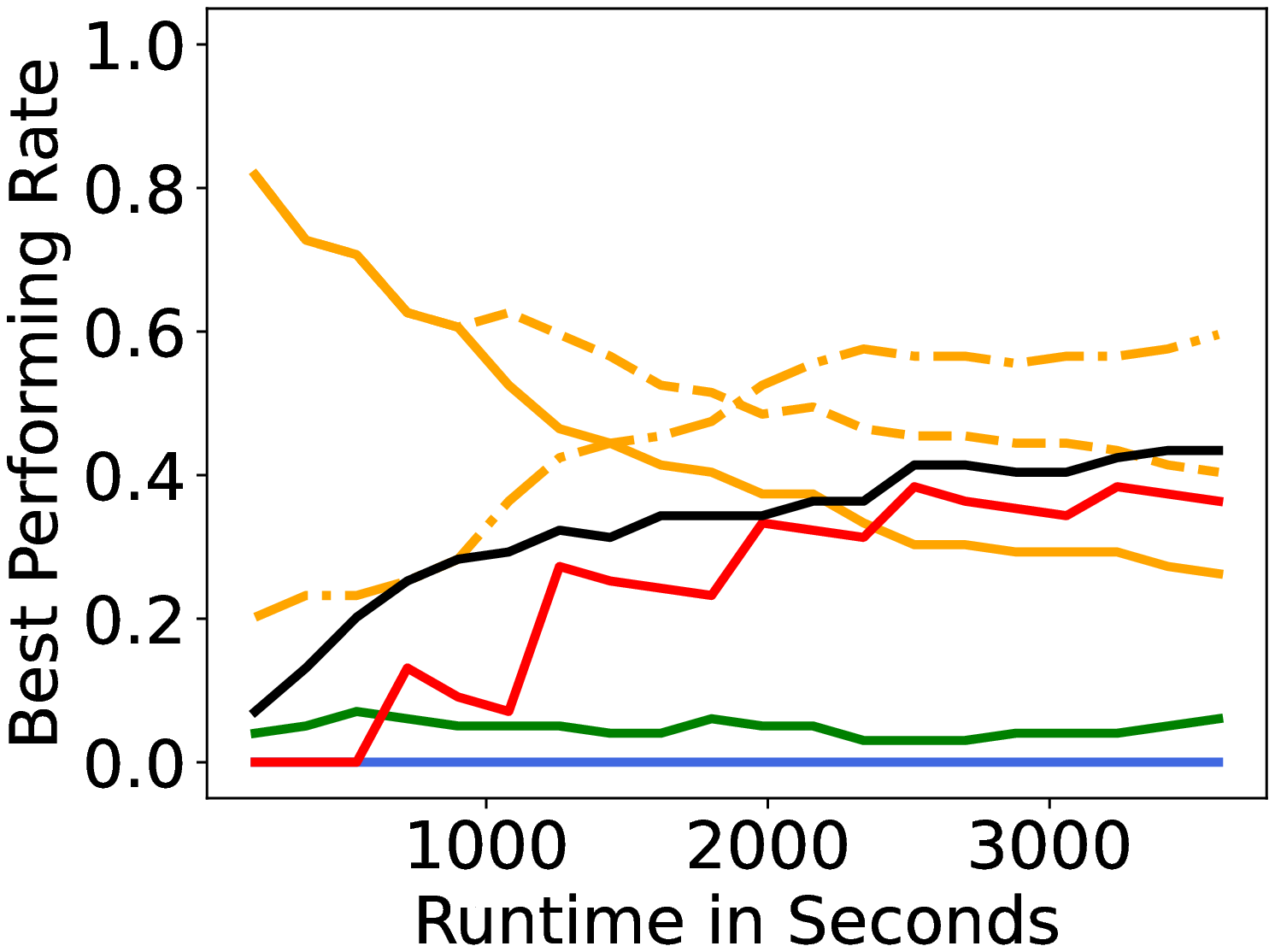}
			\caption{SC\label{fig::exmp3}}
	\end{subfigure}
	\begin{subfigure}[htbp]{0.49\textwidth}
		\centering
		\includegraphics[height=4cm]{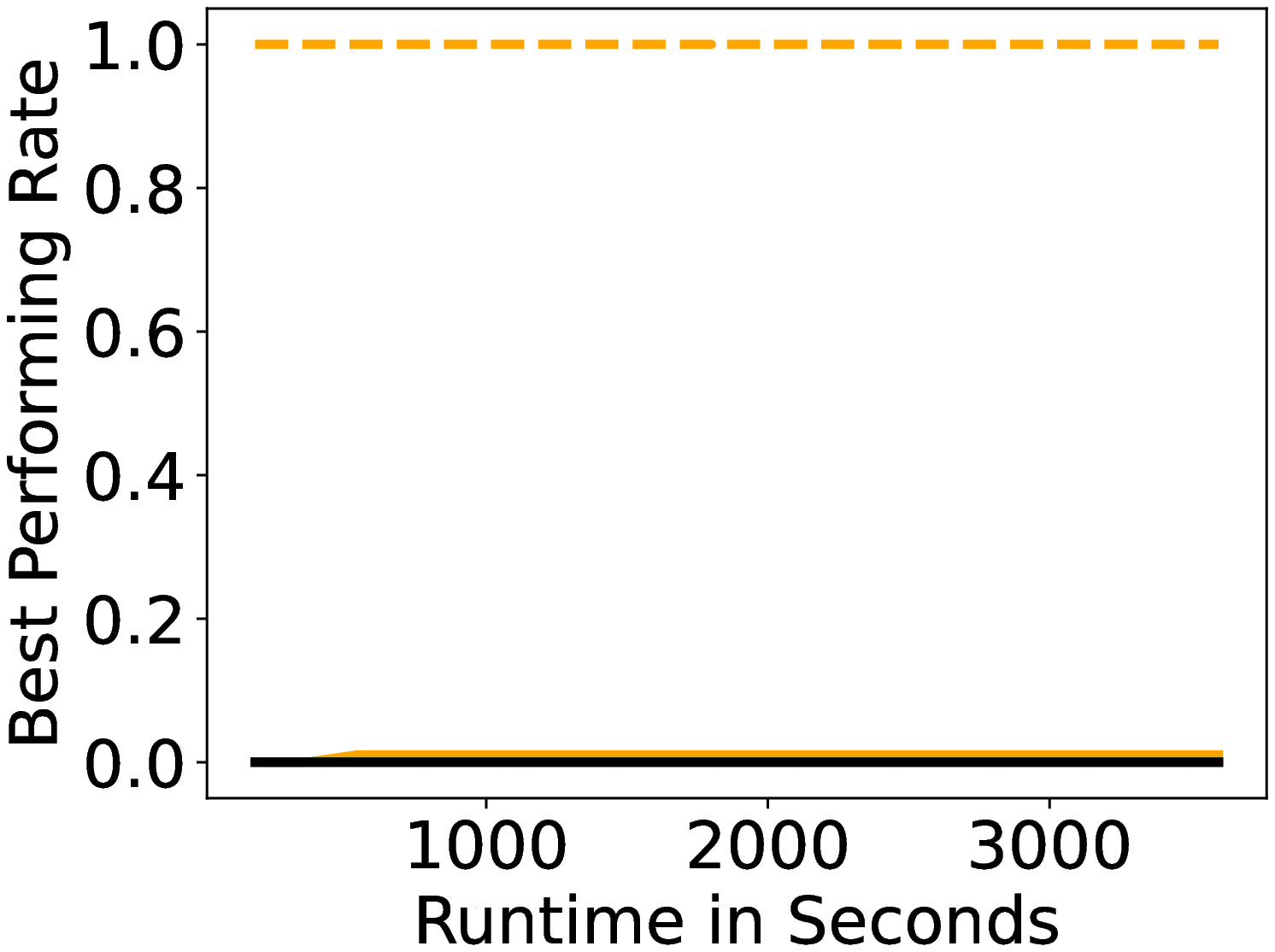}
			\caption{MK\label{fig::exmp4}}
	\end{subfigure}
    \caption{Comparison with non-ML approaches: The best performing rate as a function of time, calculated over 100 instances. The best performing rate of an approach  is the fraction of instances on which it has the best primal bound (including ties) at a given  time cutoff among all approaches. The sum of the best performing rates at a given time might sum up greater than 1 since ties are counted multiple times. \label{res::winrate}}
\end{figure}

\begin{figure}[htbp]
    \centering
    \includegraphics[width=\textwidth]{figure/legend_timeVSobj.eps}
    
	\begin{subfigure}[htbp]{1\textwidth}
		\centering
  \includegraphics[height=4cm]{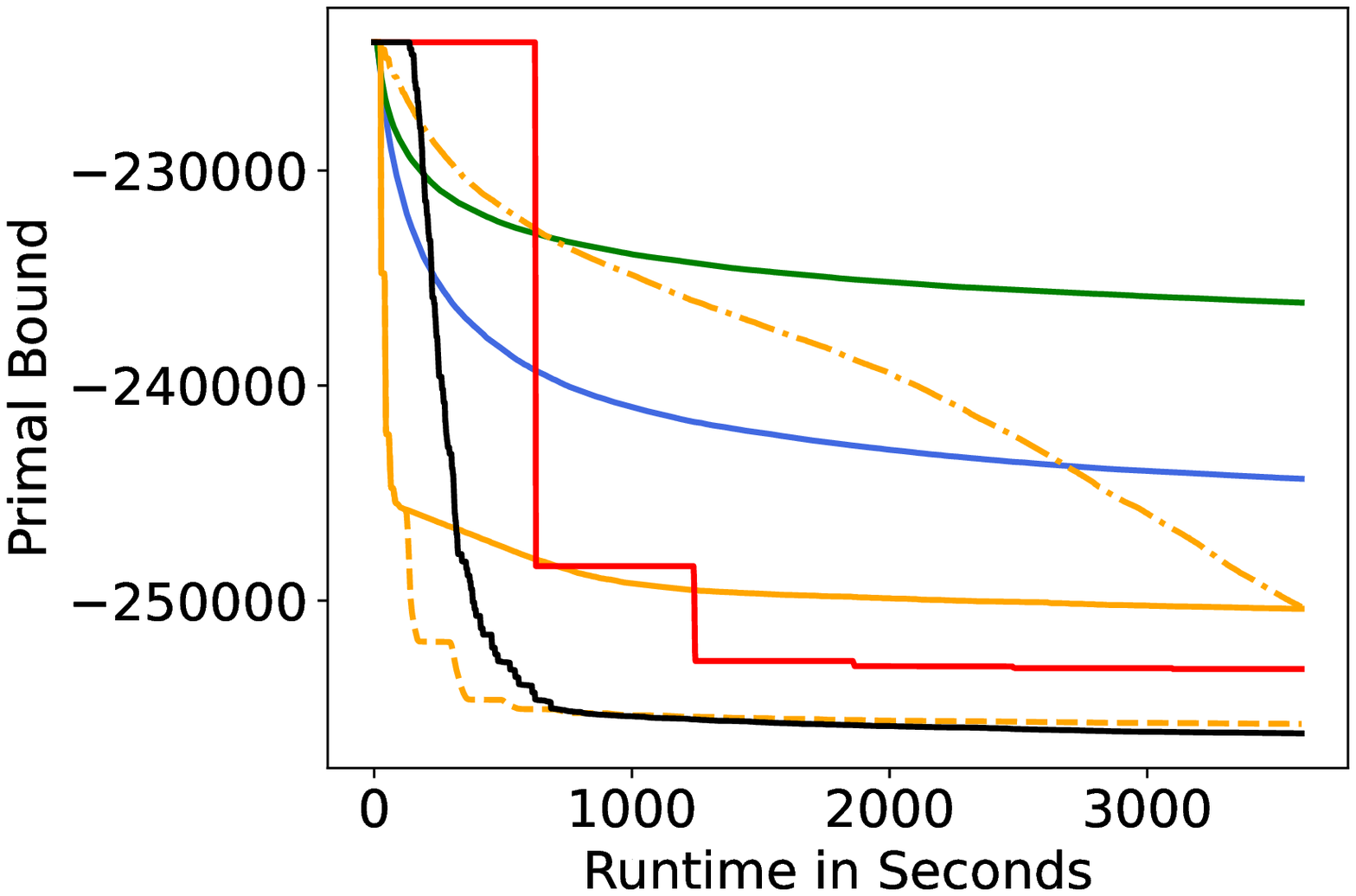}
		\includegraphics[height=4cm]{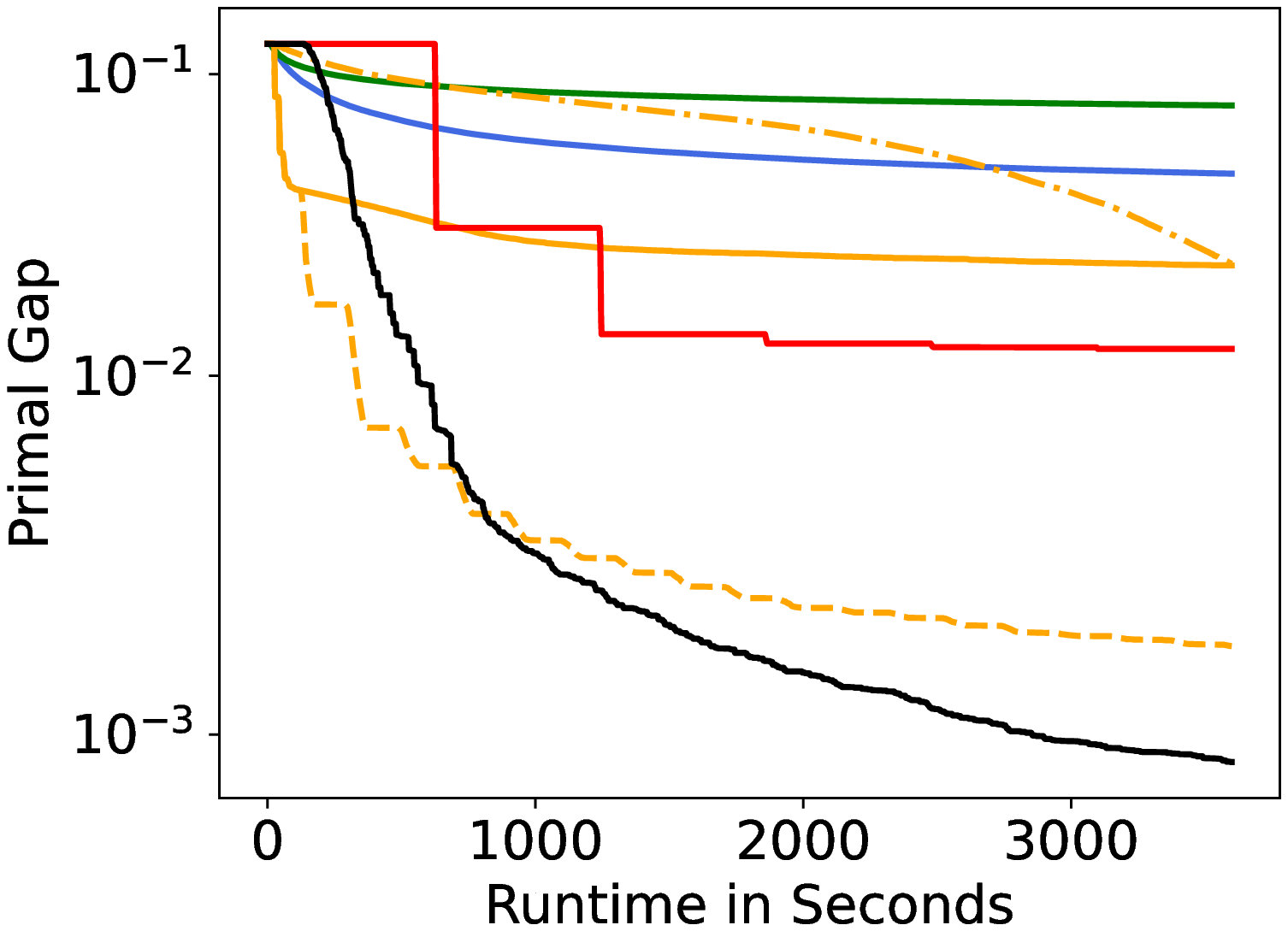}
		\includegraphics[height=4cm]{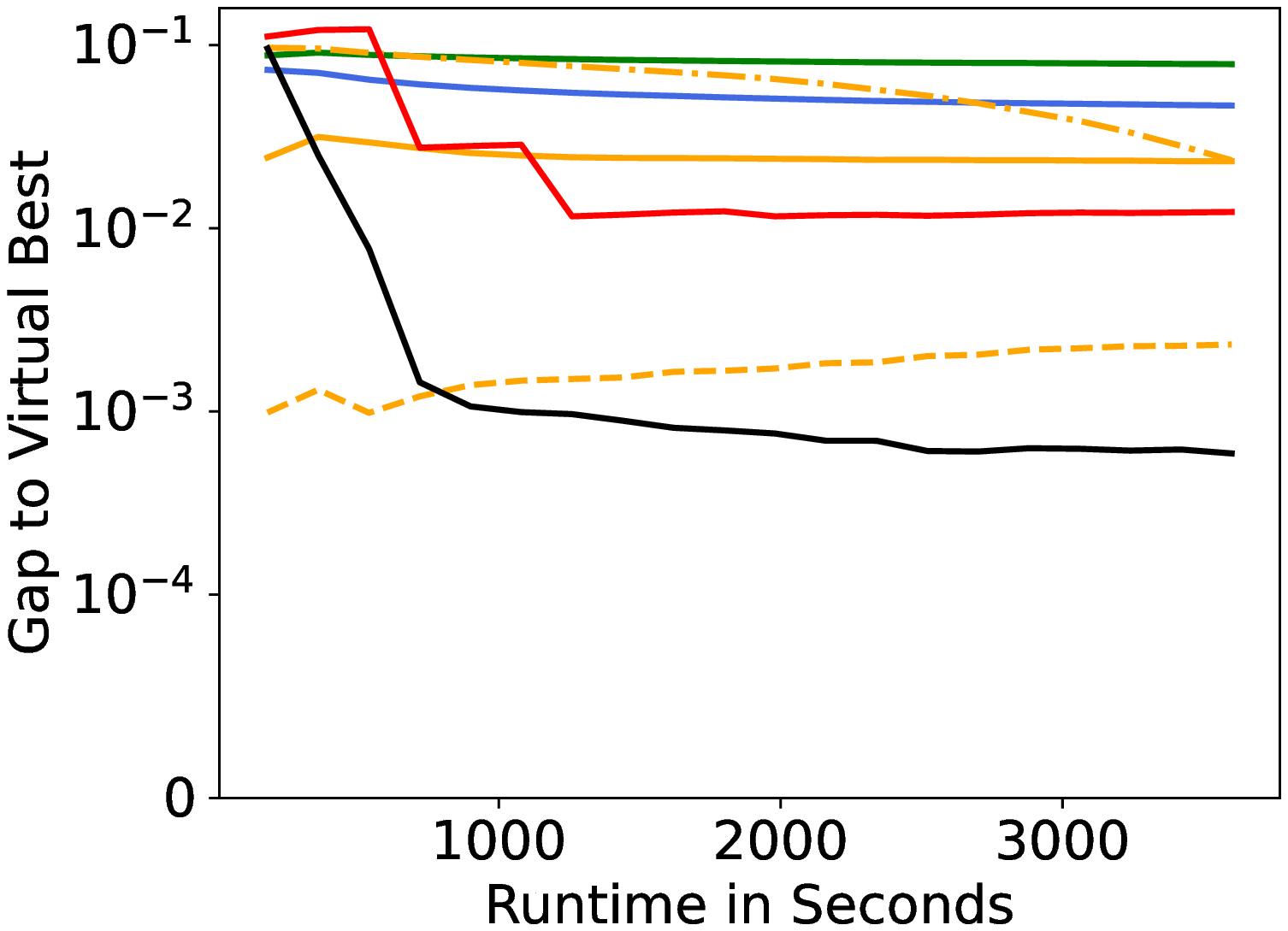}
  \includegraphics[height=4cm]{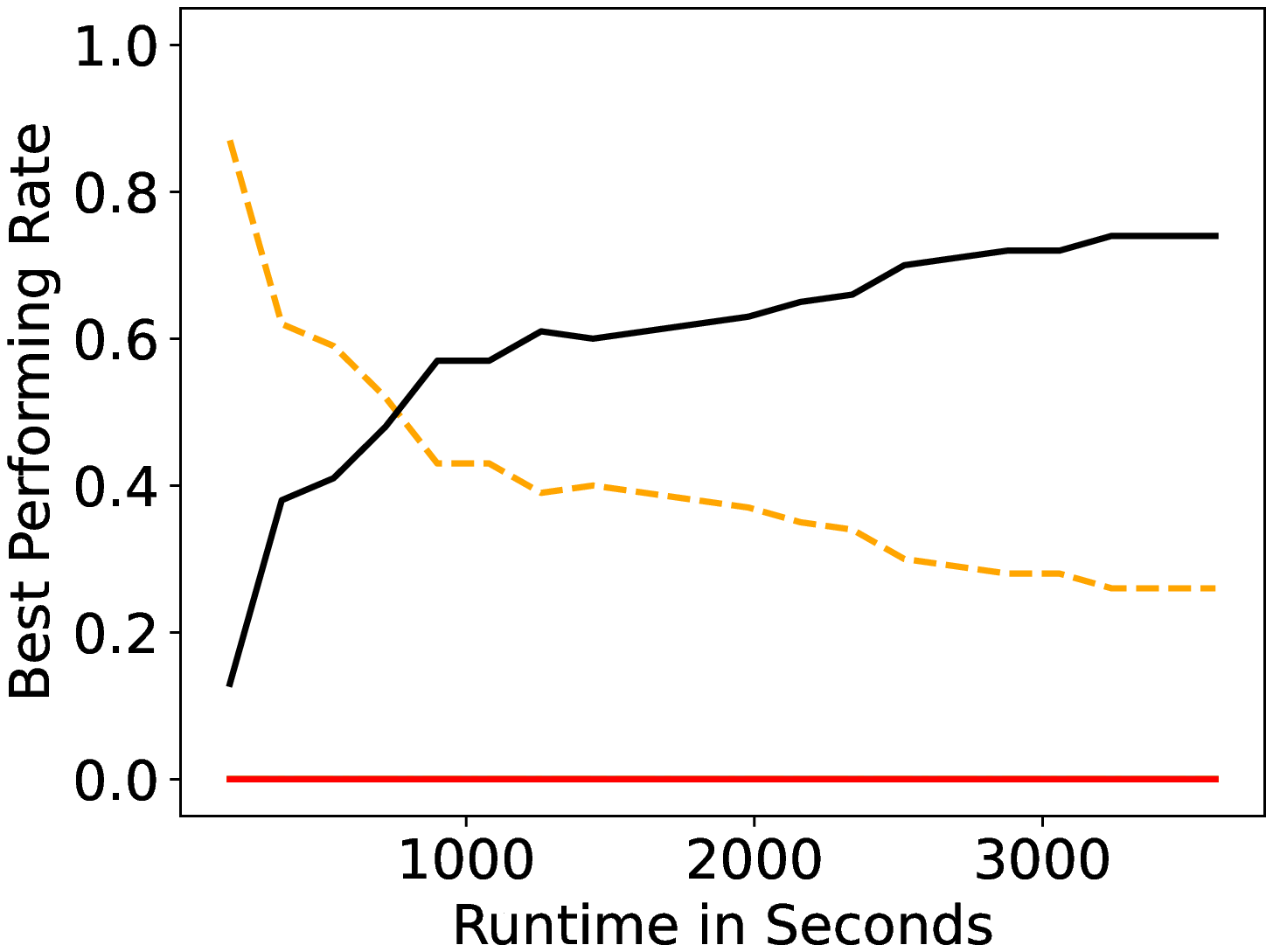}
	\end{subfigure}

    \caption{Results on CAT instances.\label{res::CAT}}
\end{figure}

\section{Details on Tuning Hyperparameters}

For $\beta$ which upper bounds the neighborhood size, we tried different values from $\{0.25,0.5,0.6,0.7\}$. $\beta=0.25$ is the worst for all approaches, resulting in the highest primal gap. For \LBRELAX and its variants, \DM and LB, all values perform similarly (because they select effective neighborhoods early in the search and their neighborhood sizes either do not reach the upper bounds or they already converge to good solutions before reaching it). For \RANDOM and \GRAPH, beta=0.5 is the best for them. So we set beta=0.5 consistently for all approaches.

For initial neighborhood sizes $k_0$, it is sensitive for approaches that need long runtime to select variables, such as \LBRELAX and its variants, LB and IL-LNS (they need to solve an LP relaxation, an LB ILP and run a forward pass of GCN in each iteration, respectively). They need the right $k_0$ from the beginning and we fine tune it for them. For \RANDOM and \GRAPH, their runtime for selecting variables is short, and with the adaptive neighborhood size mechanism, they could very quickly find the right neighborhood size and are insensitive to $k_0$. They converge to the same primal gaps ($<1\%$ relative differences) with similar primal integrals ($<2\%$ relative differences) using different $k_0$.

For $\alpha$ that controls the rate at which $k_t$ increases, we tried values from $\{1,1.01,1.02,1.05\}$. Overall, $\alpha$ does not have a big impact on the performance if $\alpha>1$, however $\alpha=1$ is far worse than the others.

For the runtime limit for repair operations, we tried 0.5, 1, 2 and 5 minutes. All approaches are not sensitive to it since most repairs are finished within 20 seconds. Except for \LBRELAX and \DM on SC instances, they both select neighborhoods that require a longer time to repair and a 2-minute runtime limit is necessary and best for them. We therefore use 2 minutes consistently.

\section{Additional Comparison with Non-ML Approaches}
We present additional results on comparison with non-ML approaches.

Figure \ref{res::bound} shows the primal gap as a function of time, averaged over 100 instances. 

Tables \ref{res::nonMLtable15},  \ref{res::nonMLtable30} and  \ref{res::nonMLtable45} present the average primal gap and primal integral at 15, 30 and 45 minutes time cutoff, respectively.

We also introduce another metric for comparison: the \textit{gap to virtual best} of an approach at time $q$. It is defined as the normalized difference between its best primal bound found up to time $q$ and the best primal bound found up to time $q$ by any approach in the portfolio for comparison, similar to the primal gap. Figure \ref{res::virtualBest} shows the gap to virtual best as a function of time, averaged over 100 instances.

Figure \ref{res::winrate} shows the winning rate as a function of time, averaged over 100 instances. The best performing rate at time $q$ for an approach is  the fraction of instances on which it achieves the best performance (including ties) compared to all approaches in the portfolio.

\begin{figure}[tbp]
    \centering
    \includegraphics[width=\textwidth]{figure/legend_ML_timeVSobj.eps}
    \begin{subfigure}[htbp]{0.325\textwidth}
		\centering
		\includegraphics[height=3.cm]{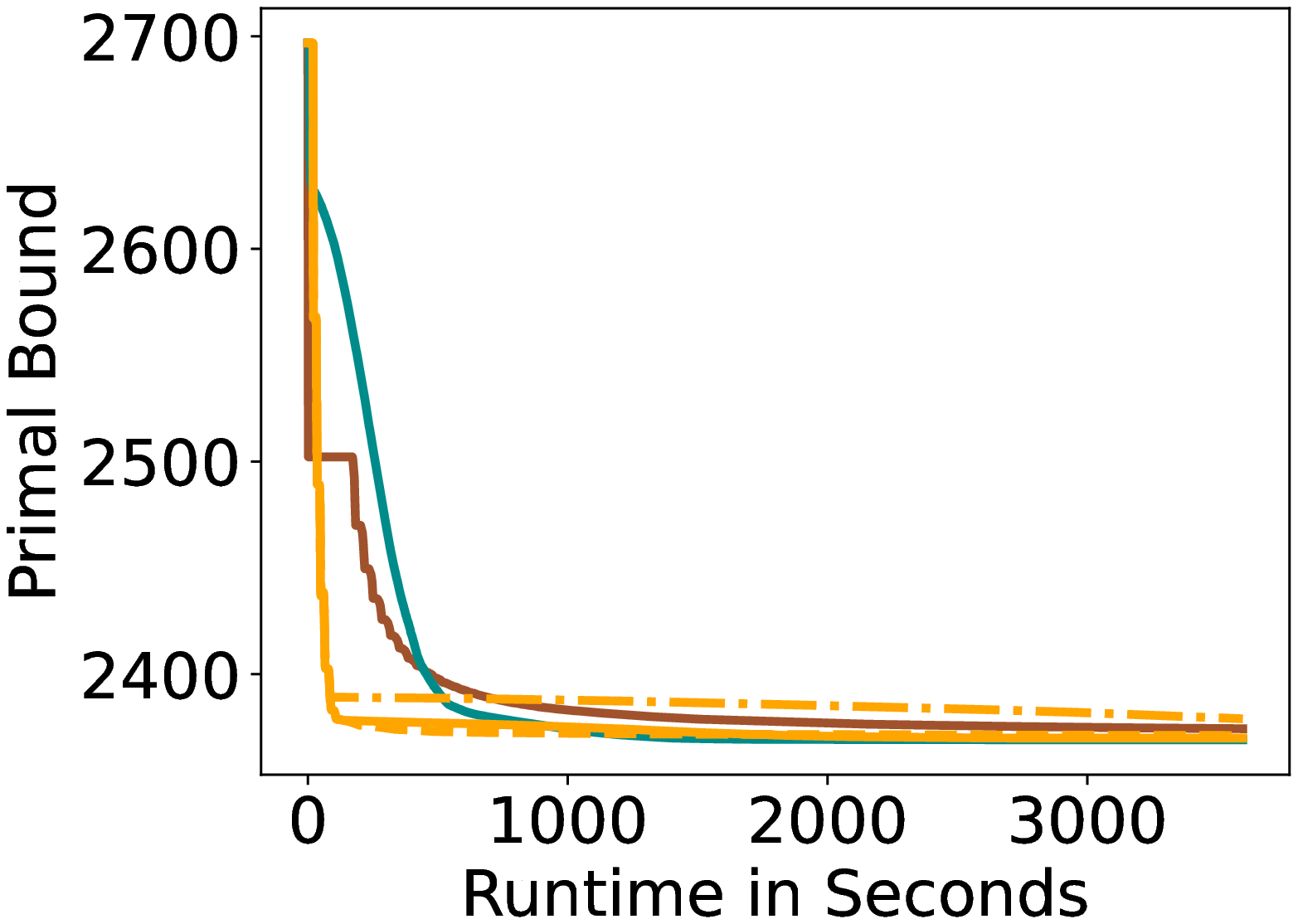}
		\caption{MVC}
	\end{subfigure}
	\begin{subfigure}[htbp]{0.325\textwidth}
		\centering
		\includegraphics[height=3.cm]{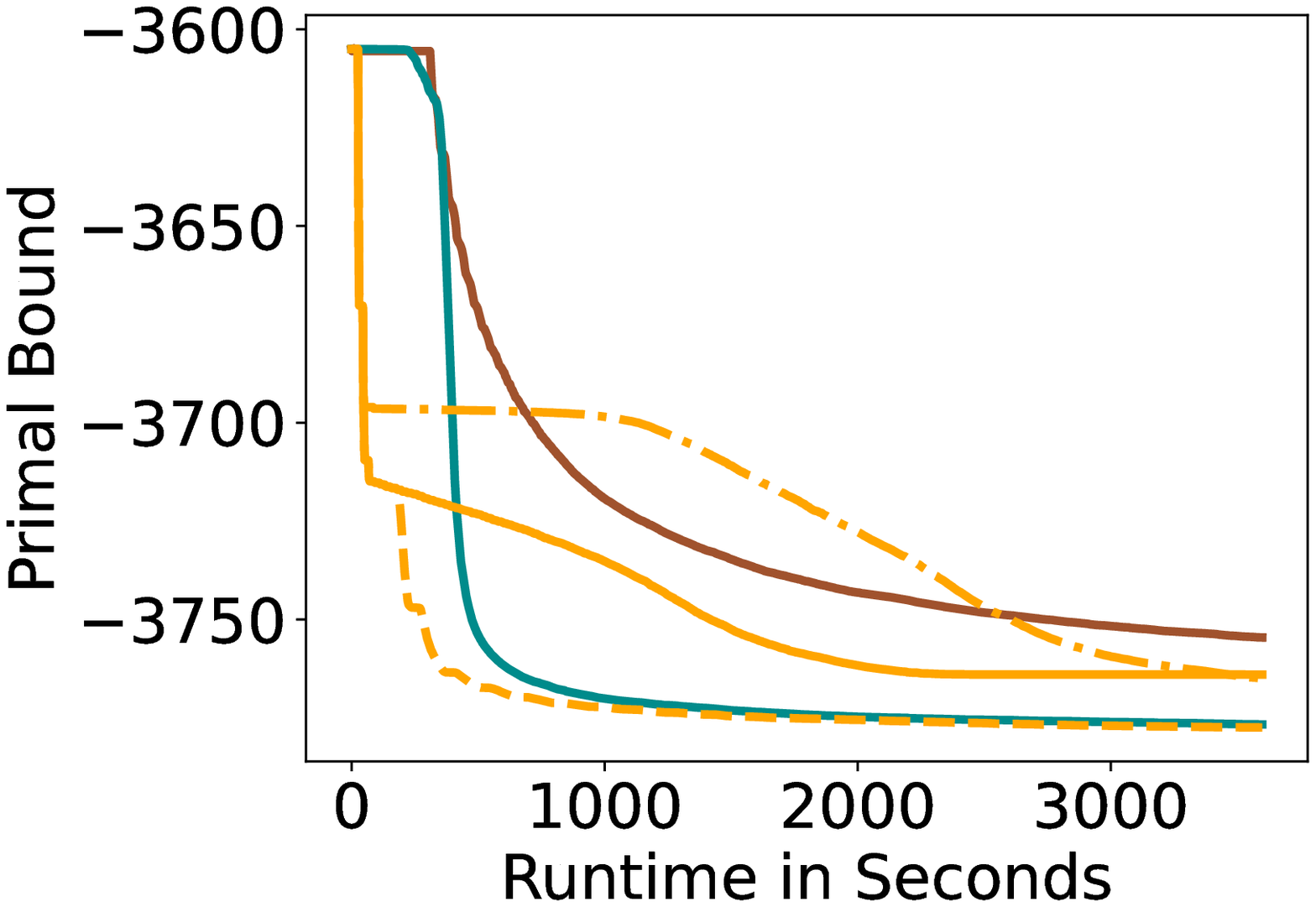}
			\caption{MIS}
	\end{subfigure}
	\begin{subfigure}[htbp]{0.325\textwidth}
		\centering
		\includegraphics[height=3.cm]{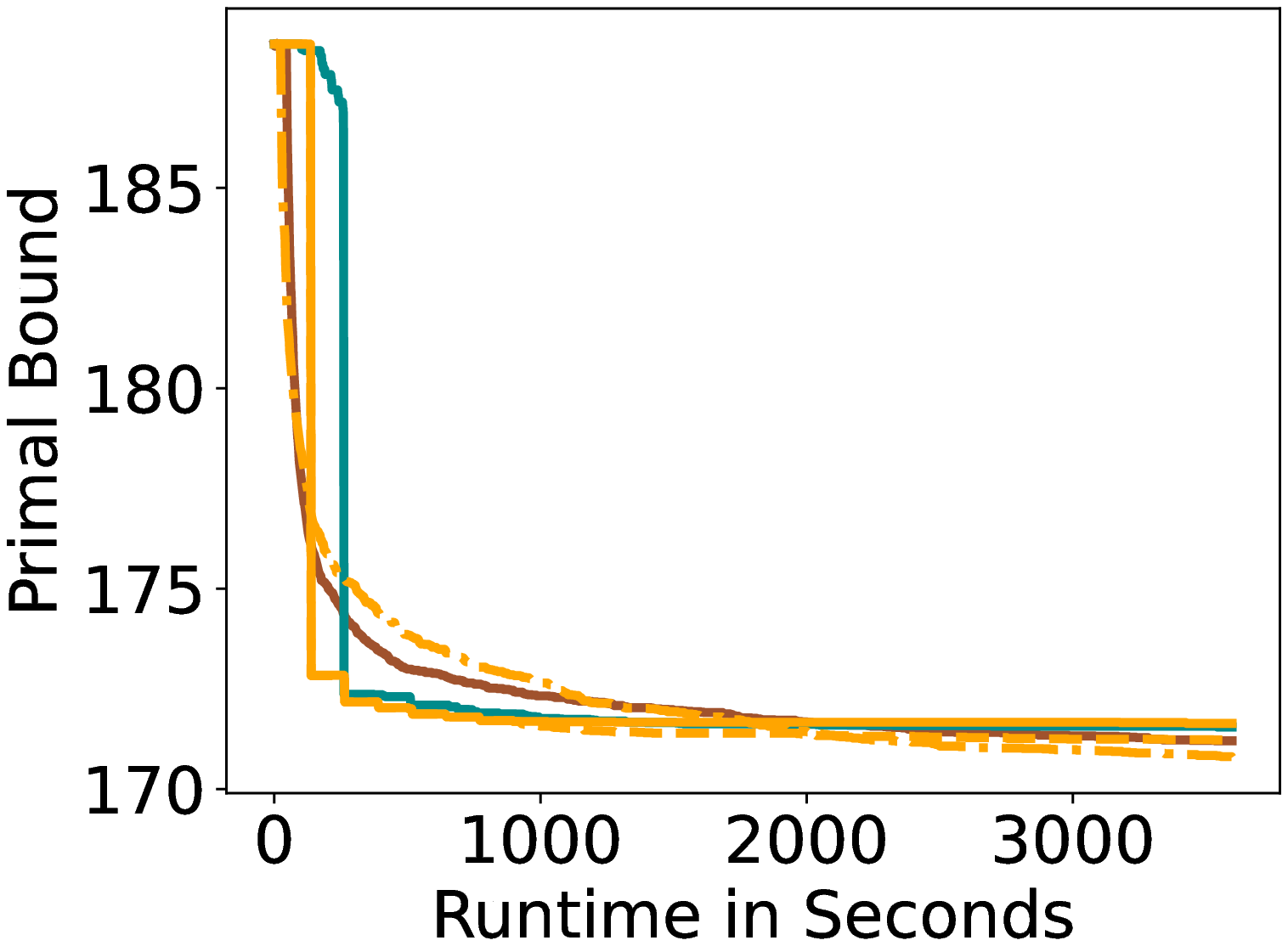}
			\caption{SC}
	\end{subfigure}

    \caption{Comparison with ML approaches: The primal bound as a function of time, averaged over 100 instances.\label{res::MLbound}}
\end{figure}

\begin{figure}[tbp]
    \centering
    \includegraphics[width=\textwidth]{figure/legend_ML_timeVSobj.eps}
    \begin{subfigure}[htbp]{0.325\textwidth}
		\centering
		\includegraphics[height=3.cm]{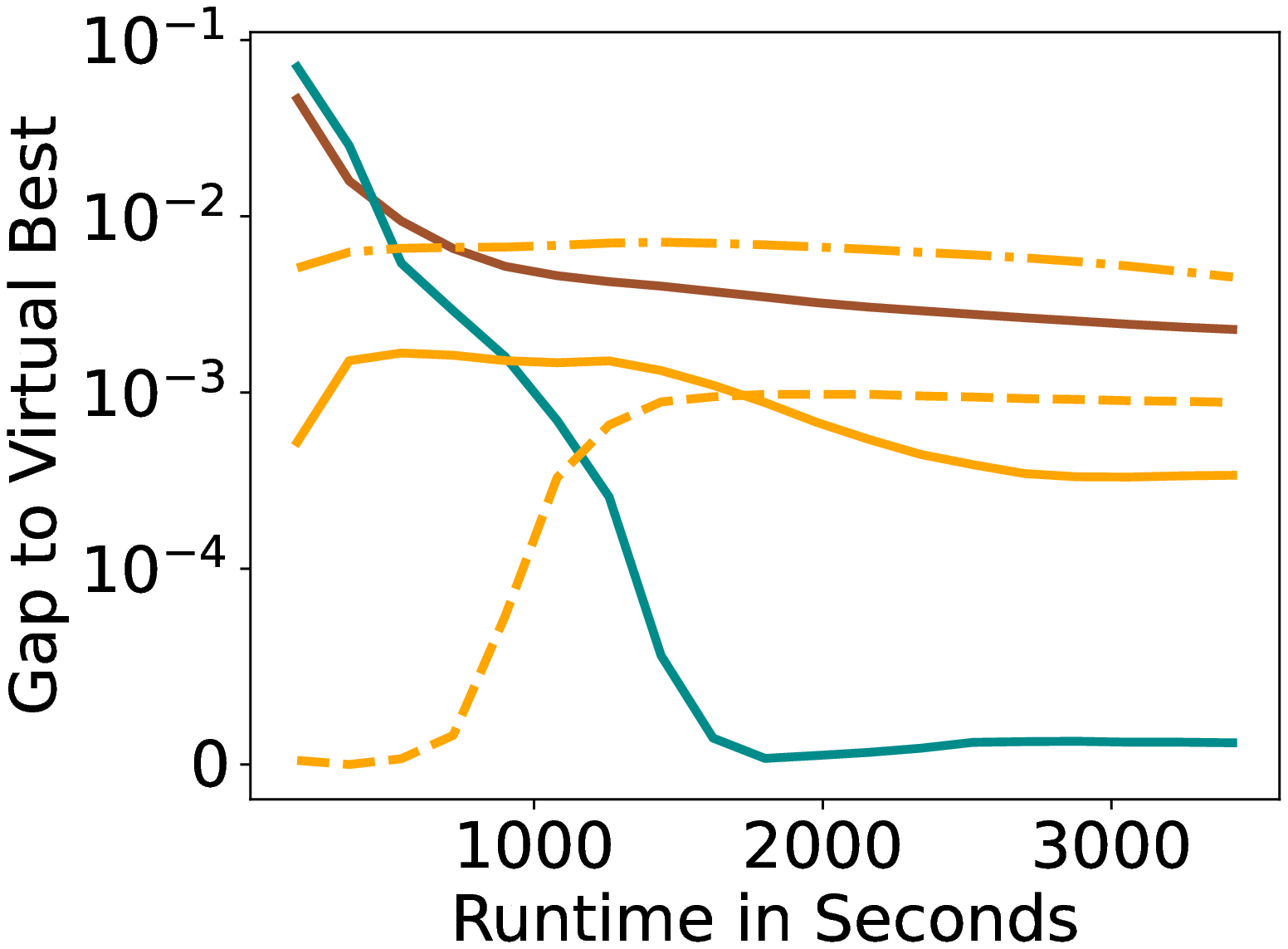}
		\caption{MVC}
	\end{subfigure}
	\begin{subfigure}[htbp]{0.325\textwidth}
		\centering
		\includegraphics[height=3.cm]{figure/ML_timeVSgap_INDSET2_BA_9000_3599_nh_200.eps}
			\caption{MIS}
	\end{subfigure}
	\begin{subfigure}[htbp]{0.325\textwidth}
		\centering
		\includegraphics[height=3.cm]{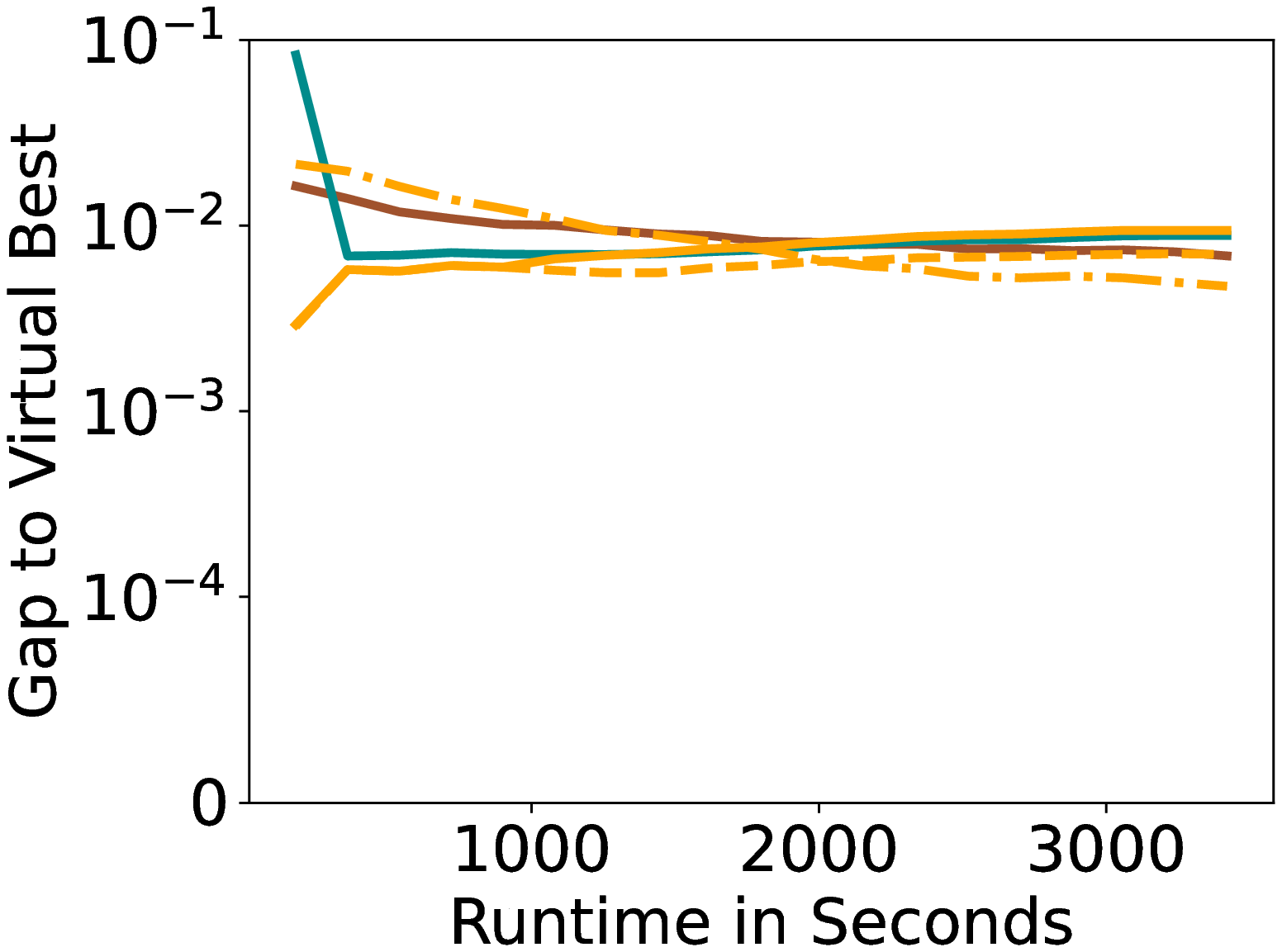}
			\caption{SC}
	\end{subfigure}

    \caption{Comparison with ML approaches: The gap to virtual best as a function of time, averaged over 100 instances.\label{res::MLvirtualBest}}
\end{figure}

\begin{figure}[tbp]
    \centering
    \includegraphics[width=\textwidth]{figure/legend_ML_timeVSobj.eps}
    \begin{subfigure}[htbp]{0.325\textwidth}
		\centering
		\includegraphics[height=3.cm]{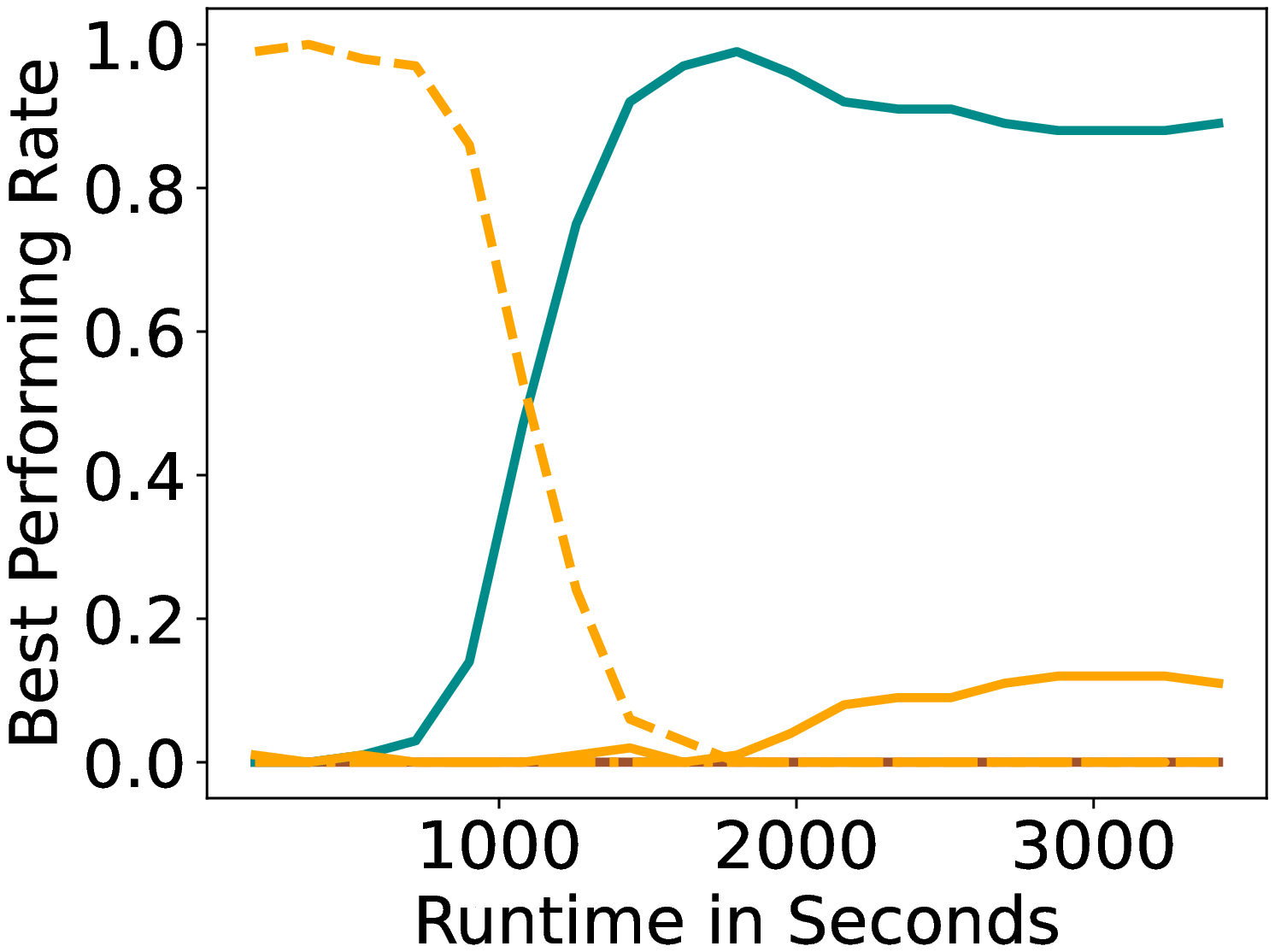}
		\caption{MVC}
	\end{subfigure}
	\begin{subfigure}[htbp]{0.325\textwidth}
		\centering
		\includegraphics[height=3.cm]{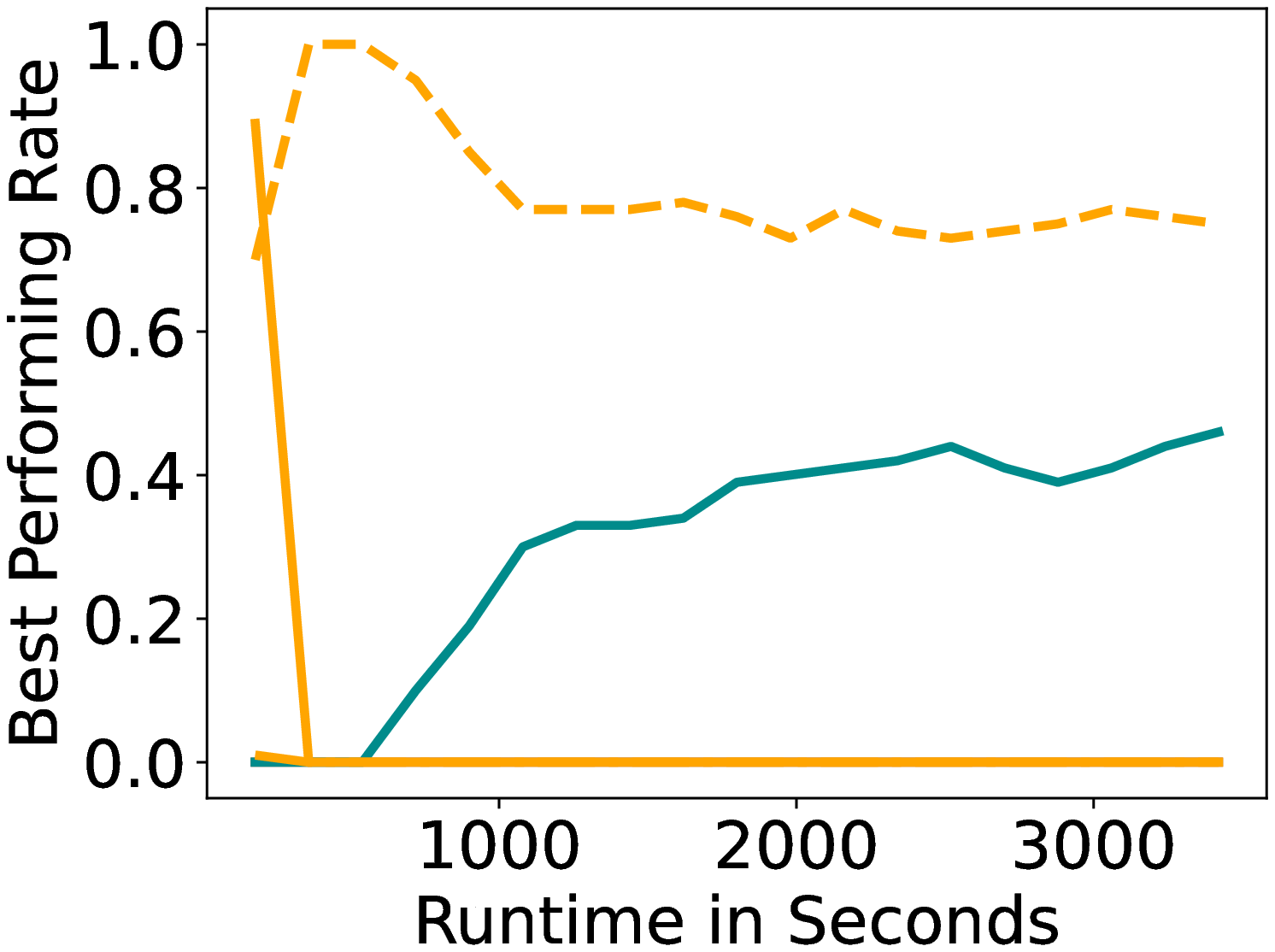}
			\caption{MIS}
	\end{subfigure}
	\begin{subfigure}[htbp]{0.325\textwidth}
		\centering
		\includegraphics[height=3.cm]{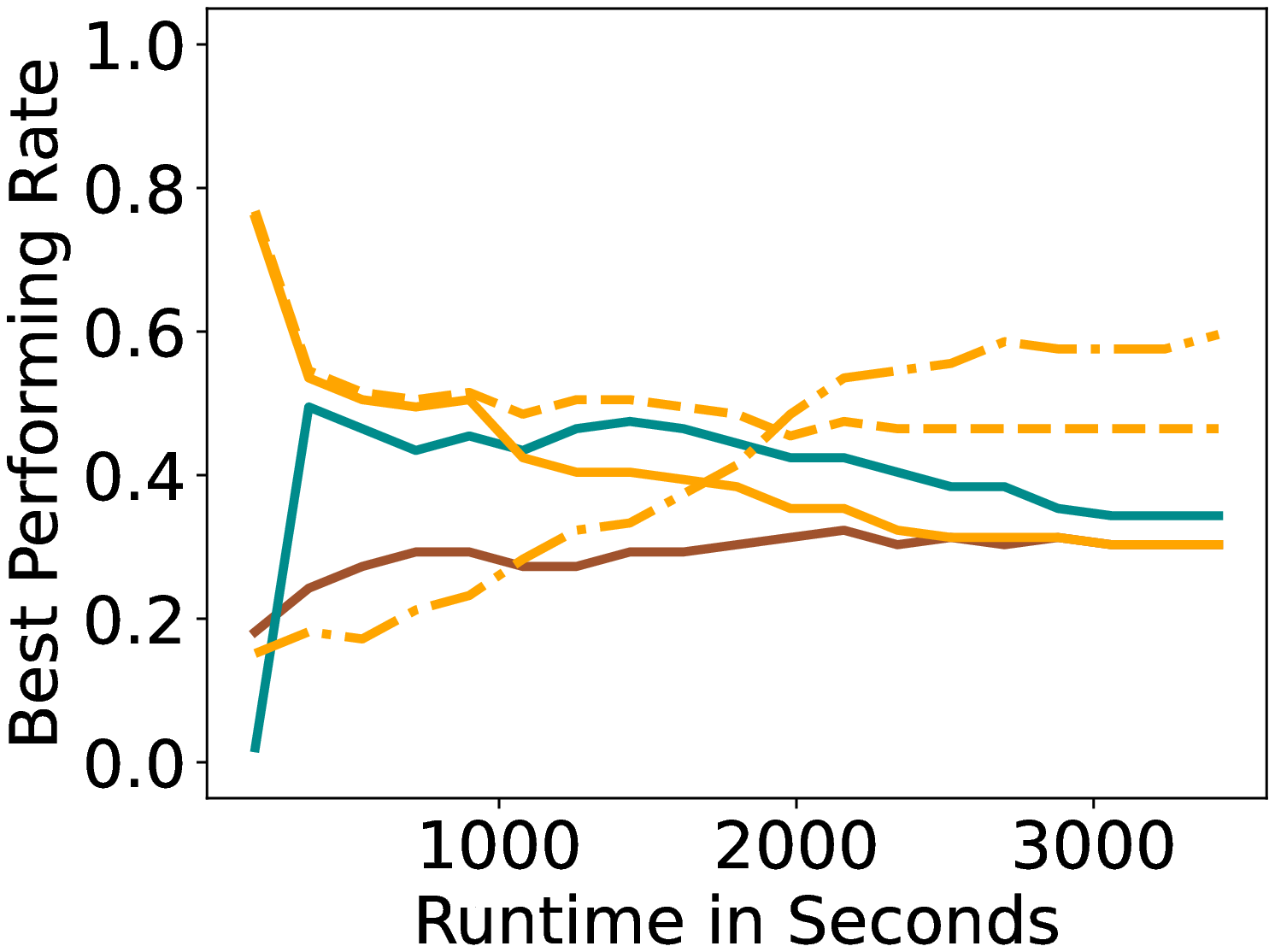}
			\caption{SC}
	\end{subfigure}

    \caption{Comparison with ML approaches: The gap to virtual best as a function of time, averaged over 100 instances.\label{res::MLwinRate}}
\end{figure}

\section{Experimental Results on CAT Instances}
We generate 100 combinatorial auction (CAT) instances with 4000 items and 8,000 bids according to arbitrary relationships in \cite{leyton2000towards}. The main results are shown in Figure \ref{res::CAT}. We show that \LBRELAX can improve the primal bound fast early in the search but get stuck at local minima and afterwards get outperformed  by BnB and LB. \LBRELAXRR can help escape the local minima but on average converges to slightly worse solutions than BnB.

\section{Additional Comparison with ML Approaches}

We present additional results on comparison with ML approaches.
Figure \ref{res::MLbound} shows the primal gap as a function of time, averaged over 100 instances. 
Figure \ref{res::MLvirtualBest} shows the gap to virtual best as a function of time, averaged over 100 instances. 
Figure \ref{res::MLwinRate} shows the winning rate as a function of time, averaged over 100 instances. 

\section{Additional Results on Selected MIPLIB Instances}
\begin{figure}[tbp]


    \centering
    \includegraphics[width=\textwidth]{figure/legend_timeVSobj.eps}
    \begin{subfigure}[htbp]{0.49\textwidth}
		\centering
		\includegraphics[height=3.6cm]{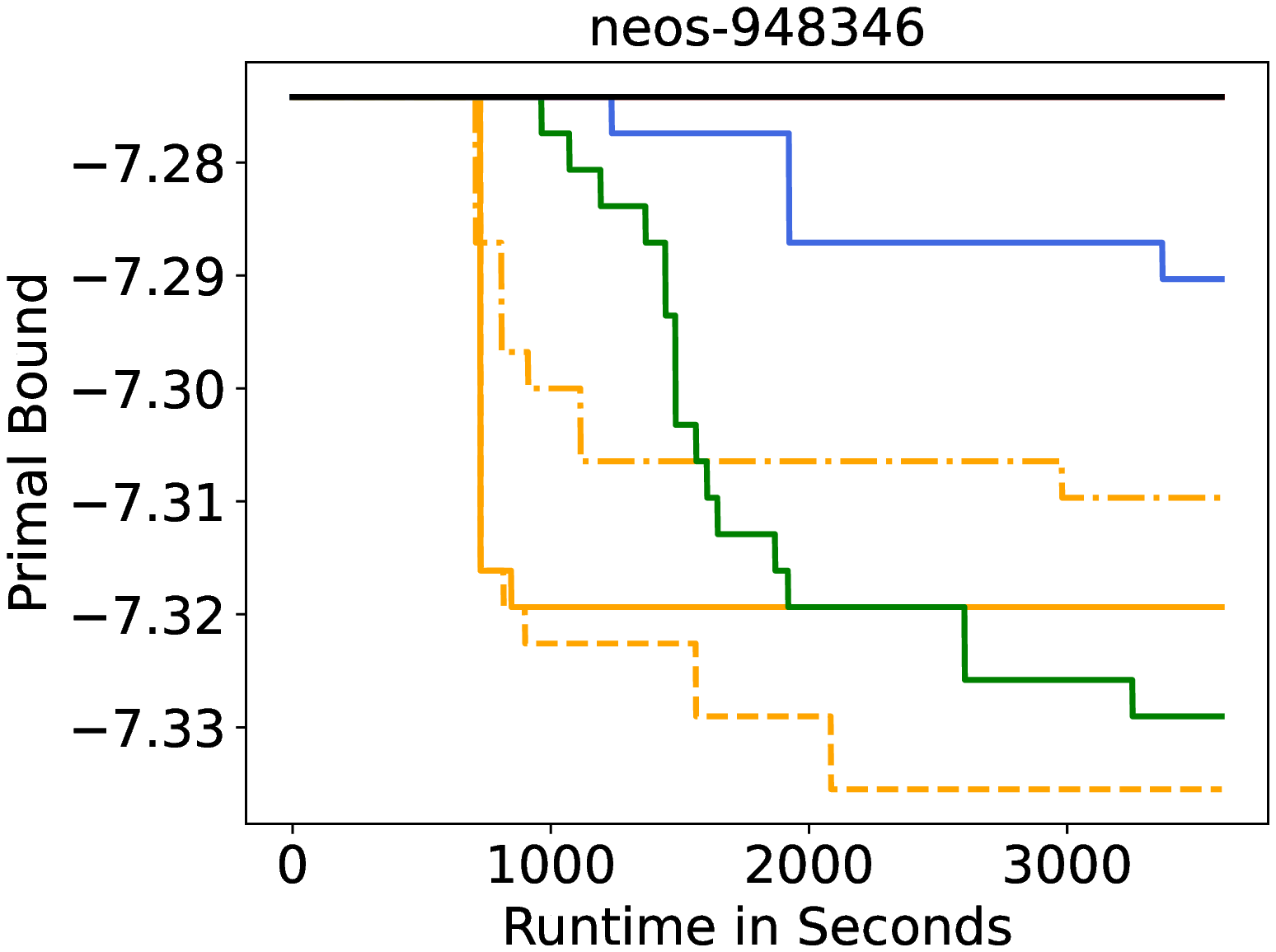}
	\end{subfigure}\,\,
	\begin{subfigure}[htbp]{0.49\textwidth}
		\centering
		\includegraphics[height=3.6cm]{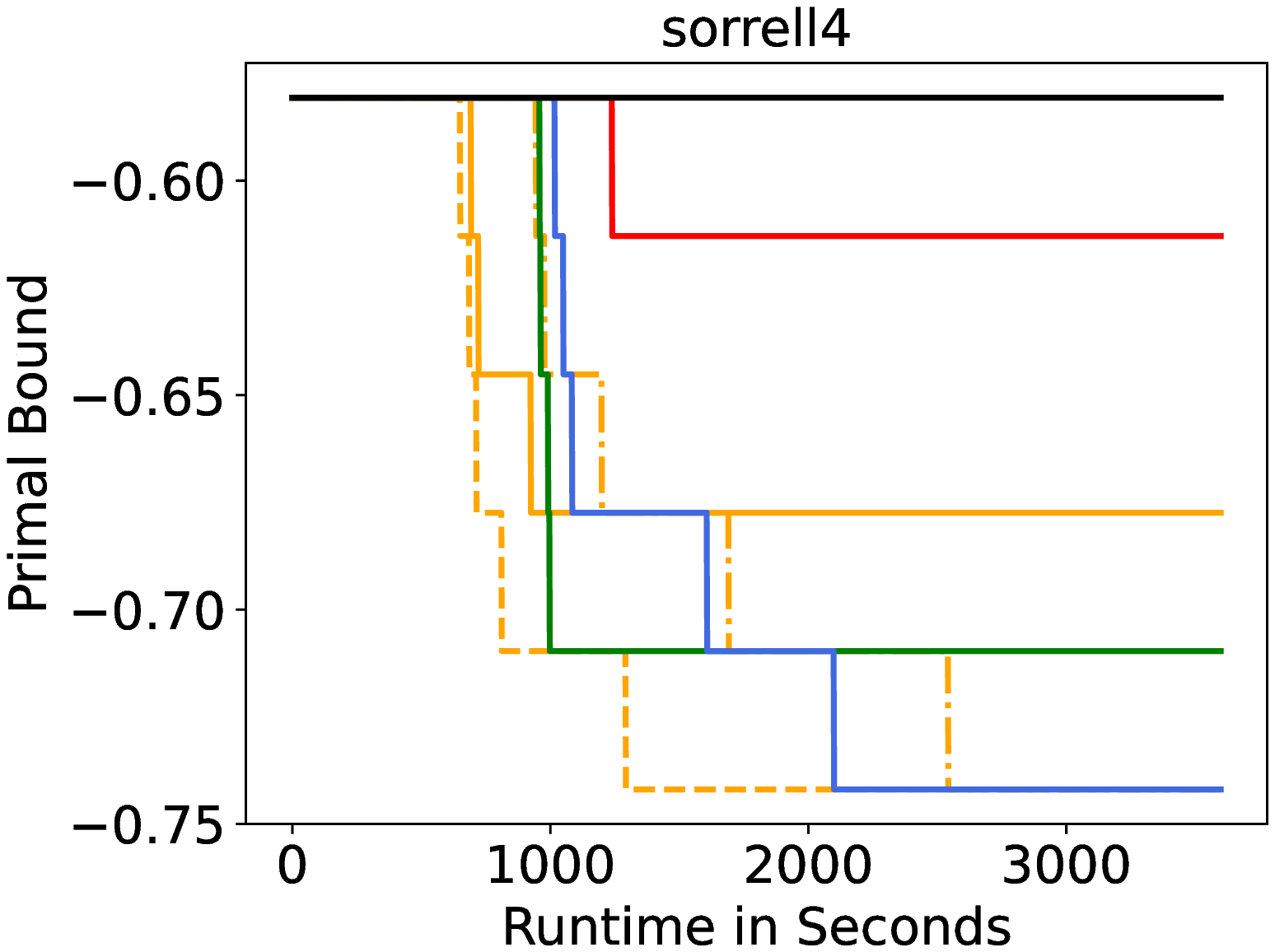}
	\end{subfigure}
	\begin{subfigure}[htbp]{0.49\textwidth}
		\centering
		\includegraphics[height=3.6cm]{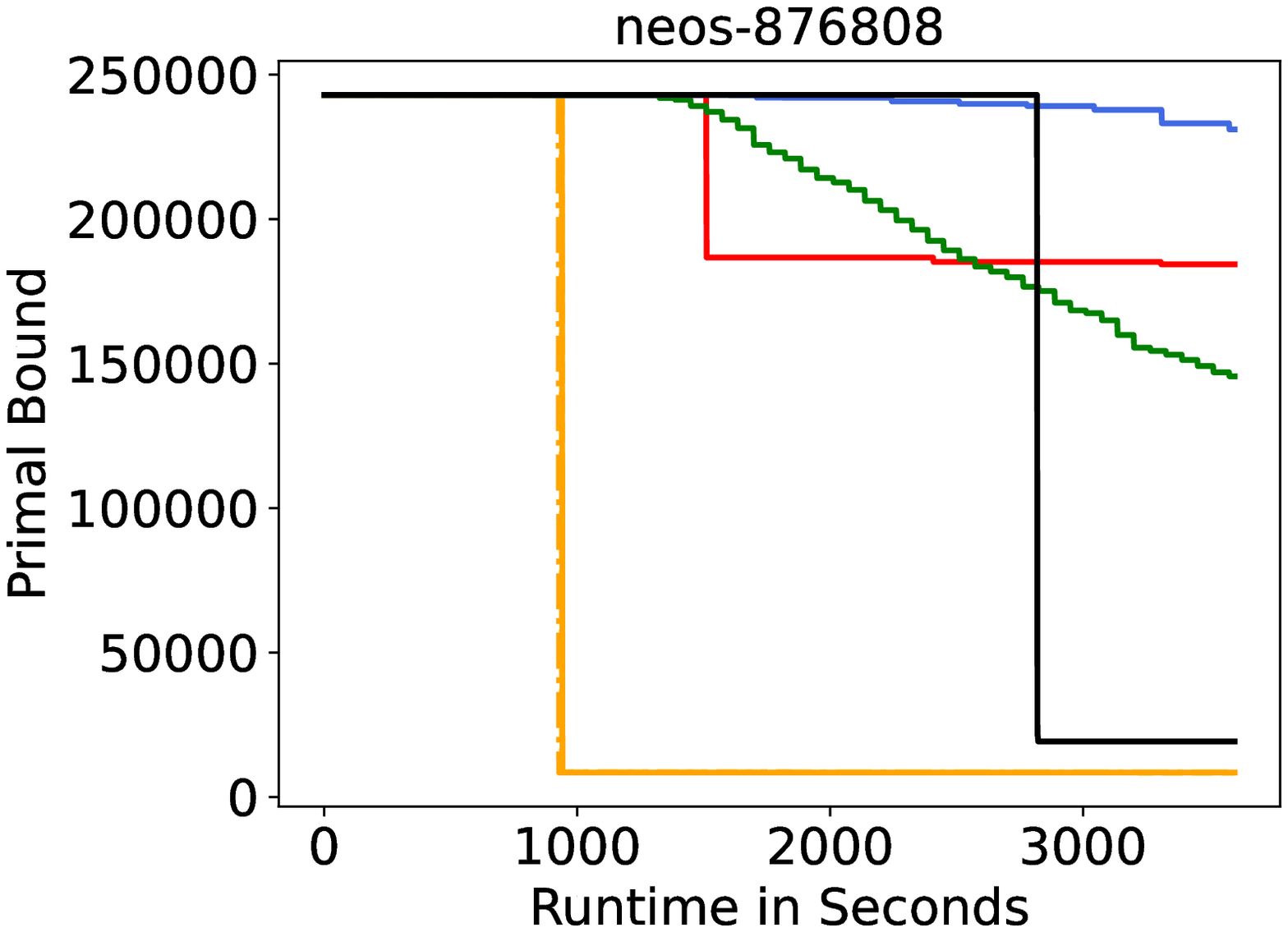}
	\end{subfigure}
	\begin{subfigure}[htbp]{0.49\textwidth}
	\centering
	\includegraphics[height=3.6cm]{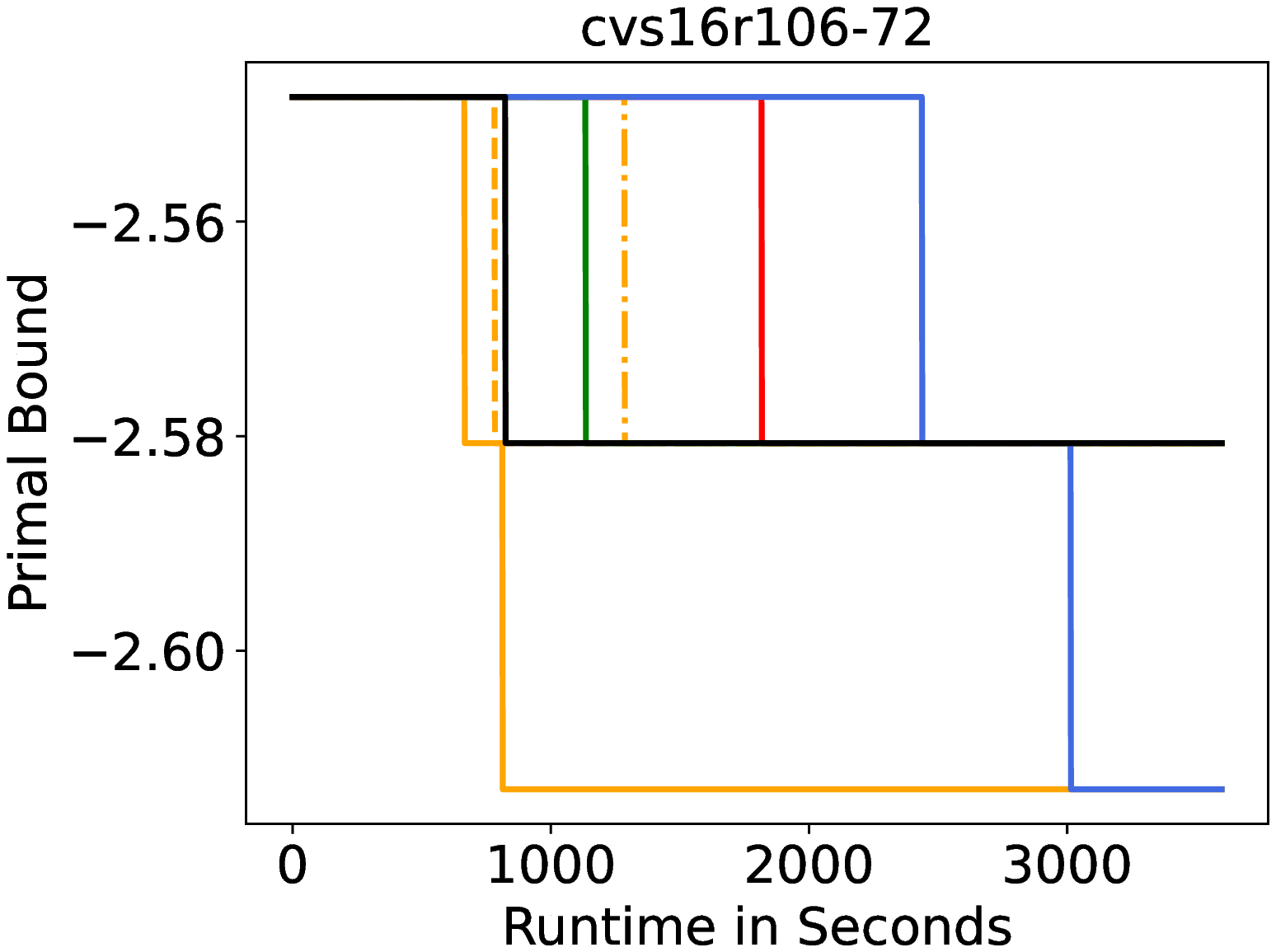}
	\end{subfigure}
	\begin{subfigure}[htbp]{0.49\textwidth}
	\centering
	\includegraphics[height=3.6cm]{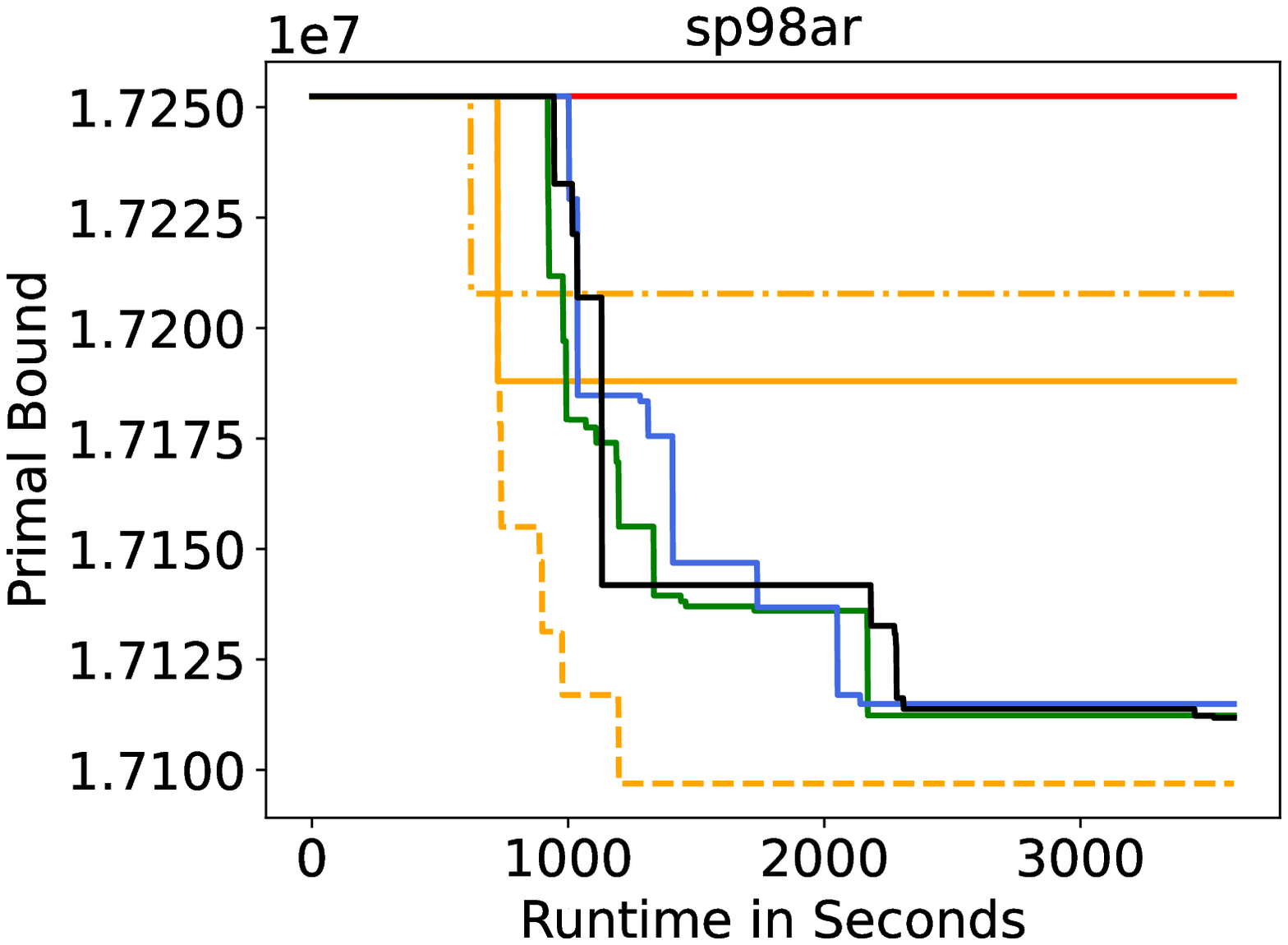}
	\end{subfigure}
	\begin{subfigure}[htbp]{0.49\textwidth}
	\centering
	\includegraphics[height=3.6cm]{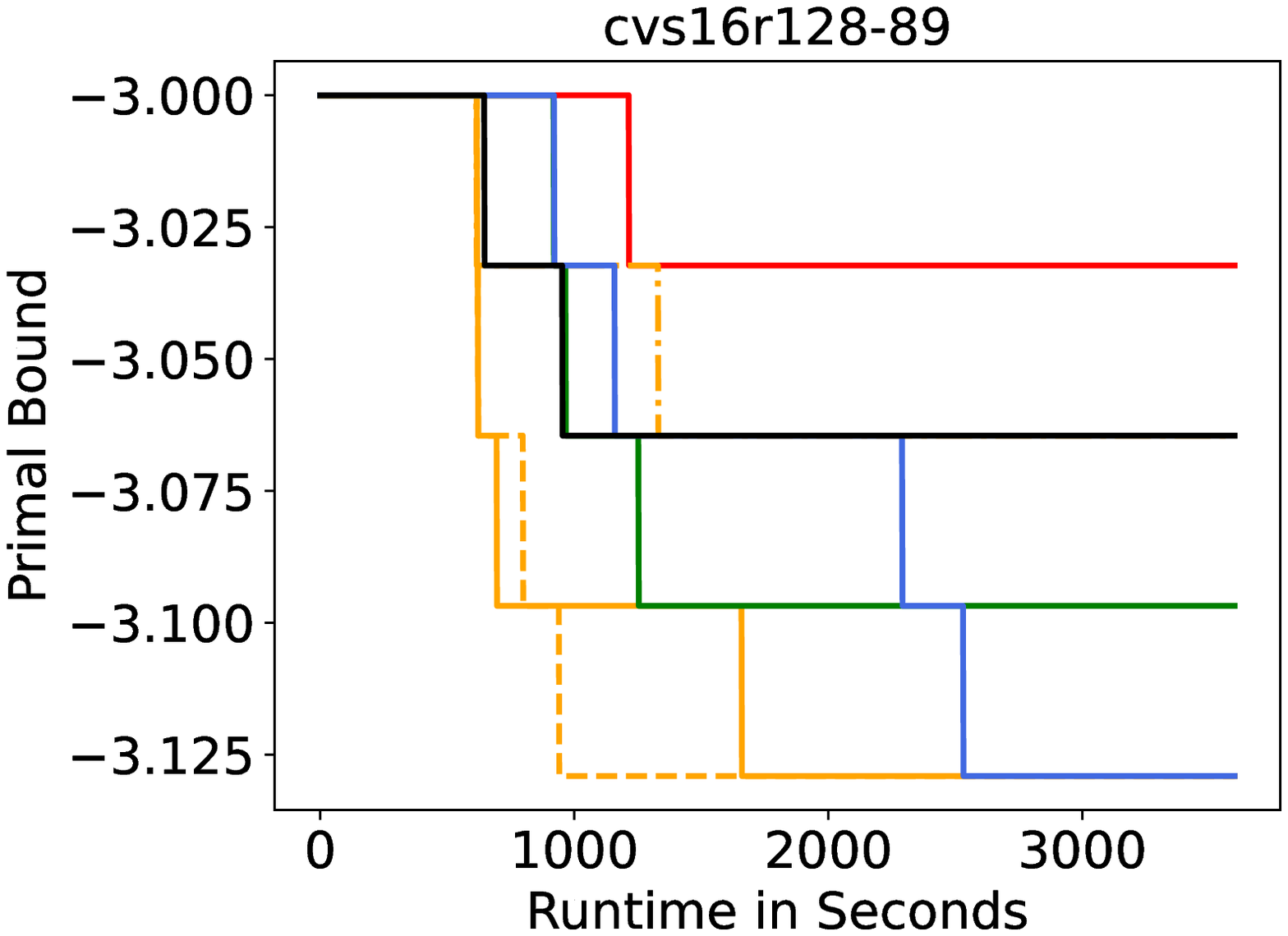}
	\end{subfigure}
    \begin{subfigure}[htbp]{0.49\textwidth}
	\centering
	\includegraphics[height=3.6cm]{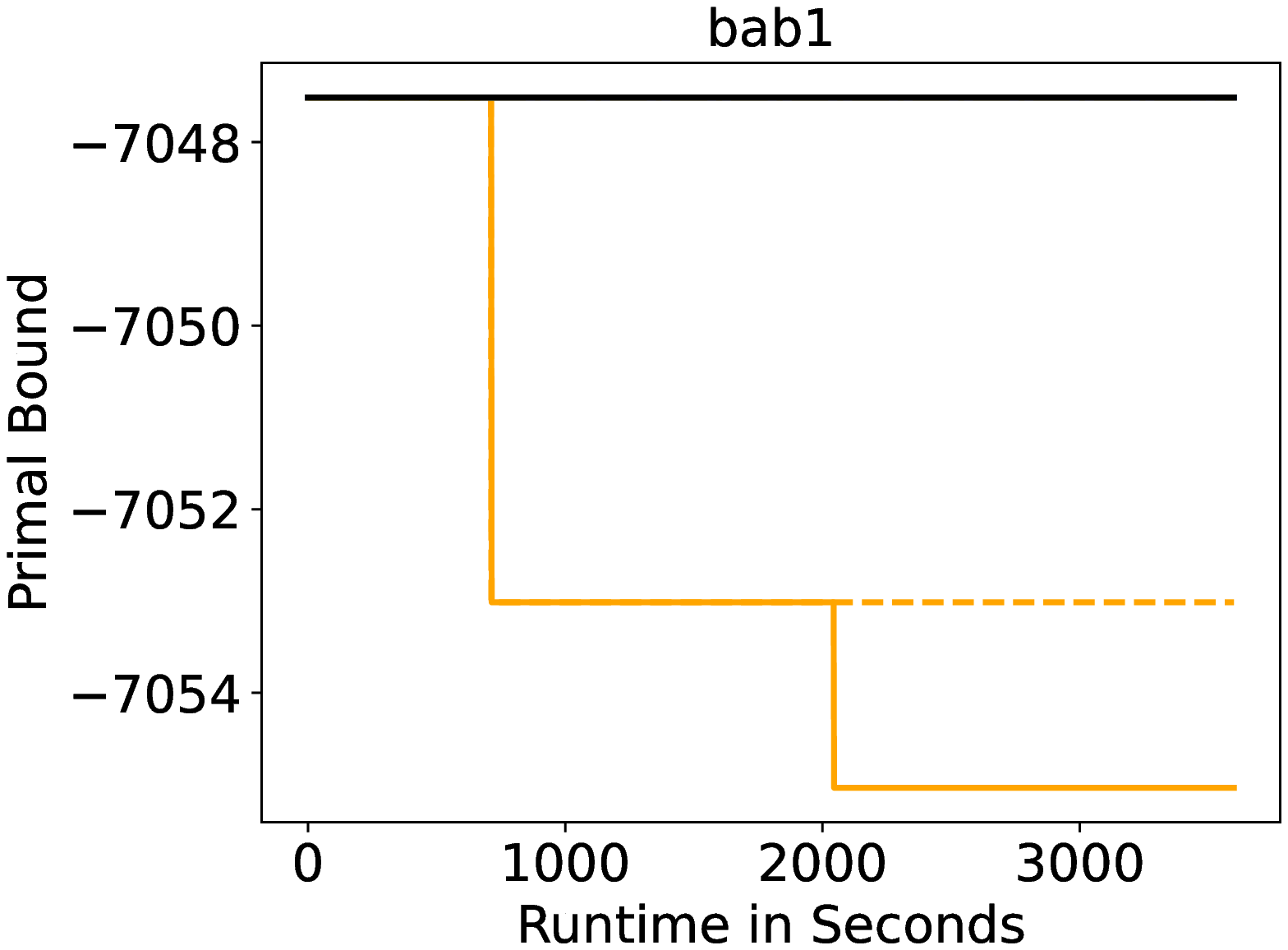}
	\end{subfigure}
	\begin{subfigure}[htbp]{0.49\textwidth}
	\centering
	\includegraphics[height=3.6cm]{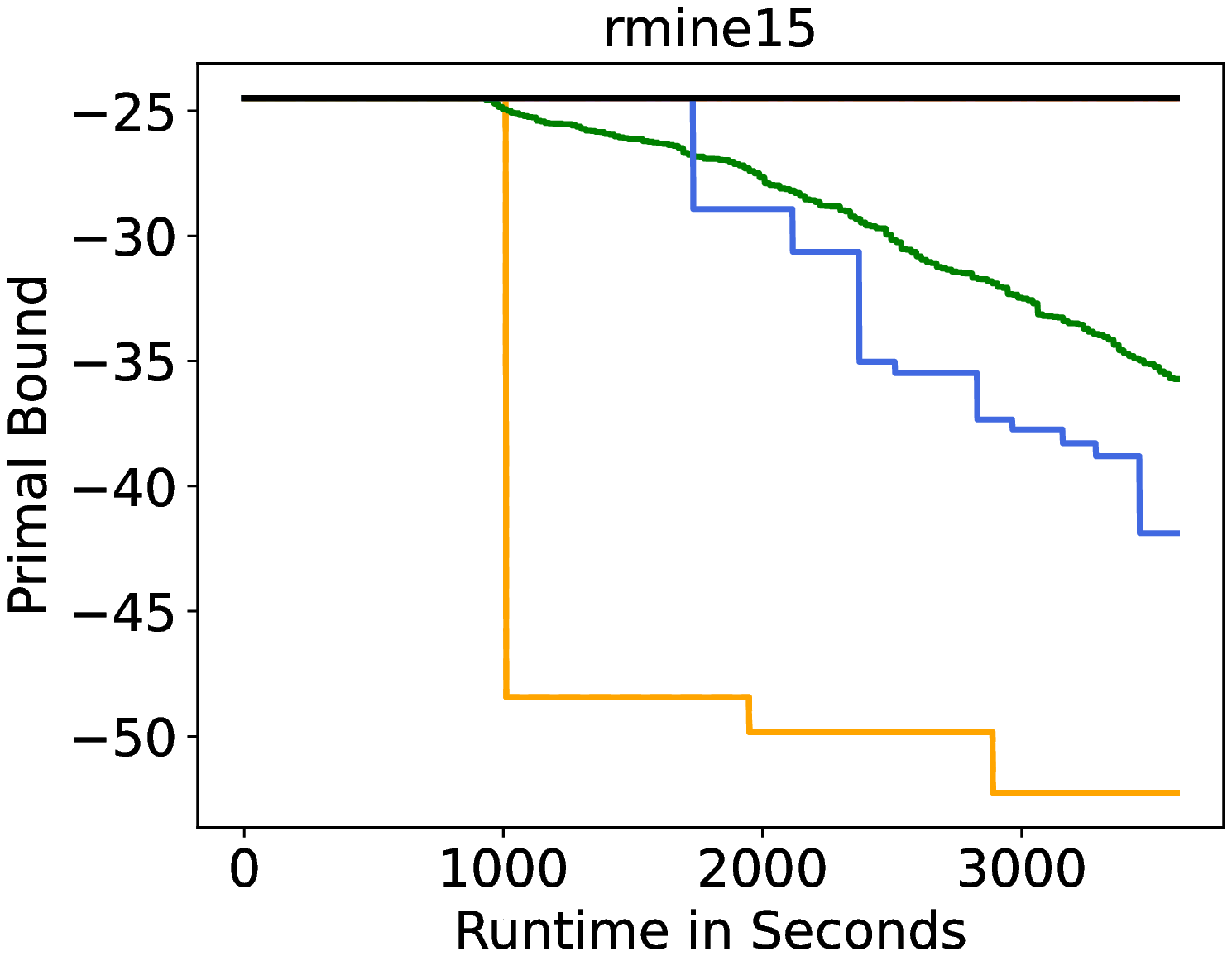}
	\end{subfigure}
    \caption{The primal bound as a function of time on selected MIPLIB instances. The titles of the plots are the instance names for which the results are presented.\label{res::individualMIPLIB}}
\end{figure}

On some instances,  \LBRELAX and its variants can significantly outperform the baselines and we present the anytime performance on those.
Figure \ref{res::individualMIPLIB} shows the primal bound as a function of time on 8 selected MIPLIB instances.

\end{document}